\documentclass[journal,10pt]{IEEEtran}
\usepackage[utf8]{inputenc} 
\usepackage{float} 
\usepackage{cite}
\ifCLASSINFOpdf
  \usepackage[pdftex]{graphicx}
  \DeclareGraphicsExtensions{.pdf,.jpeg,.png}
\else
  \usepackage[dvips]{graphicx}
  \DeclareGraphicsExtensions{.eps}
\fi
\usepackage[cmex10]{amsmath}
\usepackage{algorithmic}
\usepackage{array}

\usepackage{amssymb}
\usepackage{gensymb}
\usepackage{multirow}
\usepackage{color,array,dcolumn}
\usepackage{fixltx2e}
\usepackage{stfloats}
\usepackage{url}
\usepackage{amssymb}
\usepackage{gensymb} 
\usepackage{multirow} 

\usepackage{enumitem}
\usepackage{amsmath}
\usepackage{graphicx}
\usepackage{multirow,bigdelim}
\usepackage{booktabs}
\usepackage{textcomp}
\usepackage{adjustbox}
\usepackage{verbatim}
\usepackage{enumitem}
\usepackage{caption}
\usepackage{subcaption}
\usepackage{mathtools}
\usepackage{commands}
\usepackage{amsmath,amsfonts,amssymb}
\usepackage{xifthen}
\usepackage{tikz}
\usetikzlibrary{graphs, graphs.standard, quotes}
\usepackage{todonotes}
\usepackage{wrapfig}
\usepackage{color}
\usepackage[normalem]{ulem}

\usepackage[ruled,vlined]{algorithm2e}

\newcommand{\vect}[1]{\boldsymbol{\mathrm{#1}}}
\newcommand{\vtheta}{{\vect{\theta}}}
\newcommand{\vphi}{{\vect{\phi}}}

\newcommand{\vectall}[1]{\vect{#1}_{0:T}}

\newcommand{\qphi}{q_\vphi}

\newcommand{\rv}[1]{\textcolor{black}{ #1}}

\makeatletter
\DeclareRobustCommand\onedot{\futurelet\@let@token\@onedot}
\def\@onedot{\ifx\@let@token.\else.\null\fi\xspace}

\def\eg{\emph{e.g}\onedot} 
\def\ie{\emph{i.e}\onedot} 
 
\def\etc{\emph{etc}\onedot}

\makeatother

\begin{document}


\title{Variational Deep Learning for the Identification and Reconstruction of Chaotic and Stochastic Dynamical Systems from Noisy and Partial Observations}


\author{\IEEEauthorblockN{Duong Nguyen, ~\IEEEmembership{Member,~IEEE,}
Said Ouala, ~\IEEEmembership{Member,~IEEE,} \\
Lucas Drumetz, ~\IEEEmembership{Member,~IEEE,} and
Ronan Fablet, ~\IEEEmembership{Senior Member,~IEEE.}}


\thanks{Duong Nguyen, Said Ouala, Lucas Drumetz and Ronan Fablet are with IMT Atlantique, Lab-STICC, 29238 Brest, France (email: \{van.nguyen1, said.ouala, lucas.drumetz, ronan.fablet\}@imt-atlantique.fr)}

\thanks{This work was supported by Labex Cominlabs (grant SEACS), CNES (grant OSTST-MANATEE), Microsoft (AI EU Ocean awards), ANR Projects Melody and OceaniX. It benefited from HPC and GPU resources from Azure (Microsoft EU Ocean awards) and GENCI-IDRIS (Grant 2020-101030).
}
}

%
\maketitle

\vspace{-2mm}

\begin{abstract}
The data-driven recovery of the unknown governing equations of dynamical systems has recently received an increasing interest. However, the identification of governing equations remains challenging when dealing with noisy and partial observations. Here, we address this challenge and investigate variational deep learning schemes. Within the proposed framework, we jointly learn an inference model to reconstruct the true states of the system and the governing laws of these states from series of noisy and partial data. In doing so, this framework bridges classical data assimilation and state-of-the-art machine learning techniques. We also demonstrate that it generalises state-of-the-art methods. Importantly, both the inference model and the governing model embed stochastic components to account for stochastic variabilities, model errors, and reconstruction uncertainties. Various experiments on chaotic and stochastic dynamical systems support the relevance of our scheme w.r.t. state-of-the-art approaches.
\end{abstract}

\vspace{-2mm}
\section{Introduction}
\label{sec:introduction}

The identification of the governing equations of dynamical processes, usually stated as Ordinary Differential Equations (ODEs), Stochastic Differential Equations (SDEs) or Partial Differential Equations (PDEs),  is critical in many disciplines. For example, in geosciences, it provides the basis for the simulation of climate dynamics, short-term and medium-range weather forecast, short-term prediction of ocean and atmosphere dynamics, etc. In aerodynamics or in fluid dynamics, it is at the core of the 
design of aircraft and control systems, of the optimisation of energy consumption, etc. 

Classically, the  derivations of governing equations are  based on some prior knowledge of the intrinsic nature of the system \cite{lorenz_deterministic_1963, hilborn_chaos_2000, sprott_chaos_2003, hirsch_differential_2012}. The derived models can then be combined with the measurements (observations) to reduce errors, both in the model and in the measurements. This approach forms the discipline of Data Assimilation (DA) \cite{lahoz_data_2010}. 
However, in many cases, the underlying dynamics of the system are unknown or only partially known, while a large number of observations are available. This has motivated the development of learning-based approaches \cite{brunton_data-driven_2019}, where one aims at identifying the governing equations of a process from time series of measurements. 
Recently, the ever increasing availability of data thanks to developments in sensor technologies, together with advances in Machine Learning (ML), has made this issue a hot topic. Numerous methods have successfully captured the hidden dynamics of systems under ideal conditions, {\em i.e.} noise-free and high sampling frequency, using a variety of data-driven schemes, including analog methods \cite{lguensat_analog_2017}, sparse regression schemes \cite{brunton_discovering_2016}, reservoir computing  \cite{pathak_using_2017, pathak_model-free_2018} and neural approaches \cite{fablet_bilinear_2018, raissi_multistep_2018, qin_data_2018, ayed_learning_2019, vlachas_data-driven_2018}. However, real life data are often corrupted by noise and/or observed partially, as for instance encountered in the monitoring of ocean and atmosphere dynamics from satellite-derived observation data \cite{pierce_distinguishing_2001,johnson_very_2005, isernfontanet_diagnosis_2014}. In such situations, the above-mentioned approaches are most likely to fail to uncover unknown governing equations. 

To address this challenge, we need to jointly solve the reconstruction of the hidden dynamics and the identification of governing equations. This may be stated within a data assimilation framework \cite{bocquet_data_2019} using state-of-the-art assimilation schemes such as the Ensemble Kalman Smoother (EnKS) \cite{evensen_ensemble_2000}. Deep learning approaches are also particularly appealing to benefit from their flexibility and computational efficiency. Here, we investigate a variational deep learning framework. More precisely, we frame the considered issue as a variational inference problem with an unknown transition distribution associated with the underlying dynamical model. The proposed method generalises learning-based schemes such as \cite{fablet_bilinear_2018, nguyen_em-like_2019, nguyen_assimilation-based_2020, brajard_combining_2019, bocquet_bayesian_2020}, and also explicitly relates to data assimilation formulations. Importantly, it can account for errors and uncertainties both within the dynamical prior and the inference model. Overall, our key contributions are:
\begin{itemize}
    \item a general deep learning framework which bridges classical data assimilation and modern machine learning techniques for the identification of dynamical systems from noisy and partial observations. This framework uses variational inference and random-$n$-step-ahead forecasting, which can be considered as two complementary regularisation strategies to improve the learning of governing equations;
    \item insights on the reason why many existing methods for learning dynamical systems do not work when the available data are not perfect, {\em i.e.} noisy and/or partial;
    \item numerical experiments with chaotic systems which support the relevance of the proposed framework to improve the learning of governing equations from noisy and partial observation datasets compared to state-of-the-art schemes;
    \item numerical experiments which demonstrate that our method can also capture the characteristics of dynamical systems where stochastic factors are significant. 
\end{itemize}

The paper is organised as follows. In Section \ref{sec:problem_formulation}, we formulate the problem of learning non-linear dynamical systems. We review state-of-the-art methods and analyse their drawbacks in Section \ref{sec:related_work}. Section \ref{sec:proposed_framework} presents the details of the proposed framework, followed by the experiments and results in Section \ref{sec:experiments_results}. We close the paper with conclusions and perspectives for future work in Section \ref{sec:conclusions}.

\section{Problem formulation}
\label{sec:problem_formulation}

Let us consider a dynamical system, described by an Ordinary Differential Equation (ODE) as follows:
\begin{align}
    \frac{\mathrm{d}\vect{z}_{t}}{\mathrm{d}t} &= f\big(\vect{z}_{t}\big)
    \label{eq:ode}
\end{align}
where $\vect{z}_{t} \in \mathbb{R}^{d_z}$ is a geometrical point, called the \textit{state} of the system ($d_z$ is the dimension of $\vect{z}_{t}$), $f: \mathbb{R}^{d_z} \longrightarrow \mathbb{R}^{d_z}$ is a \textit{deterministic} function, called the \textit{dynamical model}. 

We aim at learning the dynamics of this system from some observation dataset, that is to say identifying governing equations $f$, given a series of observations $\vect{x}_{t_k}$:
\begin{align}
    \vect{x}_{t_k} &= \Phi_{t_k}\big(\mathcal{H}\big(\vect{z}_{t_k}\big)+\vect{\varepsilon}_{t_k}\big)
    \label{eq:obs}
\end{align}
where $\mathcal{H}: \mathbb{R}^{d_z} \longrightarrow \mathbb{R}^{d_x}$ is the observation operator, usually known ($d_x$ is the dimension of $\vect{x}_{t_k}$), $\vect{\varepsilon}_{t_k} \in \mathbb{R}^{d_x}$ is a zero-mean additive noise and $\{t_k\}_k$ refers to the time sampling, typically regular such that $t_k = t_0 +k \times \delta $ with respect to a fine time resolution $\delta$ and a starting point $t_0$. We introduce a masking operator $\Phi_{t_k}$ to account for the fact that observation $\vect{x}_{t_k}$ may not be available at all time steps $t_k$ ($\Phi_{t_k,j} = 0$ if the $j^{th}$ variable of $\vect{x}_{t_k}$ is missing). For the sake of simplicity, from now on in this paper, we use the notation $\vect{x}_{k}$ for $\vect{x}_{t_k}$ and $\vect{x}_{k+n}$ for $\vect{x}_{t_{k+n \times \delta}}$.   

From Eqs. \eqref{eq:ode} and \eqref{eq:obs}, we derive a state-space formulation:
\begin{align}
    \vect{z}_{k+n} &= \mathcal{F}^{n}\big(\vect{z}_{k}\big) + \vect{\omega}_{k+n}
    \label{eq:dyn_x} \\
    \vect{x}_{k} &= \Phi_{k}\big(\mathcal{H}\big(\vect{z}_{k}\big) + \vect{\varepsilon}_{k}\big)
    \label{eq:obs_y}
\end{align}
where $\vect{z}_{k+n}$ results from an integration of operator $f$ from state $\vect{z}_{k}$: $\mathcal{F}^{n}\big(\vect{z}_{k}\big) = \vect{z}_{k} + \int_{k}^{k+n}f\big(\vect{z}_{u}\big)\mathrm{d}u$. \rv{$n$ is the number of timesteps ahead that $\mathcal{F}^{n}$ forecasts, given the current state $\vect{z}_k$. $\mathcal{F}^{n}$ is hence called the $n$-step-ahead model.} $\vect{\omega}_{k} \in \mathbb{R}^{d_z} $ is a zero-mean noise process, called the \textit{model error}. $\vect{\omega}_{k}$ may come from neglected physics, numerical approximations and/or modelling errors. $\vect{\varepsilon}_{k}$ is the \textit{observation error} (or \textit{measurement error}). \rv{Note that $f$ is continuous, the time discretisation only happens because we want to calculate the integral $\mathcal{F}^{n}$ over the interval $\left[t_k, t_{k+n}\right]$. Furthermore, as detailed later, the parametrisation of $\mathcal{F}^{n}$ may explicitly depend on $n$ or not. In this paper, to simplify the notation, $k+n$ includes both $k+1$ (\ie $n=1$) and $k+n$ (\ie $n \neq 1$). If we specify $k+1$ and $k+n$ in one sentence, it means $n \neq 1$ in those contexts.}

Within this general formulation, the identification of governing equations $f$ amounts to maximising the log likelihood $\ln{p(\vectall{x})}$.


\section{Related work}
\label{sec:related_work}

The identification of dynamical systems has attracted attention for several decades and closely relates to  data assimilation (DA) for applications to geoscience. Proposed approaches typically consider a parametric model for operator $\mathcal{F}^{n}$, for example, a linear function in \cite{ghahramani_parameter_1996} or Radial Basis Functions (RBFs) in \cite{ghahramani_learning_1999}. While data assimilation mostly focuses on the reconstruction of the hidden dynamics given some observation series, a number of studies have investigated the situation where the dynamical prior is unknown. They typically learn the unknown parameters of the model using an iterative Expectation-Maximisation (EM) procedure. The E-step involves a DA scheme (\eg the Kalman filter \cite{welch_introduction_1995}, the Extended Kalman filter \cite{hoshiya_structural_1984}, the Ensemble Kalman filter \cite{evensen_data_2009}, the particle filter \cite{doucet_tutorial_2009}, etc.) to reconstruct the true states $\{\vect{z}_k\}$ from observations $\{\vect{x}_k\}$, whereas the M-step retrieves the parameters of $\mathcal{F}^n$ best describing the reconstructed state dynamics. Such methods address 
the fact that the observations may not be ideal. They can also account for model errors and uncertainties ($\vect{\omega}_k$ in Eq. \eqref{eq:dyn_x}). However, since they rely on analytic solutions, the choices of the candidates for $\mathcal{F}^n$ are generally limited. For a comprehensive introduction as well as an analysis of the limitations of those methods, the reader is referred to \cite{voss_nonlinear_2004}. 

Recently, the domain of dynamical system identification has received a new wave of contributions. Advances in machine learning open new means for learning the unknown dynamics. In this line of work, one of the pioneering contributions is the Sparse Identification of Nonlinear Dynamics (SINDy) presented in \cite{brunton_discovering_2016}. SINDy assumes that the governing equations of a dynamical model consist of only a few basic functions such as polynomial functions, trigonometric functions, exponential functions, etc. The method creates a dictionary of such candidates and uses sparse regression to retrieve the corresponding coefficients of each basic function. Under ideal conditions, SINDy can find the exact solution of Eq. \eqref{eq:ode}. The key advantage of SINDy is the interpretability of its solutions, {\em i.e.} the parametric form of the governing equations can be recovered. Another advantage is that the solutions comprise only a few terms, which improves the generalisation properties of the learnt models. However, SINDy requires the time derivative $\frac{d\vect{x}_{t}}{dt}$ to be observed. $\frac{d\vect{x}_{t}}{dt}$ might be highly corrupted by noise for noisy and partial observation datasets, which may strongly affect the performance of SINDy. Besides, it requires some prior knowledge about the considered system to create a suitable dictionary of the basic functions.

Analog methods \cite{nagarajan_evaluation_2015,mcdermott_modelbased_2016,zhao_analog_2016}, including the Analog Data Assimilation (AnDA) presented in \cite{lguensat_analog_2017}, propose a non-parametric approach for data assimilation. AnDA implicitly learns Eqs. \eqref{eq:dyn_x} and \eqref{eq:obs_y} by remembering every seen pair $\{state,successor\} = \{\vect{x}_k,\vect{x}_{k+1}\}$ and storing them in a catalog. To predict the evolution of a new query point $\vect{x}_k$, AnDA looks for $k$ similar states in the catalog, the prediction is then a weighted combination of the corresponding successors of these states.  The performance of this method heavily depends on the quality of the catalog. If the catalog contains enough data and the data are clean, AnDA provides a good and straightforward solution for data assimilation. However, since AnDA relies on a k-Nearest Neighbor (k-NN) approach, it may be strongly affected by noisy data, especially when considering high-dimensional systems.

A number of neural-network-based (NN-based) methods have been introduced recently. These methods leverage deep neural networks as universal function approximators. They vary from direct applications of standard NN architectures, such as LSTMs in \cite{yeo_deep_2019}, ResNets in \cite{qin_data_2018}, etc. to some more sophisticated designs, dedicated to dynamical systems and often referred to as neural ODE schemes \cite{chen_neural_2018,fablet_bilinear_2018,raissi_multistep_2018, rubanova_latent_2019}. The reservoir computing, whose idea is derived from Recurrent Neural Networks (RNNs), used in \cite{pathak_using_2017} and \cite{pathak_model-free_2018} can also be regarded as a NN-based model. As illustrated in \cite{chen_neural_2018, fablet_bilinear_2018,raissi_multistep_2018, rubanova_latent_2019}, through the combinations of a parametrisation for differential operator $f$ and some predefined integration schemes (\eg, explicit Runge-Kutta 4 scheme (RK4) in \cite{fablet_bilinear_2018}, black-box ODE solvers in \cite{chen_neural_2018, rubanova_latent_2019}), neural ODE schemes provides significantly better forecasting performance than that of standard NN models, especially when dealing with chaotic dynamics.
Powered by deep learning, these methods can successfully capture the dynamics of the system under ideal conditions (noise-free and regularly sampled with high frequency). However, they have the following limitations: i) the network requires fully-observed data\footnote{Latent ODE \cite{rubanova_latent_2019} can apply for data observed partially in time, however, data may be observed partially in space also.} and ii) when dealing with noisy observations, no regularisation techniques have been proved effective in preventing overfitting for dynamical system identification. 

\begin{figure}
    \centering
    \includegraphics[width=0.95\linewidth]{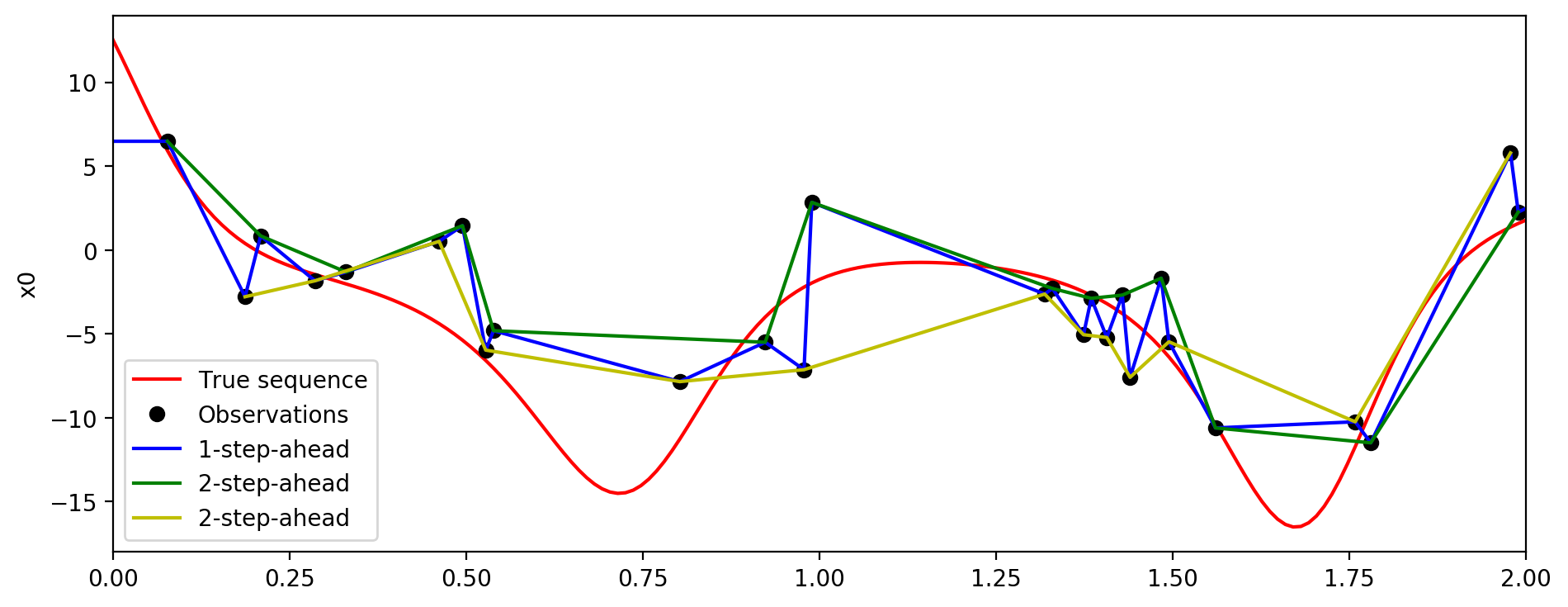}
    \caption{Problems of learning dynamical systems from imperfect data. This figure plots the first component of the Lorenz-63 system \cite{lorenz_deterministic_1963}, when the observation operator is the identity matrix. The observation is noisy and partial.  If the learning algorithm is applied directly on the observations (black dots), which are noisy and partially sampled, and a linear interpolation is used to create grid data, the dynamics seen by the network are the blue curve (for 1-step-ahead forecasting models) or the green and the yellow curves (for 2-step-ahead forecasting models, these two curves correspond to two possible starting points of the sequence) instead of the true dynamics (the red curve).}
    \label{fig:noisy_irregular_problem}
\end{figure}

Overall, the above-mentioned learning-based methods may not apply or fail when the observations are noisy and partial. Their learning step is stated as the minimisation of a short-term prediction error of the observed variables:
\begin{equation}
    \label{eq:shortterm_error}
    loss = \sum g(||\vect{x}^{pred}_{k+n} -\vect{x}_{k+n}||_2)
\end{equation}
where $\vect{x}^{pred}_{k+n} = \mathcal{F}^n(\vect{x}_{k})$ is the predicted observation at $k+n$ given the current observation $\vect{x}_k$,  $||.||_2$ denotes the $L^2$ norm, $g$ is a function of $||\vect{x}^{pred}_{k+n} -\vect{x}_{k+n}||_2$. 
As shown in Fig. \ref{fig:noisy_irregular_problem}, with this family of cost functions, the model tends to overfit the observations (the blue curve or the green and yellow curves), instead of learning the true dynamics of the system (the red curve). Another reason why those methods fail is because they violate the Markovian property of the system. Note that the process of the true states $\vectall{z}$ of the system is Markovian ({\em i.e.}, given $\vect{z}_k$, $\vect{z}_{k+1}$ does not depend on $\vect{z}_{0:k-1}$). However, when the data are damaged by noise, the process of observations $\vectall{x}$ is not Markovian. Given $\vect{x}_k$, we still need the information contained in  $\vect{x}_{0:k-1}$ to predict $\vect{x}_{k+1}$. For this reason, applying Markovian architectures like SINDy, AnDA, DenseNet, BiNN, etc. directly on the observations $\vectall{x}$ would not succeed. Models with memory like LSTMs may capture the non-Markovian dynamics in the training phase, however, in the simulation phase, they still need the memory, which implies that the learnt dynamics do not have the Markovian property of the true dynamics of the system.  

In this paper, we consider a variational deep framework which derives from a variational inference for state-space formulations (Eqs. \ref{eq:dyn_x}, \ref{eq:obs_y}). This framework accounts for uncertainty components in the dynamical prior as well as in the observation model. It jointly solves the reconstruction of the hidden dynamics and the identification of the governing equations. Importantly, this framework benefits from the computational efficiency and the modeling flexibility of deep learning frameworks for the specification of the dynamical prior and of the inference model, as well as for 
the use of a stochastic regularisation during the training phase through a randomised  $n$-step-ahead prediction loss. This framework generalises 
our recent works presented in \cite{nguyen_em-like_2019} and \cite{nguyen_assimilation-based_2020} 
and similar works, which have been developed concurrently in \cite{bocquet_data_2019}, \cite{brajard_combining_2019} and \cite{bocquet_bayesian_2020}. As detailed in the next section, \cite{bocquet_data_2019}, \cite{brajard_combining_2019} and \cite{bocquet_bayesian_2020} may be regarded as specific instances of the proposed framework with some specific settings, such as constant model error covariance matrix (we relax this hypothesis), Ensemble Kalman Smoothers for the inference scheme (we exploit both strategies: Ensemble Kalman Smoothers and NN-based schemes), EM for the optimisation (we exploit both EM and gradient-based techniques).

\section{Proposed framework}
\label{sec:proposed_framework}

In this section, we detail the proposed variational deep learning framework for the data-driven identification of the governing equations of dynamical systems from noisy and partial observations. We first present the proposed framework based on variational inference, then introduce the considered NN-based parametrisations for the dynamical prior and the inference model, along with the implemented learning scheme. We further discuss how the proposed framework relates to previous work.

\subsection{Variational inference for learning dynamical systems}

Given a series of observations $\vectall{x} = \{\vect{x}_0,..,\vect{x}_{k}\}$, instead of looking for a model $\mathcal{F}^{n}$ that minimises a loss function in a family of short-term prediction error functions as in Eq. \eqref{eq:shortterm_error}, we aim to learn operator $\mathcal{F}^{n}$ such that it maximises the log likelihood $\ln{p(\vectall{x})}$ of the observed data. We assume that $\vectall{x}$ are noisy and/or partial observations of the true states $\vectall{z}$, like in Eqs. \eqref{eq:dyn_x} and \eqref{eq:obs_y}. We can derive the log-likelihood $\ln{p(\vectall{x})}$ from the marginalisation of $\ln{p(\vectall{x},\vectall{z})}$ over $\vectall{z}$:
\begin{equation}
    \label{eq:bayesian}
     \ln{p(\vectall{x})} =\ln \int p(\vectall{x}, \vectall{z})d\vectall{z}
\end{equation}
With the exception of some simple cases, the integral in Eq. \eqref{eq:bayesian} is intractable because the posterior distribution $p(\vectall{z}|\vectall{x})$ is intractable \cite{bishop_pattern_2006}. To address this issue, Variational Inference (VI) approximates $p(\vectall{z}|\vectall{x})$ by a distribution $q$ which maximises the Evidence Lower BOund (ELBO)\footnote{For the sake of simplicity, here we present only ELBO, however, one can use other variants such as IWAE \cite{burda_importance_2016} or FIVO \cite{maddison_filtering_2017} instead.} \cite{krishnan_deep_2017, chung_recurrent_2015, fraccaro_sequential_2016}: 
\begin{align}
    \mathcal{L}(\vectall{x},p,q) &= \int q(\vectall{z}|\vectall{x}) \ln{\frac{p_\vtheta(\vectall{x}, \vectall{z})}{q(\vectall{z}|\vectall{x})}} d \vectall{z}
    \label{eq:elbo}
\end{align}
Based on the state-space formulation in Eqs. \eqref{eq:dyn_x} and \eqref{eq:obs_y}, we consider the following parametrisation for the joint likelihood $p(\vectall{x}, \vectall{z})$:
\begin{align}
    p_{\vtheta}(\vectall{x}, \vectall{z}) &= p_{\vtheta}(\vectall{z})p_{\vtheta}(\vectall{x}|\vectall{z})
    \\
    p_{\vtheta}(\vectall{z}) &= p_{\vtheta}(\vect{z}_0)\prod_{k=1}^{n-1}p_{\vtheta}(\vect{z}_{k}|\vect{z}_{k-1})\prod_{k=n}^{T}p_{\vtheta}(\vect{z}_{k}|\vect{z}_{k-n})
    \label{eq:factor_dyn}
    \\ \vspace{-10mm}
    p_{\vtheta}(\vectall{x}|\vectall{z}) &= \prod_{k=0}^{T}p_{\vtheta}(\vect{x}_k|\vect{z}_k)
    \label{eq:factor_generative}    
    \\ \vspace{-2mm}
    q_{\vphi}(\vectall{z}|\vectall{x}) &= \prod_{k=0}^T q_{\vphi}(\vect{z}_k|\vect{z}^f_k,\vectall{x})
    \label{eq:factor_q}
\end{align}
with $\vtheta$ and $\vphi$ are the sets of parameters of $p$ and $q$, respectively; \rv{$\vect{z}^f_k$ is the state forecast by $\mathcal{F}^{1}$ given $\vect{z}_{k-1}$ for $k = 1..n-1$, and by $\mathcal{F}^{n}$ given $\vect{z}_{k-n}$ for $k = n..T$}

The distributions in Eqs. \eqref{eq:factor_dyn}, \eqref{eq:factor_generative} and \eqref{eq:factor_q} are respectively the classic distributions of a state-space formulation: 1) the $n$-step ahead transition (or dynamic or prior) distribution $p_{\vtheta}(\vect{z}_{k+n}|\vect{z}_k)$ (including $n=1$); 2) the emission (or observation) distribution $p_{\vtheta}(\vect{x}_k|\vect{z}_k)$; and 3) the inference (or posterior) distribution $q_{\vphi}(\vect{z}_k|\vect{z}^f_k,\vectall{x})$. To better constrain the time consistency of the learnt dynamics, the considered 
dynamical prior embeds an $n$-step-ahead forecasting model. Given an initialisation $\vect{z}_0$, it first applies a one-step-ahead prior to propagate the initial state to the first $n$ timesteps. The application of the $n$-step-ahead prior then follows to derive the joint distribution over the entire time range $\{ 0,..,T \}$.

By explicitly separating the transition, the inference and the generative processes, the proposed framework is fully consistent with the underlying state-space formulation and the associated Markovian properties. Especially, the prior $p_{\vtheta}(\vect{z}_{k+n}|\vect{z}_k)$ will embed a Markovian architecture; by contrast, the posterior $q_{\vphi}(\vect{z}_k|\vect{z}^f_k,\vectall{x})$ shall capture the non-Markovian characteristics of the observed data. Given the learnt model, the generation of simulated dynamics only relies on the dynamical prior $p_{\vtheta}(\vect{z}_{k+1}|\vect{z}_k)$ to simulate state sequences, which conform to the Markovian property. Overall, for a given observation dataset, the learning stage comes to  maximising Eq. \eqref{eq:elbo} w.r.t. both $\vphi$ and $\vtheta$, which comprise all the parameters of the inference and generative models, {\em i.e.} the parameters of $\mathcal{F}$, $\mathcal{H}$, $\vect{\omega}_{k}$ and $\vect{\epsilon}_k$.

So far we have introduced the general form of the proposed variational inference framework for learning dynamical systems from noisy and potentially partial observations. In the following sub-sections, we will analyse some specific instances of the proposed framework and provide insights into the associated implicit hypotheses.

\subsection{Parametrisation of generative model $p_{\vtheta}$}
\label{subsec:phi}

Model $p_{\vtheta}$ involves two sets of parameters:
(i) $\vtheta_z$---the parameters of transition distribution $p_{\vtheta}(\vect{z}_{k+n}|\vect{z}_k)$ and (ii) $\vtheta_x$---the parameters of emission distribution $p_{\vtheta}(\vect{x}_{k}|\vect{z}_k)$.

Regarding the latter, similarly to \cite{brajard_combining_2019} and \cite{bocquet_bayesian_2020}, we assume the observation noise to be a white noise process with a multivariate covariance $\mathbf{R}$ such that  $p_{\vtheta}(\vect{x}_{k}|\vect{z}_k)$ is a 
conditional multivariate normal distribution:
\begin{equation}
    p_{\vtheta}(\vect{x}_{k}|\vect{z}_k) = \mathcal{N}(\mathcal{H}(\vect{z}_k),\mathbf{R})
    \label{eq:p_gen}
\end{equation}
We may consider different experimental settings, with known or unknown observation operator $\mathcal{H}$. 

Regarding the $n$-step-ahead dynamical prior  $p_{\vtheta}(\vect{z}_{k+n}|\vect{z}_k)$ (including $n=1$), we consider a conditional Gaussian distribution where the mean path is driven by the governing equation $\mathcal{F}^n$: $\vect{z}_{k+n} = \mathcal{F}^n(\vect{z}_k)$ and the dispersion is represented by a covariance matrix $\mathbf{Q}_{k}$ (usually called the \textit{model error covariance} in DA) \cite{evensen_data_2009}. Any state-of-the-art architecture for learning dynamical systems can be used to model $\mathcal{F}^n$. Here, we consider NN-based methods associated with explicit integration schemes. To account for second-order polynomial model, as proposed in  \cite{fablet_bilinear_2018}, we consider a bilinear architecture to model $f$ in Eq. \eqref{eq:ode} and a NN implementation of the RK4 integration scheme to derive the flow operator in Eq. \eqref{eq:dyn_x}. Regarding the covariance dynamics, the covariance matrix $\mathbf{Q}_{k}$ is approximated by a diagonal matrix $diag(\vect{d}^f_k)$, with $\vect{d}^f_k$ the output of a MultiLayer Perceptron (MLP):
\begin{equation}
    \label{eq:d_dyn}
    \vect{d}^f_{k} = MLP^{var\_dyn}(\vect{z}_{k-n},\mathcal{F}^n(\vect{z}_{k-n}))
\end{equation}

\subsection{Parametrisation of inference model $q_{\vphi}$}
\label{subsec:q}

\begin{figure}
    \centering
    \includegraphics[trim=30 30 0 40, width=1.0\linewidth]{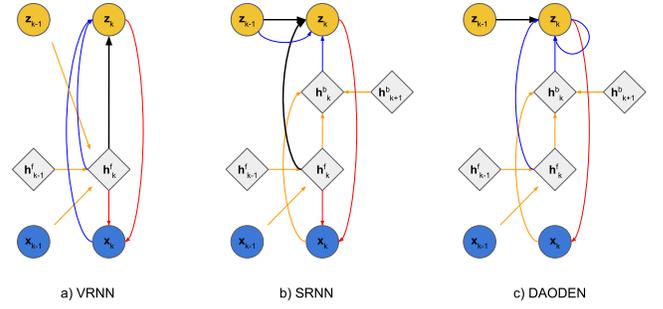}
    \caption{Architecture of VRNN, SRNN and DAODEN when $n=1$. We denote as $\vect{x}_k$ the observations, $\vect{z}_k$  the system's states, $\vect{h}^f_k$ the latent states of the forward LSTM and $\vect{h}^b_k$ the latent states of the backward LSTM. The black,  red, blue  and orange arrows denote respectively the transition of the system's states, the emission of the observations, the inference of the system's states and recurrence of the LSTMs, respectively. In VRNN (a) and SRNN (b), the dynamic $\vect{z}_k \rightarrow \vect{z}_{k+1}$ is not independent of the observation $\vect{x}_k$. The generation of the observation is also entangled with the recurrence of the LSTMs.}
    \label{fig:architecture}
\end{figure}

There is no restriction for the parametrisation of posterior $q_\vphi$. However, the parametrisation clearly affects the performance of the overall optimisation. Here, we investigate two strategies for $q_\vphi$: 1) an Ensemble Kalman Smoother (EnKS) \cite{evensen_ensemble_2000} and 2) an LSTM Variational Auto Encoder (LSTM-VAE). 
The former is a classic DA scheme that is widely used in many domains in which dynamical systems play an important role, for example in geosciences \cite{khare_investigation_2008}. We use the implementation presented in \cite{evensen_ensemble_2000}. The latter is a modern NN architecture, which has been proven effective for modelling stochastic sequential data \cite{chung_recurrent_2015} \cite{fraccaro_sequential_2016}. The backbone of the LSTM-VAE is a bidirectional LSTM which captures the long-term correlations in data. Specifically, we parameterise the inference scheme as follows: the forward LSTM is given by:
\begin{equation}
    \label{eq:lstm_f}
    \vect{h}^f_k = lstm(\vect{h}^f_{k-1},MLP^{enc}(\vect{x}^f_{k-1}))
\end{equation}
and the backward LSTM is given by:
\begin{equation}
    \label{eq:lstm_b}
    \vect{h}^b_k = lstm(\vect{h}^b_{k+1},\vect{h}^f_{k},MLP^{enc}(\vect{x}^f_{k}))
\end{equation}
where $\vect{h}^f_k$, $\vect{h}^b_k$ are the hidden states of the forward and backward LSTMs, respectively; $lstm$ is the recurrence formula of LSTM \cite{hochreiter_long_1997}; $MLP^{enc}$ is an encoder parameterised by an MLP. 
We parameterise the posterior $q_\vphi$ by a conditional Gaussian distribution with mean $\vect{\mu}^q_k$ and a diagonal covariance matrix $diag(\vect{d}^q_k)$:
\begin{align}
    q_\vphi(\vect{z}_k) &= \mathcal{N}(\vect{\mu}^q_k, diag(\vect{d}^q_k))
    \\
    \vect{\mu}^q_k, \vect{d}^q_k &= MLP^{dec}(\mathcal{F}^n(\vect{z}_{k-n}),\vect{h}^f_k,\vect{h}^b_k)
    \label{eq:q_dec}
\end{align}
with $MLP^{dec}$ is a decoder parameterised by an MLP. Note that in Eq. \eqref{eq:q_dec}, $q_\vphi$ depends on $\vect{z}^f_k = \mathcal{F}^n(\vect{z}_{k-n})$. This idea is inspired by DA, where $\mathcal{F}^n(\vect{z}_{k-n})$ is analogous to the forecasting step and $q_\vphi(\vect{z}_k|\vect{z}^f_k,\vectall{x})$ is analogous to the analysis step, which depends on the forecasting step. The whole model, called Data-Assimilation-based ODE Network (DAODEN) is illustrated in Fig. \ref{fig:architecture}. 

To our knowledge, DAODEN is the first stochastic RNN-based  model introduced for the identification of dynamical systems from noisy and partial observations. In this respect, the model used in \cite{yeo_deep_2019} is a purely deterministic RNN-based network. However, similar architectures have been used in Natural Language Processing (NLP) such as the Variational Recurrent Neural Network (VRNN) presented in \cite{chung_recurrent_2015}, the Sequential Recurrent Neural Network (SRNN) presented in \cite{fraccaro_sequential_2016}. Fig. \ref{fig:architecture} shows how DAODEN differs from those architectures. The main difference is that the transition $\vect{z}_k \rightarrow \vect{z}_{k+1}$ is independent of  observation $\vect{x}_k$ ({\em i.e.} the dynamic is autonomous). Besides, the emission $\vect{z}_k \rightarrow \vect{x}_{k}$ is also independent of the historical state ${\vect{z}_{0},..,\vect{z}_{k-1}}$. These differences relate to domain-related priors. 
In dynamical systems theory and associated application domains such as geoscience, the underlying dynamics follow physical principles. Therefore, they are autonomous and are not affected by the measurements (the observations). As a consequence,  $\vect{z}_{k+1}$ does not depend on $\vect{x}_{1:k-1}$ conditionally to $\vect{z}_k$. At a given time $k$, observation $\vect{x}_k$ is a measurement of state $\vect{z}_k$ of the system, this measurement does not depend on any other state $\vect{z}_{k' \neq k}$, {\em i.e.} given $\vect{z}_k$, $\vect{x}_k$ and $\vect{z}_{k'}$ are independent for any $k' \neq k$. For this reason, architectures used in NLP like VRNN, SRNN do not apply for dynamical system identification.

{}

\subsection{Objective function}

Following a variational Bayesian setting, the learning phase comes to minimising a loss given the negative of ELBO:
\begin{equation}
    loss_{ELBO} = - \mathcal{L}(\vectall{x},p_{\vtheta},q_{\vphi})
    \label{eq:loss_elbo}
\end{equation}

Instead of solving Eq. \eqref{eq:elbo}, one can solve its Maximum A Posteriori (MAP) solution by restricting $q_\vphi$ to Dirac distributions:
\begin{multline}
    \mathcal{L}_{MAP} = \sum_{k=0}^{T} \ln{p_{\vtheta}(\vect{x}_k|\vect{z}^*_k)} \\
    + \ln{p_{\vtheta}(\vect{z}^*_{0})}
    + \sum_{k=1}^{n-1}  \ln{p_{\vtheta}(\vect{z}^*_{k}|\vect{z}^*_{k-1})} 
    + \sum_{k=n}^{T}  \ln{p_{\vtheta}(\vect{z}^*_{k}|\vect{z}^*_{k-n})}
    \label{eq:elbo_MAP}
\end{multline}
with $\vect{z}_k^* =  \mathbb{E}\left[ \qphi(\vect{z}_k|\vect{z}^f_k,\vectall{x}) \right]$ if $q_\vphi$ is parameterised by an EnKS and $\vect{z}_k^* =  \qphi(\vect{z}_k|\vect{z}^f_k,\vectall{x}) = \delta(\vect{z}_k|\vect{z}^f_k,\vectall{x})$ if $q_\vphi$ is parameterised by a neural network.
If we remove the covariance part in Eq. \eqref{eq:q_dec}, the LSTM-VAE becomes an LSTM Auto Encoder (LSTM-AE) :
\begin{equation}
    \label{eq:q_dirac}
    \vect{z}_k^* = \vect{\mu}^q_k = dec(\mathcal{F}^n(\vect{z}_{k-n}),\vect{h}^f_k,\vect{h}^b_k)
\end{equation}
The MAP loss function, which relates to the weak-constraint 4D-Var in DA \cite{courtier_strategy_1994}, is given by:
\begin{equation}
    loss_{MAP} = - \mathcal{L}_{MAP}(\vectall{x},p_{\vtheta},q_{\vphi})
    \label{eq:loss_MAP}
\end{equation}
This is the objective function used in \cite{bocquet_data_2019}, \cite{brajard_combining_2019} and \cite{bocquet_bayesian_2020}, with the assumption that $\mathbf{Q}_k$ is time invariant, {\em i.e.} $\mathbf{Q}_k = \mathbf{Q}$.

One may further assume that the covariance matrices of transition distribution $p_{\vtheta}(\vect{z}^*_{k}|\vect{z}^*_{k-n})$ and the covariance matrices of the observation distribution $p_{\vtheta}(\vect{x}_k|\vect{z}^*_k)$ are diagonal and constant, both in time and in space, Eq. \eqref{eq:elbo_MAP} then becomes\footnote{The derivation of \eqref{eq:elbo_determ} can be found in our previous paper \cite{nguyen_em-like_2019}.}:
\begin{multline}
    \mathcal{L}_{determ} = 
    -\lambda\sum_{k=0}^{T}||\vphi_k(\mathcal{H}(\vect{z}^*_k)) - \vect{x}_{k}||_2^2 
    \\
    -\sum_{k=1}^{n-1} ||\vect{z}^*_{k} - \mathcal{F}^1(\vect{z}^*_{k-1}) ||_2^2
    -\sum_{k=n}^{T} ||\vect{z}^*_{k} - \mathcal{F}^n(\vect{z}^*_{k-n})||_2^2 
    \label{eq:elbo_determ}
\end{multline}
The associated loss function is given by:
\begin{equation}
    loss_{determ} = - \mathcal{L}_{determ}(\vectall{x},p_{\vtheta},\qphi)
    \label{eq:loss_determ}
\end{equation}
which is the objective function used in \cite{nguyen_em-like_2019} and \cite{nguyen_assimilation-based_2020}.  We may note that if $\vect{x}_k = \vect{z}_k$, \eqref{eq:loss_determ} becomes the short-term prediction error widely used in the literature \cite{pathak_using_2017, fablet_bilinear_2018, qin_data_2018, pathak_model-free_2018}. In other words, \cite{pathak_using_2017, fablet_bilinear_2018, qin_data_2018, pathak_model-free_2018} implicitly suppose that the observations are ideal.

\subsection{Optimisation strategy}
\label{subsec:optimisation_strategy}

To learn parameters $\vtheta$ and $\vphi$ (\ie the parameters of the generative and the inference models), there are two optimisation strategies: 1) alternatively optimise $\vtheta$ then $\vphi$ (Expectation-Maximisation-like or EM-like) to minimise the loss function; or 2) jointly optimise the loss function over $\vtheta$ and $\vphi$. 

For models whose posterior $q_\vphi$ is implemented by an EnKS, since EnKS uses analytic formulas and the NN-based parametrisation of $p_\vtheta$ is usually optimised by Gradient Descent (GD) techniques, we consider an alternated EM procedure as the optimisation strategy for the whole model. In the E-step, the EnKS computes the posterior $q_\vphi$, represented by an ensemble of states $\vect{z}^{(i)}_k$. Given this ensemble of states, the M-step minimises the loss function over $\vtheta$ using a stochastic gradient descent algorithm.

For DAODEN settings, we can fully benefit from the resulting end-to-end architecture, as both generative model $p_\vtheta$ and posterior model $q_\vphi$ are parameterised by neural networks, to jointly optimise all model parameters using a stochastic gradient descent technique. The gradient descent technique may be regarded as a particular case of EM where the M-step takes only one single gradient step. For NN-based models, gradient descent strategies usually work better than EM \cite{goodfellow_deep_2016}. 

\subsection{Random-$n$-step-ahead training}
\label{subsec:random_n_step_ahead}

Within the considered framework, we noted experimentally that the model may overfit the data, when the number of the forecasting timesteps is fixed. 
For example, if the observation operator $\mathcal{H}$ is an identity matrix, a possible overfitting situation is when the inference scheme also becomes an identity operator: $\mathbb{E}\left[q_\vphi(\vect{z}_k|\vectall{x})\right] \rightarrow \vect{x}_k$. In such situations, the dynamics seen by the dynamical sub-module would be the noisy dynamics. 

To deal with these overfitting issues, we further exploit the flexibility of the proposed $n$-step-ahead dynamical prior during the training phase. For each mini-batch iteration in the training phase, we draw a random value of $n$ between 1 and a predefined maximum number of forecasting steps $n$-step-ahead\_max. We then apply a gradient descent step with the sampled value of $n$. The resulting randomised training procedure is  detailed in Alg. \ref{alg:random_n_step_ahead}.

This randomised procedure is regarded as a regularisation procedure to fit a time-consistent dynamical operator $\mathcal{F}^{n}$. We noted in previous works that neural ODE schemes may not distinguish well the dynamical operator from the integration scheme \cite{ouala_residual_2019}. Here, through the randomisation of parameter $n$, we constrain the end-to-end architecture by applying it for different prediction horizons, which in turn constrains the identification of the dynamical model $f$. Asymptotically, the proposed procedure would be similar to a weighted sum of loss \eqref{eq:loss_elbo} computed for different values of $n$, which have been proposed for the data-driven identification of governing equations in the noise-free case \cite{raissi_multistep_2018}.  

\begin{algorithm}
\caption{Random-$n$-step-ahead training.}
\label{alg:random_n_step_ahead}
 \KwResult{The set of parameters $\{ \vtheta, \vphi \}$ of the learnt model.}
 \textbf{Inputs}: $\vectall{x}$, $\vect{z}_0$, $\mathbf{R}$, the initial values of $\{ \vtheta, \vphi \}$, $n$-step-ahead\_max, n\_iteration\_max\;
 
 iter = 0\;
 
  \While{iter $<$ n\_iteration\_max}{
  t = 0\;
  
 $n$-step-ahead = randint(1,$n$-step-ahead\_max)\;
 
 \While{$t < k-n$}{
  \eIf{$t <$ $n$-step-ahead $- 2$}{
  $n = 1$\;
  }{
  $n = $ $n$-step-ahead\;
  }
  $\vect{z}^f_{k+n} = \mathcal{F}^n(\vect{z}_k)$\;
  
  $\vect{d}^f_{k+n} = MLP^{var\_dyn}(\vect{z}_k,\mathcal{F}^n(\vect{z}_k))$\;
  
  $p_\vtheta(\vect{z}_{k+n}|\vect{z}_k) = \mathcal{N}(\vect{z}^f_{k+n}, \vect{d}^f_{k+n})$\;
  
  Calculate $q_\vphi(\vect{z}_{k+n}|\vect{z}^f_{k+n},\vectall{x})$\;
  
  Sample $\vect{z}_{k+n} \sim q_\vphi(\vect{z}_{k+n}|\vect{z}^f_{k+n},\vectall{x})$\;
  
  $p_\vtheta(\vect{x}_{k+n}|\vect{z}_{k+n}) = \mathcal{N}(\mathcal{H}(\vect{z}_{k+n}), \mathbf{R})$\;  
 }
 {
 Calculate $loss$\;
 
 Optimise $loss$ w.r.t. $\{ \vtheta, \vphi \}$\;
 }
 }
\end{algorithm}

\subsection{Initialisation by optimisation}

In this section, we present the initialisation technique used in the experiments in this paper. Although this technique is not compulsory, it improves the stability of the training.  

To calculate the state of the system at any time $k$, we need both the true dynamics and the precise initial condition $\vect{z}_{0}$. If we use DAODEN, we also have to initialise $\vect{h}^f_{0}$ and $\vect{h}^b_{T+1}$. The common approach is ``wash out" \cite{jaeger_tutorial_2002}, \ie to initialise $\vect{h}^f_{0}$ and $\vect{h}^b_{T+1}$ to zeros or random values and run the LSTMs until the effect of the initial values disappears. However, this initialisation technique may not be suitable for learning dynamical systems, because during the wash out period, the network is not stable, especially when using an explicit integration scheme (here is the RK4). These instabilities may make the training fail. The value of the objective function also varies highly during this period, leading to an unreliable outcome of the final loss. 

Sharing a similar idea with \cite{rubanova_latent_2019} and \cite{mohajerin_multistep_2019}, we use a different initialisation strategy. We add two auxiliary networks, a Forward Auxiliary Net to provide $\vect{h}_0$ and $\vect{z}_0$, and a Backward Auxiliary Net to provide  $\vect{h}_{T+1}$ for the main model. Each auxiliary network is an LSTM. We use one segment at the beginning of the sequence and one segment at the end of the sequence as the inputs of these networks. 
\begin{figure}
    \centering
    \includegraphics[trim = 40 130 40 50, width=1.0\linewidth]{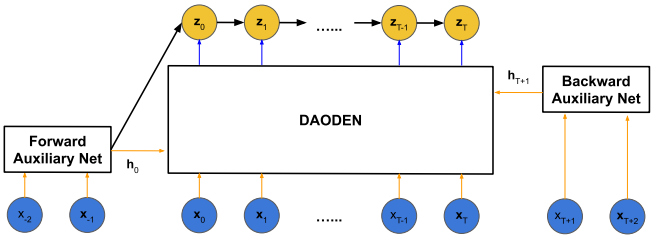}
    \caption{Initialization by optimisation. An auxiliary network is added for the initialization of $\vect{x}_0$ and $\vect{h}_0$.}
    \label{fig:auxiliary_net}
\end{figure}{}

\section{Experiments and results}
\label{sec:experiments_results}

In this section, we report numerical experiments to evaluate the proposed framework. We include a comparison with respect to state-of-the-art methods. Beyond the application to deterministic dynamics as considered in previous work \cite{nguyen_em-like_2019, nguyen_assimilation-based_2020, bocquet_data_2019, bocquet_bayesian_2020, brajard_combining_2019, pathak_model-free_2018, pathak_using_2017, qin_data_2018}, we also investigate an application to stochastic dynamics and a reduced-order modelling, where observation operator $\mathcal{H}$ is unknown. As case-study models, we focus on Lorenz-63 and Lorenz-96 dynamics, which provides a benchmarking basis w.r.t. previous work \cite{brunton_data-driven_2019, lguensat_analog_2017, fablet_bilinear_2018, champion_data-driven_2019}.

\subsection{Benchmarking dynamical models}
\label{subsec:datasets}

We report numerical experiments for three chaotic dynamical systems: a Lorenz 63 system (L63) \cite{lorenz_deterministic_1963}, a Lorenz 96 system (L96) \cite{lorenz_predictability:_1996} and a stochastic Lorenz 63 system (L63s)  \cite{chapron_large-scale_2018}. The details of the L63, the L96 and the L63s are presented in the Appendices. Note that these models are chaotic, {\em i.e.} they are highly sensitive to initial conditions such that a small difference in a state may lead to significant changes in future. Because of this chaotic nature, applying directly standard deep neural network architectures would not be successful. 

We chose the L63 as a benchmarking system because of its famous butterfly attractor. The system involves 3-dimensional states, making it easy to visualise for a qualitative interpretation. Experiments on the L96 provides a means to evaluate how the proposed schemes can scale up to higher-dimensional systems. The last system---the L63s, is considered to show the benefit of stochastic architectures over deterministic ones.   

For each system, we generated 200 sequences of length of 150 using 200 different initial conditions $\vect{z}_0$ with time step $\delta = 0.01$, $\delta = 0.05$ and $\delta = 0.01$ for the L63, L96 and L63s, respectively\footnote{This is the setting used in \cite{champion_data-driven_2019}}. In total, the training set of each system comprises 30000 points. Those training sets are relatively small in comparison with those in \cite{champion_data-driven_2019} (512000 points) and \cite{brajard_combining_2019} (40000 points). Another setting when we generated only one long sequence of length of 4000 from one initial condition $\vect{z}_0$, then split it into smaller segments of length of 150 also gave similar results\footnote{This is the setting used in \cite{lguensat_analog_2017}, \cite{pathak_model-free_2018},  \cite{brajard_combining_2019} \cite{bocquet_bayesian_2020}} (not reported in this paper). 

For the test sets, we generated 50 sequences of length of 150 using 50 different initial conditions $\vect{z}_0$ which are not observed in the training set. Let us recall that the true hidden states $\vectall{z}$ of sequences are never used during the training phase, however, they are used in the test phase to give a quantitative evaluation. 
As in \cite{nguyen_em-like_2019, nguyen_assimilation-based_2020, bocquet_data_2019, brajard_combining_2019, bocquet_bayesian_2020}, we first consider an experimental setting where $\mathcal{H}$ is an identity operator, and $\vect{\varepsilon}_k$ a zero-mean Gaussian white noise. We tested several signal-to-noise ratio values $r = \frac{std_{\epsilon}}{std_{\vect{z}}}$. Then we tested the proposed framework on a setting where $\mathcal{H}$ is unknown, as in \cite{champion_data-driven_2019}.

\subsection{Baseline schemes}
\label{subsec:baselines}

In the reported experiments, we considered different state-of-the-art schemes for benchmarking purposes, namely the Analog Data Assimilation (AnDA) \cite{lguensat_analog_2017}, the Sparse Identification of Nonlinear  Dynamics (SINDy) \cite{brunton_discovering_2016}, the Bilinear Neural Network (BiNN) \cite{fablet_bilinear_2018}, and the Latent ODE \cite{rubanova_latent_2019}, the latter being among the state-of-the-art schemes in the deep learning literature. As explained earlier in this paper, regardless of the network architecture, as long as the objective function does not take into account the fact that the observations are noisy and potentially partial, the method would not work. BiNN and Latent ODE embed the true solution of the L63 and the L96. Under ideal conditions, they should work as well as other NN-based ODE models (such as those in \cite{raissi_multistep_2018}, \cite{qin_data_2018}, \cite{yeo_deep_2019}, \etc) do. The difference between BiNN and Latent ODE is BiNN uses an explicit integration scheme (the RK4), while Latent ODE uses a black-box ODE solver. Latent ODE also uses an additional network to infer the initial condition $\vect{z}_{0}$. Since VRNN \cite{chung_recurrent_2015} and SRNN \cite{fraccaro_sequential_2016} are not designed for dynamical system identification (no autonomous dynamics in the hidden space), we do not consider these architectures in this paper.

\subsection{Instances of the proposed framework}
\label{subsec:instances}

\begingroup
\setlength{\tabcolsep}{4pt} 
\begin{table}[t!]
    \caption {Implementations of the proposed framework.}
    \label{tab:models}
    \centering
    \begin{tabular}{l*{4}c}
    \toprule
    \multirow{2}{*}{Model name} & \multirow{2}{*}{$p_{\theta}(\vect{z}_{k+1}|\vect{z}_{k})$}  &
    \multirow{2}{*}{$q_{\phi}(\vect{z}_{k}|\vect{z}_{k}^f,\vectall{x})$} & Objective  &  \multirow{2}{*}{Optimiser} \\
    & & & function & \\
    \midrule 
    \midrule 
    BINN\_EnKS & BiNN & EnKS & Eq. \eqref{eq:loss_determ} & EM \\
    \midrule 
    DAODEN\_determ & BiNN & LSTM-AE & Eq. \eqref{eq:loss_determ} & GD \\
    \midrule 
    DAODEN\_MAP & BiNN & LSTM-AE & Eq. \eqref{eq:loss_MAP} & GD \\
    \midrule 
    DAODEN\_full & BiNN & LSTM-VAE & Eq. \eqref{eq:loss_elbo} & GD \\
    \bottomrule
    
    \end{tabular}
    \normalsize
\end{table}
\endgroup
We synthesise in Table. \ref{tab:models} the different configurations of the proposed framework that we implemented in our numerical experiments. We may point out that BiNN\_EnKS configuration is similar to \cite{bocquet_bayesian_2020}.
All configurations use a BiNN with a fourth-order Runge-Kutta scheme to parameterise $\mathcal{F}^{n}$. As presented above, other architectures can also be used to parameterise $\mathcal{F}^{n}$, we choose BiNN to highlight the performance of learning dynamical systems with and without inference schemes (by comparing the performance of BiNN and that of models following the proposed framework). The hyper-parameters of each model are presented in the Appendices. We provide the code to reproduce the result presented in this paper at: https://github.com/CIA-Oceanix/DAODEN. Interested users are highly encouraged to try those models on different dynamical systems or to replace the dynamical sub-module by different learning methods to see the improvement of its performance on noisy and partial observations.

In this paper, unless specified otherwise $n$-step-ahead\_max parameter was set to 4 for DAODEN models and 1 for baseline models (1-step-ahead is the default setting in the original papers of those methods). As in \cite{bocquet_bayesian_2020}, for BiNN\_EnKS, we suppose that we know $\mathbf{R}$. However, for DAODEN, we do not need the exact value of $\mathbf{R}$, when using a fixed value of $\mathbf{R}$ that was from 1 to 2 times larger than the true value of $\mathbf{R}$, the results were similar.   

\subsection{Evaluation metrics}
\label{subsec:evaluation_metrics}

We evaluate both the short-term and long-term performance of the learnt models using the following metrics:
\begin{itemize}
    \item The Root Mean Square Error (RMSE) of the short-term forecast at $t_n = t_0+n \times \delta$: 
    \begin{equation}
        \label{eq:e_n}
        e_n = \sqrt{\frac{1}{n}\sum^n_{k=1}(\vect{z}^{pred}_{k} - \vect{z}^{true}_{k} )^2}
    \end{equation}
    with $\vect{z}^{pred}_{k} \overset{\Delta}{=} \mathcal{F}^k(\vect{z}_{0})$ and $\vect{z}_0$ is the first state of each sequence in the test set. 

    \item The reconstruction capacity given the observations, denoted as $rec$:
    \begin{equation}
        \label{eq:r}
        rec = \sqrt{\frac{1}{T}\sum_{k=0}^{T}(\vect{z}^{*}_{k} - \vect{z}^{true}_{k} )^2}
    \end{equation}
    with $\vect{z}_k^* =  \mathbb{E}\left[q_\vphi(\vect{z}_k|\vect{z}_k^{f},\vectall{x})\right]$.
    
    \item The first time (in Lyapunov unit) the RMSE reaches half of the standard deviation of the true system, denoted as $\pi_{0.5}$.
    
    \item The capacity to maintain the long-term topology of the system, evaluated via the first Lyapunov exponent $\lambda_1$ calculated in a forecasting sequence of length of 20000 time steps, using the method presented in \cite{wolf_determining_1985}. The true $\lambda_1$ of the L63 is 0.91 and the true $\lambda_1$ of the L96 is 1.67.
\end{itemize}
For each metric, we compute the average of the results on 50 sequences in the test set. 

As Lorenz dynamics may be interpreted in terms of geophysical dynamics, we may also give some physical interpretation to the considered metrics. 
For example, in geosciences, for experiments on the L96 system with $\delta$=0.05 (correspond to 6 hours in real-world time), $e_4$ would relate to the precision of a weather forecast model for the next day, $\pi_{0.5}$ indicates how long the forecast is still meaningful, $\lambda_1$ indicates whether a model can be used for long-term forecast such as the simulation of climate change, and $rec$ indicates the ability of a model to reconstruct the true states of a system when the observations are noisy and partial, such as reconstructing the sea surface condition from satellite images. 

\subsection{L63 case-study}
\label{subsec:lorenz63_noisy}

In this section we report the results for the L63 case-study. We first assess the identification performance on noisy but complete observations (\ie $\vphi_k$ is an identity matrix at all time steps) of the L63 system, then address cases where the observations are sampled partially, both in time and in space.

\begin{figure}
    \centering
    \includegraphics[width=\linewidth, clip, trim=0mm 0mm 0mm 0mm]{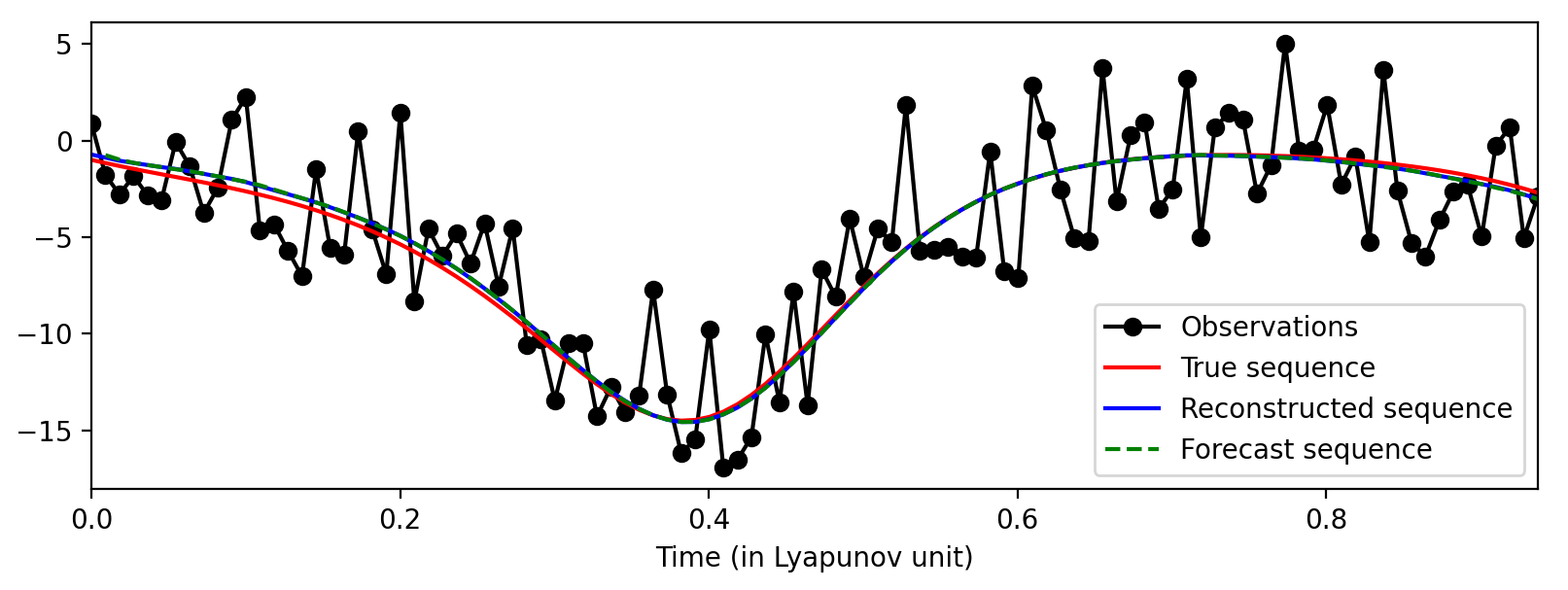}
    \caption{An example of the the first dimension of the L63 system reconstructed by the inference module of DAODEN\_determ, $r=33\%$. Given the noisy observations (black dots), inference module $q_\vphi(\vect{z}_k|\vect{z}^f_k,\vectall{x})$ reconstructs a clean sequence of the hidden state (blue curve), which is very close to the true unknown dynamic (red curve). Given this sequence, the transition network (BiNN) can successfully learn the governing laws of the system, as it can do under ideal conditions. The green dash shows the forecast $\vect{z}^*_{k+1} = \mathcal{F}^1(\vect{z}^*_{k})$ given the mean $\vect{z}^*_{k}$ of $q_\vphi$.}
    \label{fig:smoother}
\end{figure}

Table \ref{tab:l63_noisy} shows the performance of the considered models on noisy L63 data. We compare the performance of the 4 proposed models with the baselines' w.r.t the short-term prediction error and the capacity to maintain the long-term topology. All the models based on the proposed framework outperform the baselines by a large margin. This asserts the ability of the proposed framework to deal with noisy observations. In Fig. \ref{fig:smoother} we show the first component of a L63 sequence in the test set reconstructed by the inference scheme of DAODEN\_determ. $q_\vphi$ is expected to infer a mapping that converts data from the corrupted observation space (black dots) to the true space of the dynamics (the red curve). In this space, data-driven methods can successfully learn the governing equations of the system. The reconstructed sequence is very close to the true sequence. 

At first glance, we can see that no model is better than all the others in all 4 criteria. This is aligned with the finding of \cite{fablet_joint_2020}. BiNN\_EnKS and DAODEN\_full have very good forecasting scores, however, the performance of BiNN\_EnKS in reconstructing the true states is not as good as DAODEN models. The dynamics learnt by DAODEN models are also more synchronised to the true dynamics (indicated by $\pi_{0.5}$) than those learnt by BiNN\_EnKS. This might suggest that NN-based models (here are LSTM-AE and LSTM-VAE) can be an alternative for classic inference schemes like EnKS, which are among the state-of-the-art methods in data assimilation \cite{lahoz_data_2010}.  

In Fig. \ref{fig:attactors_lorenz63_noisy}, we show the attractors generated by the learnt models. AnDA is more suitable for data assimilation than for forecasting. When the noise level is small ($r$=8.5\% and $r$=16.7\%), SINDy and BiNN can still capture the dynamics of the system. When the noise level is significant ($r$=33.3\% and $r$=66.7\%), the attactors generated by SINDy and BiNN are distorted, which indicates that the learnt models are not valid for long-term simulations. On the other hand, all the models of the proposed framework successfully reconstructed the butterfly topology of the attractor, even when the noise level is high.

{}

 \begingroup
\begin{table*}
    \caption {Performance of models trained on noisy L63 data. For each index, the best score is marked in \textbf{bold} and the second best score is marked in \textit{italic}.}
    \label{tab:l63_noisy}
    \footnotesize
    \centering
    \begin{tabular}{ll*{4}c}
    \toprule
    \multicolumn{2}{c}{\multirow{2}{*}{Model}} & \multicolumn{4}{c}{$r$} \\
    & & 8.5\% & 16.7\% & 33.3\% & 66.7\%\\
    \midrule 
    
    \midrule 
    \multirow{4}{*}{AnDA}
    & $e_4$      & 0.351±0.184  & 0.777±0.350 & 1.683±0.724 & 3.682±1.346 \\
    & $rec$      & 0.416±0.019  & 0.941±0.037 & 2.134±0.076 & 4.876±0.168 \\
    & $\pi_{0.5}$      & 0.820±0.480  & 0.380±0.172 & 0.249±0.174 & 0.104±0.116 \\
    &  $\lambda_1$  &  26.517±7.665 & 27.146±42.927 & 76.267±28.150  & 127.047±0.881 \\

    \midrule 
    \multirow{3}{*}{SINDy}
    & $e_4$      &  0.068±0.052 & 0.149±0.106 & 0.311±0.196 & 0.694±0.441 \\
    & $\pi_{0.5}$      & 0.490±0.261  & 0.165±0.085 & 0.077±0.049  &  0.034±0.034 \\
    &  $\lambda_1$  &  0.898±0.008 & 0.840±0.035 &  0.840±0.035 & nan±nan \\
    
    \midrule 
    \multirow{3}{*}{BiNN}
    & $e_4$      & 0.045±0.030  & 0.119±0.085 & 0.283±0.185  & 0.684±0.408\\
    & $\pi_{0.5}$      & 3.608±1.364  & 2.053±0.666 &  0.975±0.488 & 0.308±0.125\\
    &  $\lambda_1$  & 0.900±0.011  & 0.868±0.010 & 0.122±0.208  & -0.422±0.047\\
    
    \midrule 
    \multirow{3}{*}{Latent-ODE}
    & $e_4$      & 0.051±0.027  & 0.062±0.034 & \textit{0.065±0.042}  & \textit{0.213±0.084}\\
    & $\pi_{0.5}$      & 2.504±1.332  & 2.336±1.472 &  2.852±1.352 & 2.118±1.129\\
    &  $\lambda_1$  & 0.892±0.018  & 0.877±0.018 & 0.885±0.015  & 0.675±0.027\\    
    
    \midrule 
    \multirow{4}{*}{BiNN\_EnKS}
    & $e_4$      & \textbf{0.019±0.016}  & \textbf{0.024±0.023} & \textbf{0.037±0.024}  & 0.276±0.160 \\
    & $rec$      &  0.323±0.024 & 0.431±0.042 & 0.598±0.093  & 1.531±0.332\\
    & $\pi_{0.5}$      & 2.807±1.128  & 3.004±1.355 &  2.996±1.641 & \textit{2.081±1.214}\\
    &  $\lambda_1$  &  0.856±0.031 & 0.869±0.024 & 0.826±0.065  & 0.868±0.014 \\

    \midrule 
    \multirow{4}{*}{DAODEN\_determ}
    & $e_4$      & 0.049±0.031  & 0.056±0.034 &  0.077±0.048 & 0.268±0.201\\
    & $rec$      & 0.216±0.125  & 0.269±0.110 &  \textbf{0.448±0.199} & \textit{0.873±0.216}\\
    & $\pi_{0.5}$      & \textit{3.519±1.282}  & \textit{3.488±1.327} & \textbf{3.470±1.562}  &  1.803±1.104\\
    &  $\lambda_1$  & 0.882±0.036 &  0.895±0.021 & 0.911±0.013 & 0.793±0.021\\
    
    \midrule 
    \multirow{4}{*}{DAODEN\_MAP}
    & $e_4$      & 0.038±0.027  & 0.038±0.038 &  0.101±0.070 & 0.233±0.088 \\
    & $rec$      & \textit{0.209±0.096}  & \textbf{0.234±0.065} &  0.525±0.253 & \textbf{0.817±0.330}\\
    & $\pi_{0.5}$      & 3.271±1.270  & 3.219±1.260 &  2.993±1.413 & \textbf{2.650±1.382}\\
    &  $\lambda_1$  & 0.860±0.047 & 0.876±0.029 & 0.916±0.012 & 0.920±0.008\\
    
    \midrule 
    \multirow{4}{*}{DAODEN\_full}
    & $e_4$      & \textit{0.023±0.015}  & \textit{0.027±0.016} &  0.072±0.045 & \textbf{0.187±0.127}\\
    & $rec$      & \textbf{0.178±0.050}  & \textit{0.258±0.066} &  \textit{0.469±0.168} & 1.003±0.380\\
    & $\pi_{0.5}$      & \textbf{3.533±1.139}  & \textbf{3.496±1.215} & \textit{3.426±1.512}  & 1.897±0.918\\
    &  $\lambda_1$  & 0.869±0.036 & 0.858±0.028 & 0.881±0.024 & 0.884±0.013\\

    
    
    \bottomrule
    \end{tabular}
    \normalsize
\end{table*}
\endgroup

\begingroup
\newcommand{\ltrim}{100mm}%
\newcommand{\btrim}{40mm}%
\newcommand{\rtrim}{100mm}%
\newcommand{\ttrim}{40mm}%
\newcommand{\nfwidth}{0.8\linewidth}%
\newcommand{\swidth}{0.05\linewidth}%
\newcommand{\bwidth}{0.235\linewidth}%
\begin{figure}[t!]
    \centering
	\begin{subfigure}[t]{0.04\linewidth}
		\hfill
		\caption*{}
	\end{subfigure}%
	\begin{subfigure}[t]{0.96\linewidth}
		\hspace{2.5mm} $r=8.5\%$ \hspace{\swidth} $r=16.7\%$ \hspace{\swidth} $r=33.3\%$ \hspace{\swidth} $r=66.7\%$ \hfill
	\end{subfigure}%
	\vspace{-2mm}
	\begin{subfigure}[b]{0.04\linewidth}
	    \rotatebox[origin=t]{90}{\scriptsize AnDA}\vspace{0.79\linewidth}
	\end{subfigure}%
	\begin{subfigure}[t]{0.96\linewidth}
		\centering
		\includegraphics[width=\bwidth,clip, trim=100mm 40mm 80mm 40mm]{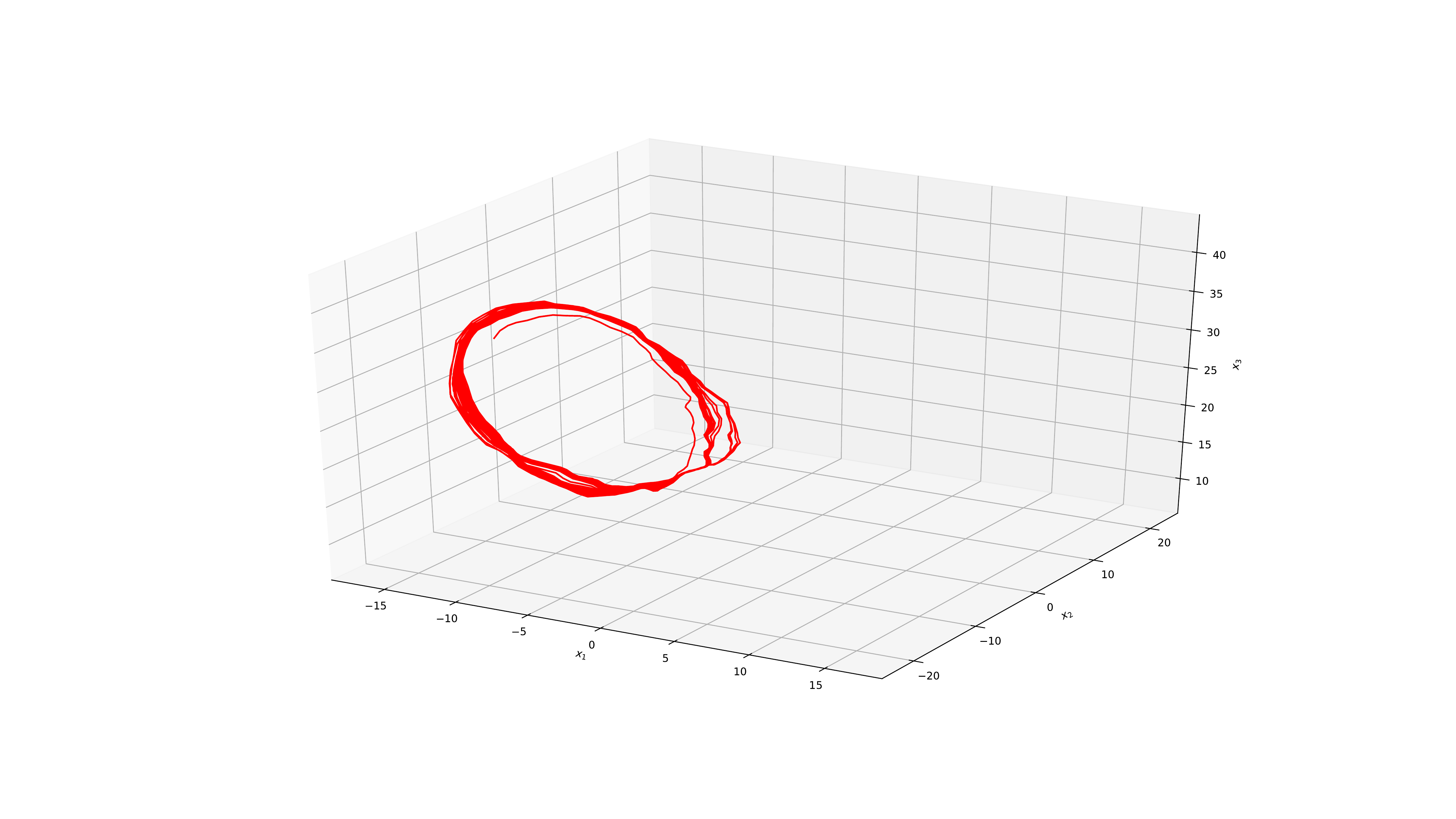}
		\includegraphics[width=\bwidth,clip, trim=100mm 40mm 80mm 40mm]{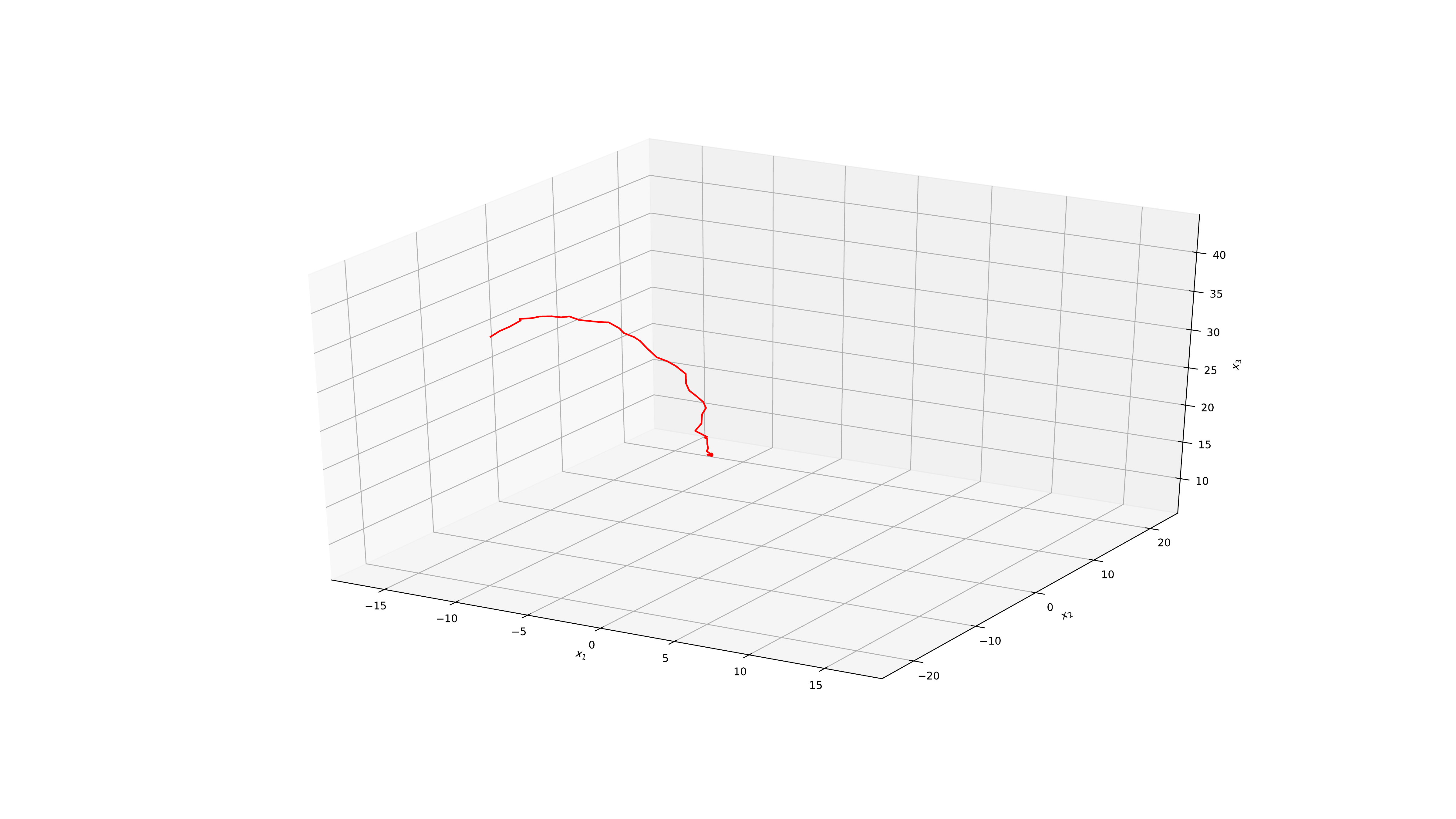}
		\includegraphics[width=\bwidth,clip, trim=100mm 40mm 80mm 40mm]{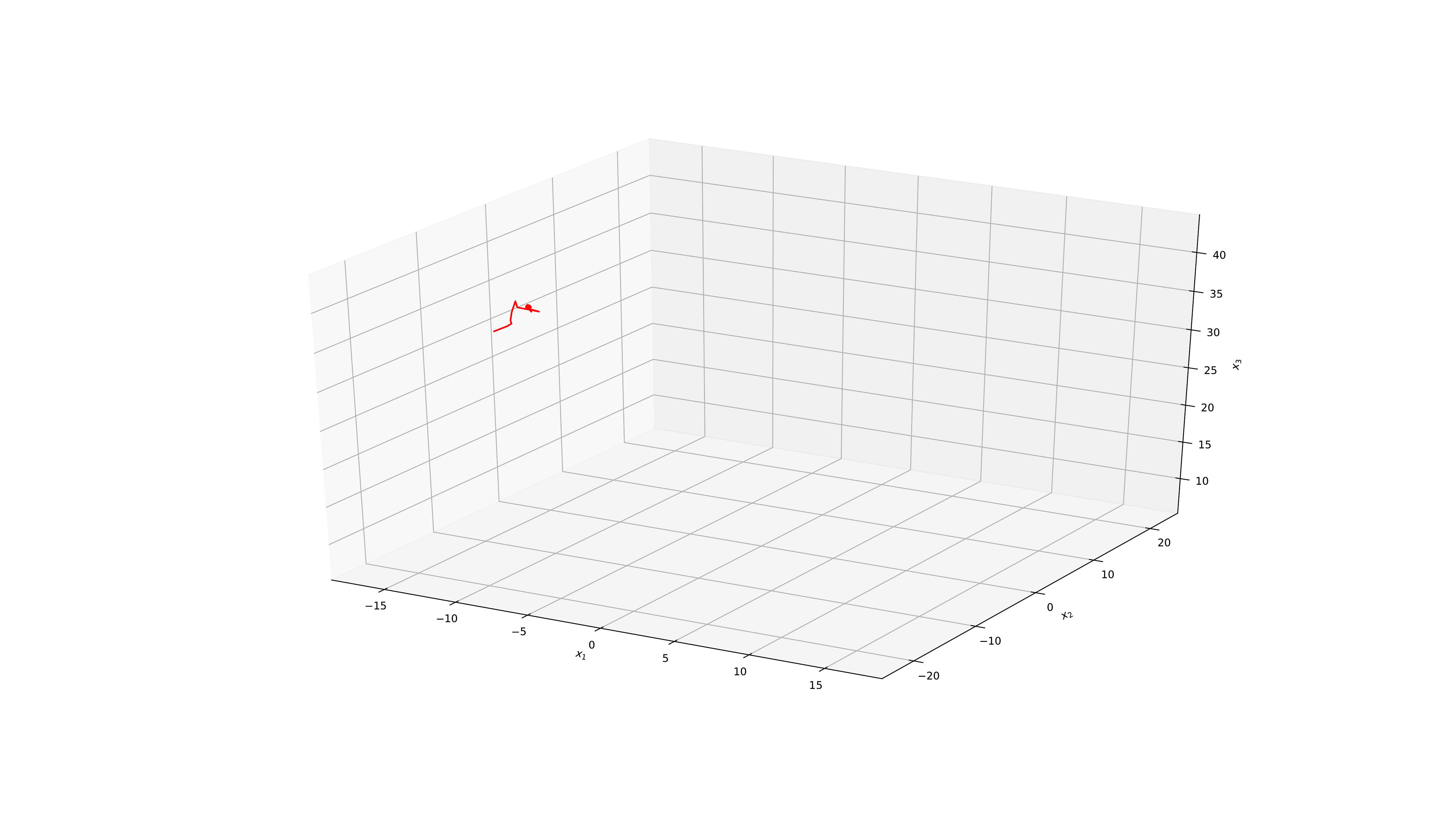}
		\includegraphics[width=\bwidth,clip, trim=100mm 40mm 80mm 40mm]{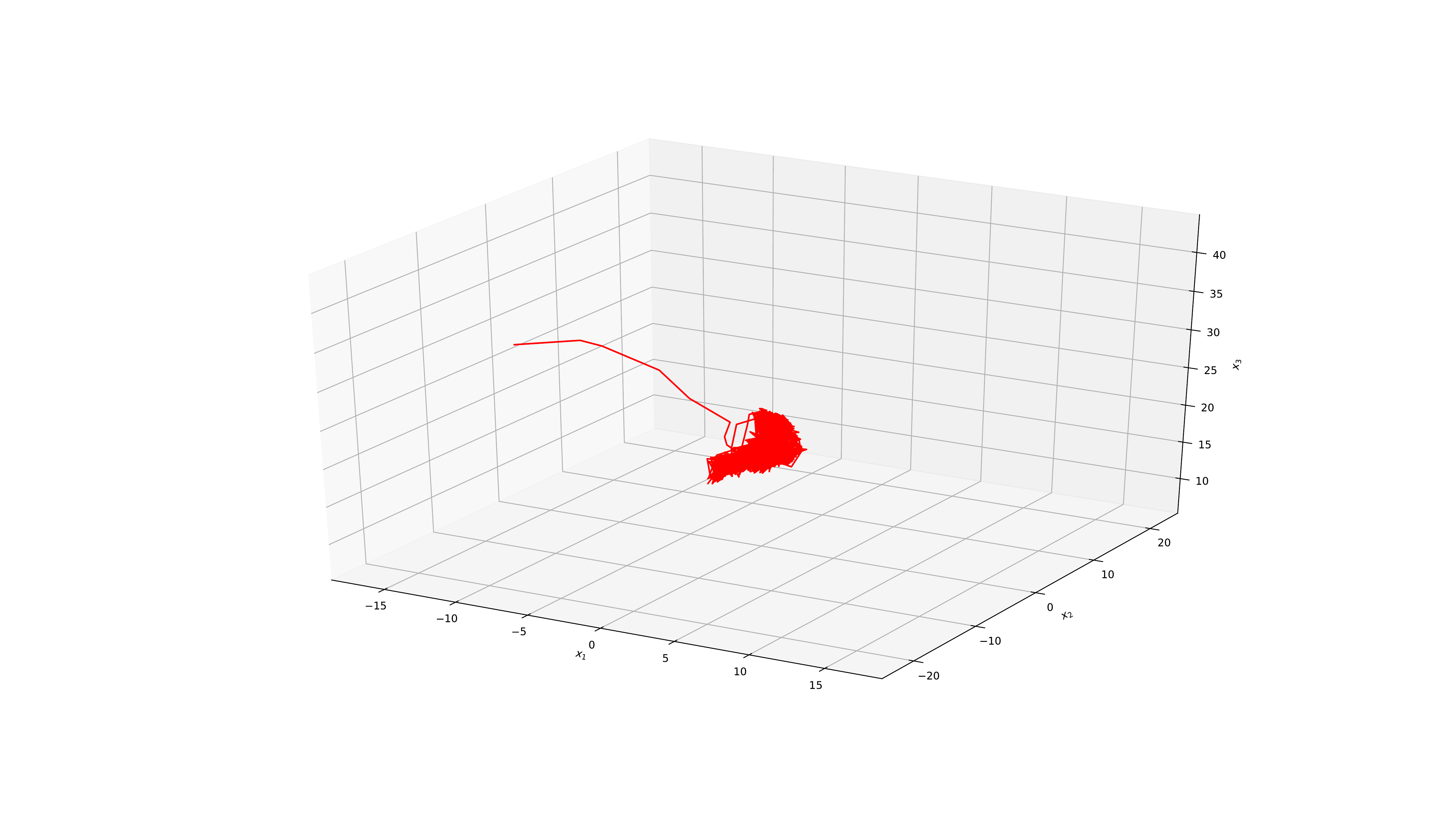}
	\end{subfigure}%
	
	\begin{subfigure}[b]{0.04\linewidth}
	    \rotatebox[origin=t]{90}{\scriptsize SINDy}\vspace{0.9\linewidth}
	\end{subfigure}%
	\begin{subfigure}[t]{0.96\linewidth}
		\centering
		\includegraphics[width=\bwidth,clip, trim=100mm 40mm 80mm 40mm]{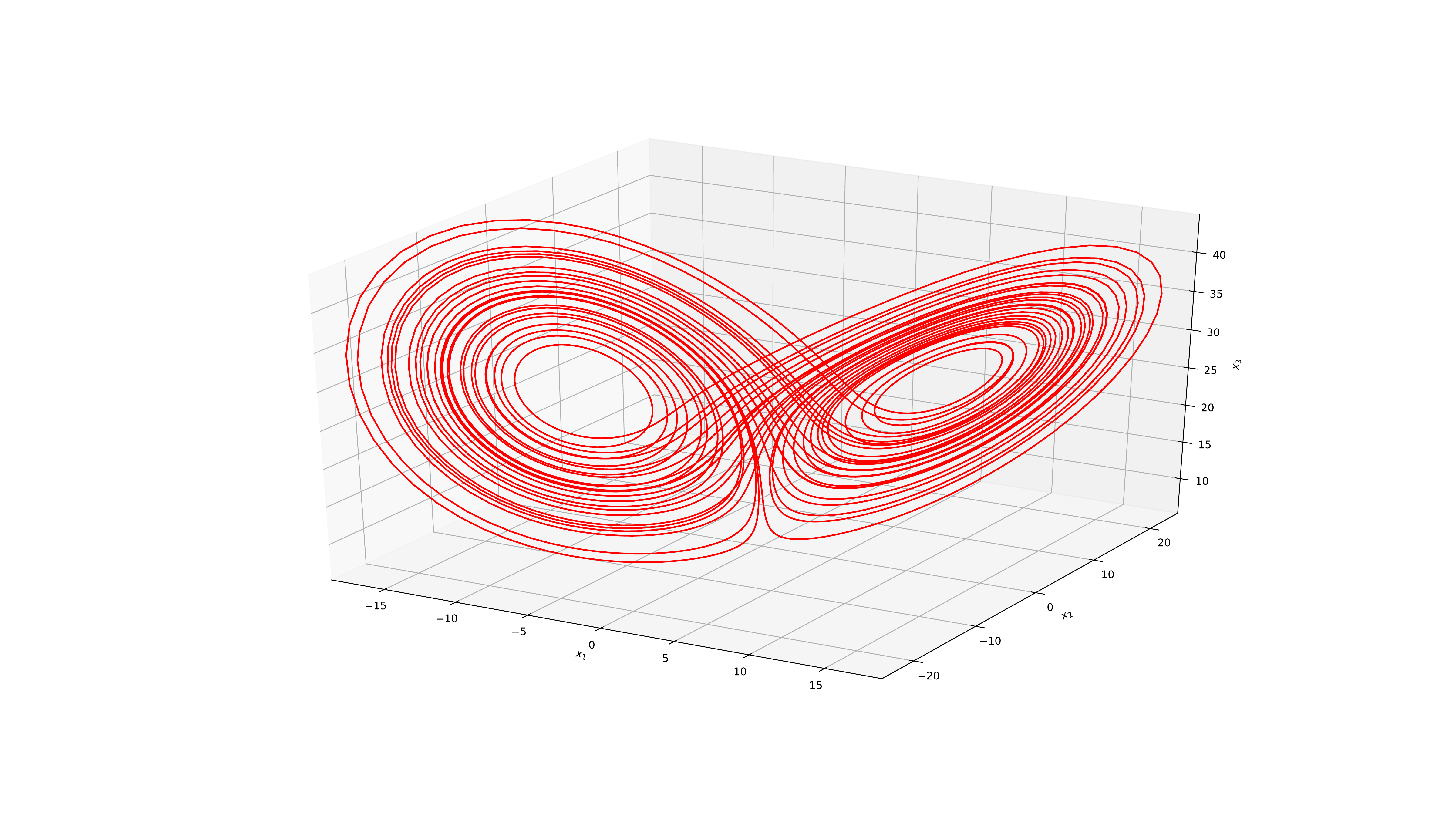}
		\includegraphics[width=\bwidth,clip, trim=100mm 40mm 80mm 40mm]{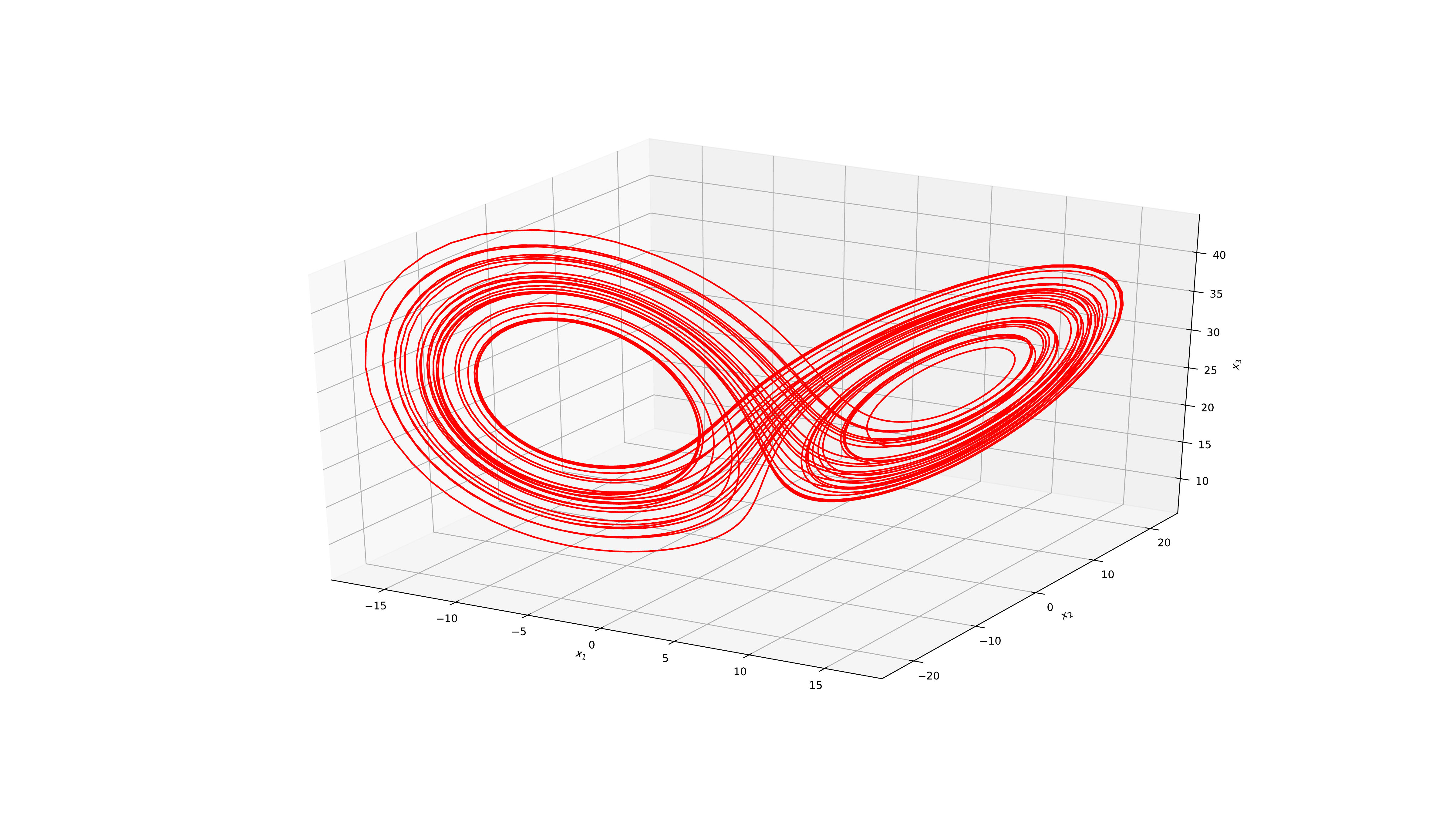}
		\includegraphics[width=\bwidth,clip, trim=100mm 40mm 80mm 40mm]{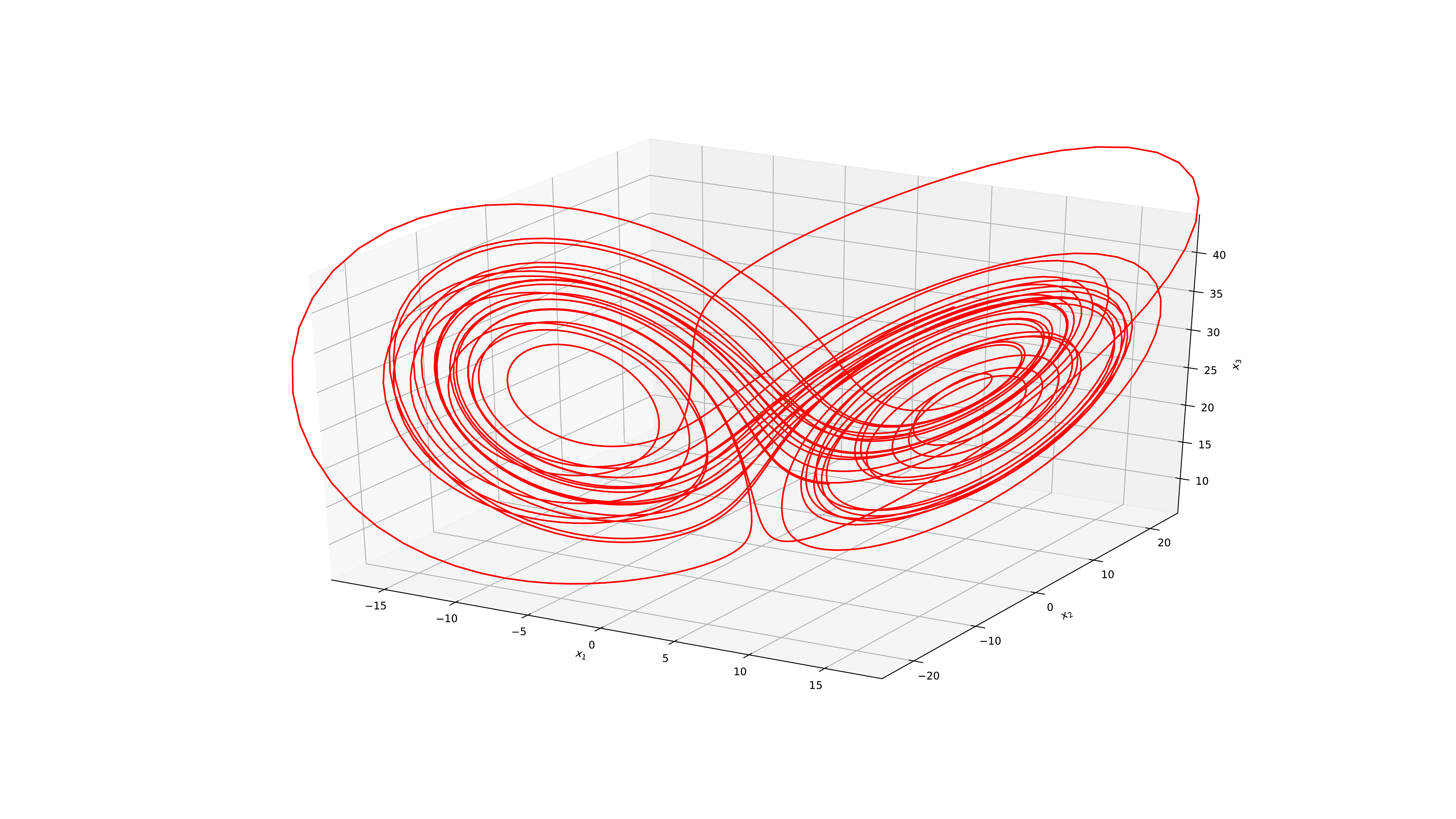}
		\includegraphics[width=\bwidth,clip, trim=100mm 40mm 80mm 40mm]{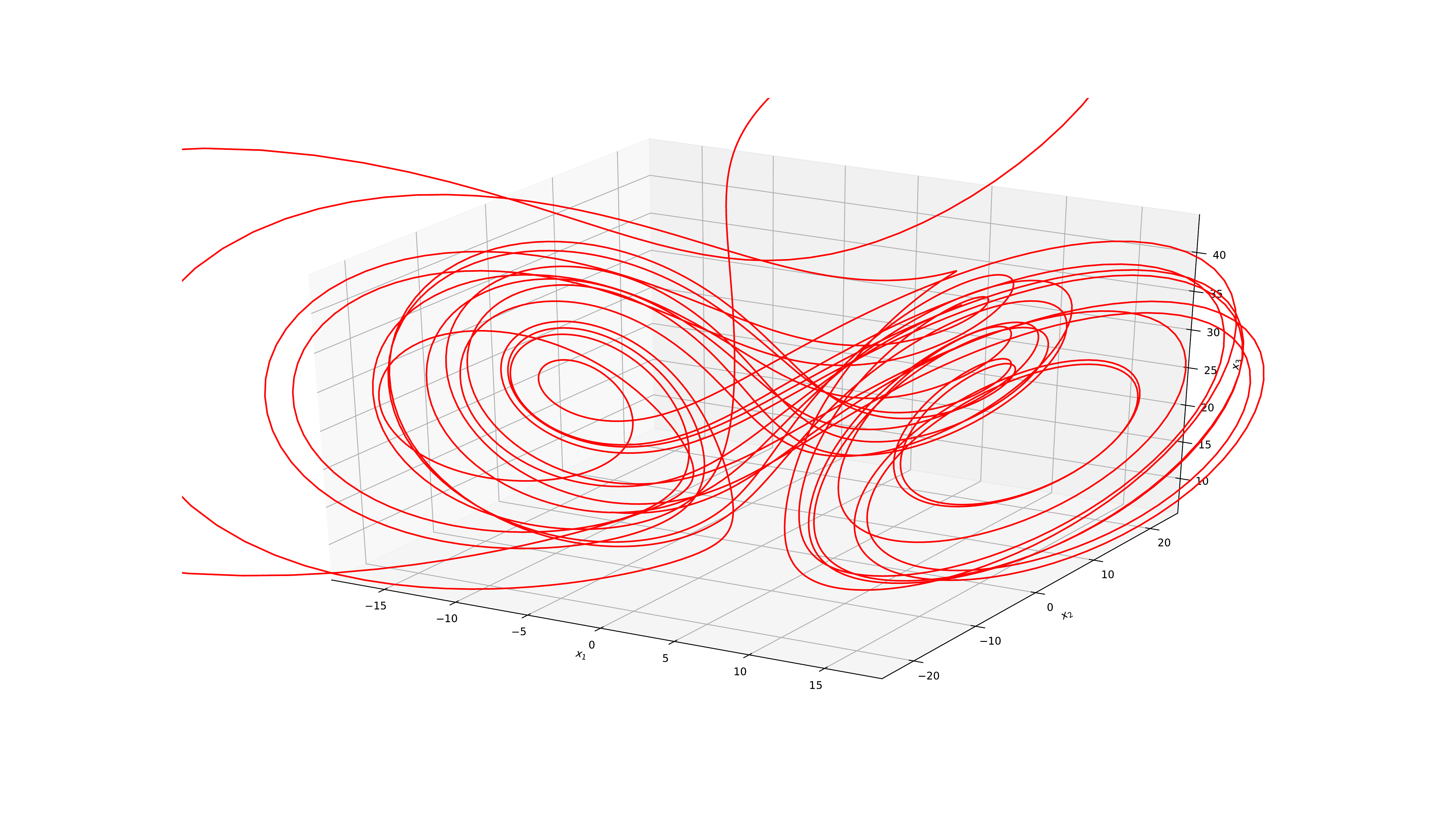}
	\end{subfigure}%

	\begin{subfigure}[b]{0.04\linewidth}
	    \rotatebox[origin=t]{90}{\scriptsize BiNN}\vspace{0.7\linewidth}
	\end{subfigure}%
	\begin{subfigure}[t]{0.96\linewidth}
		\centering
		\includegraphics[width=\bwidth,clip, trim=100mm 40mm 80mm 40mm]{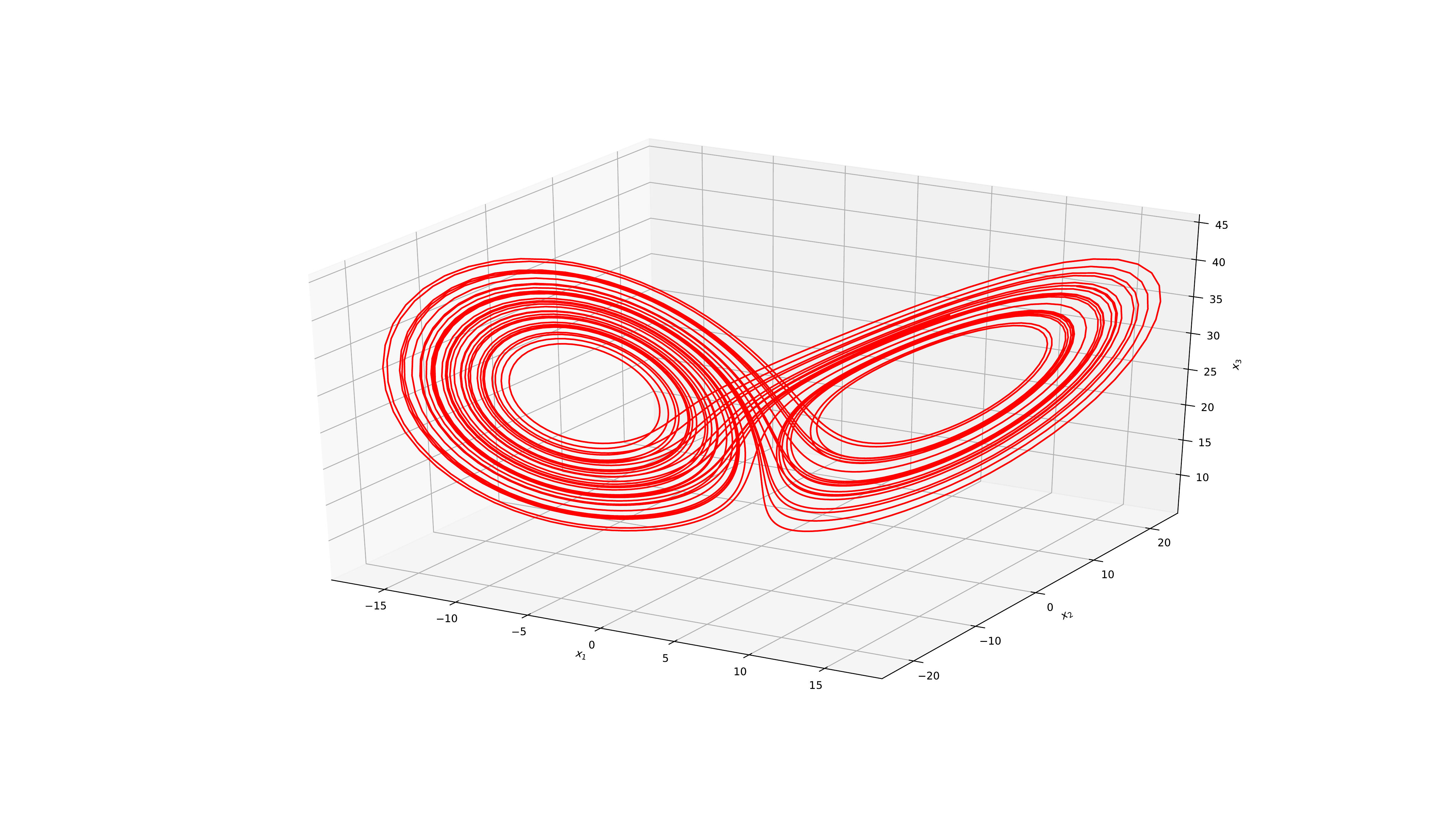}
		\includegraphics[width=\bwidth,clip, trim=100mm 40mm 80mm 40mm]{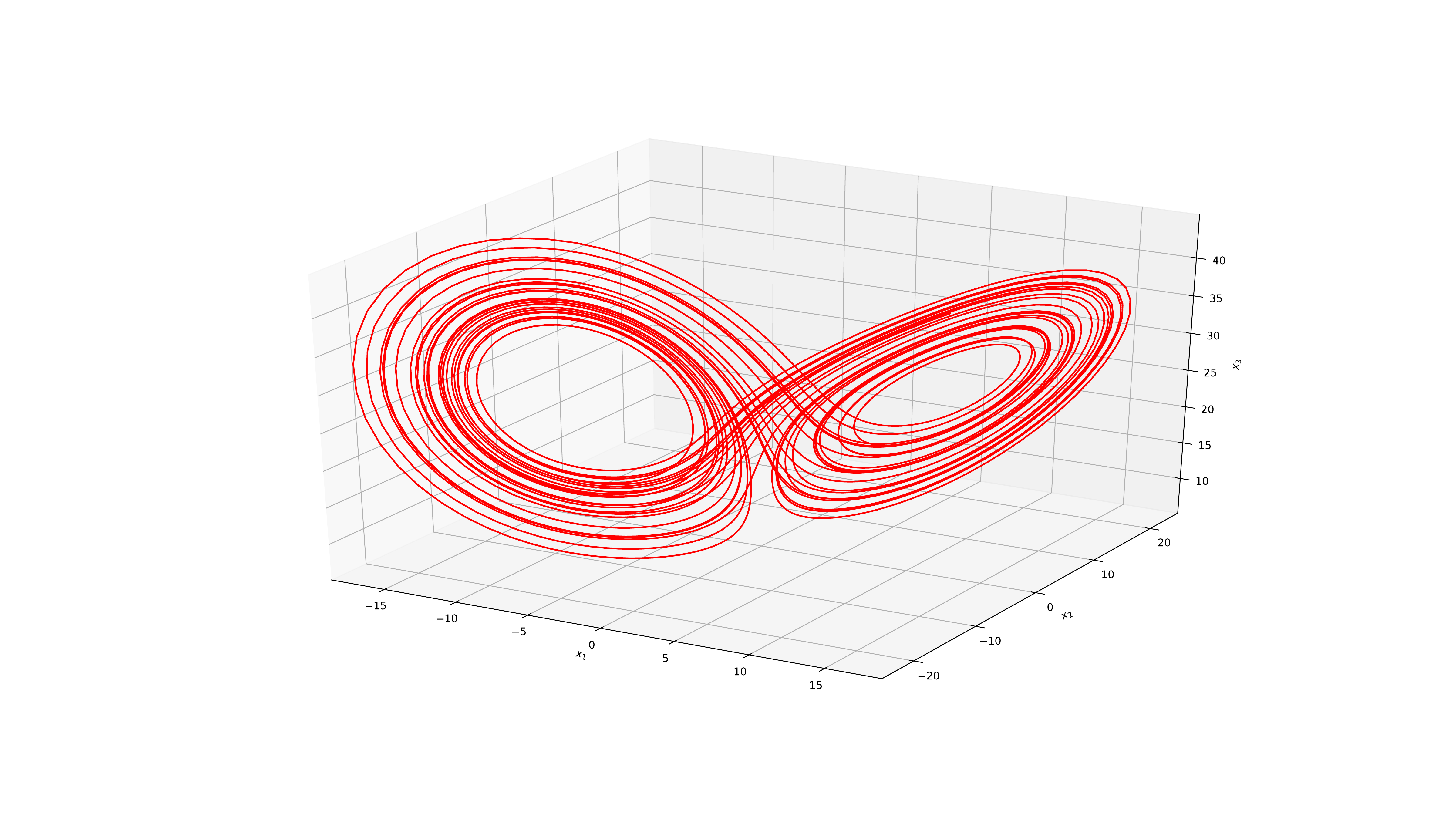}
		\includegraphics[width=\bwidth,clip, trim=100mm 40mm 80mm 40mm]{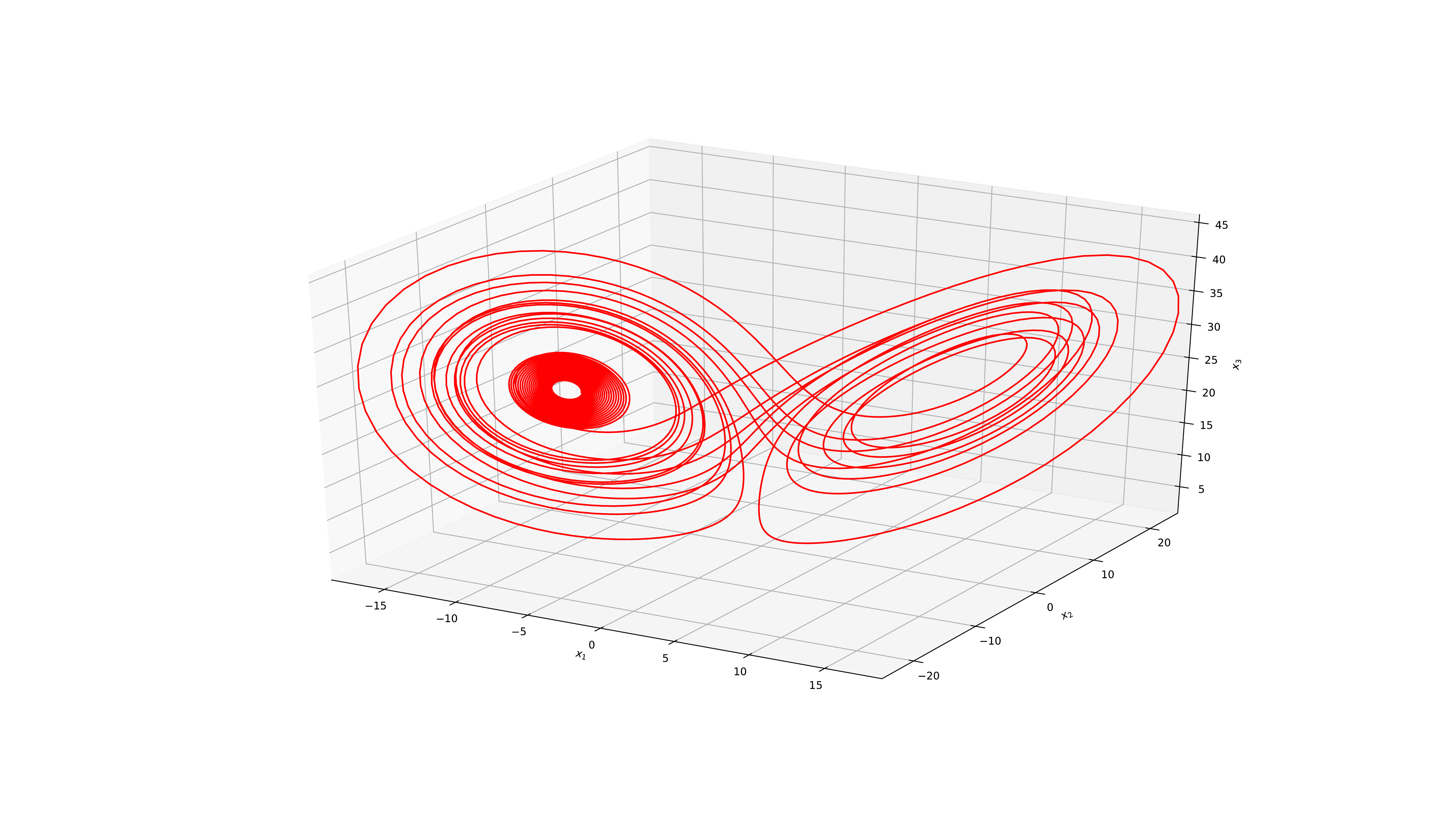}
		\includegraphics[width=\bwidth,clip, trim=100mm 40mm 80mm 40mm]{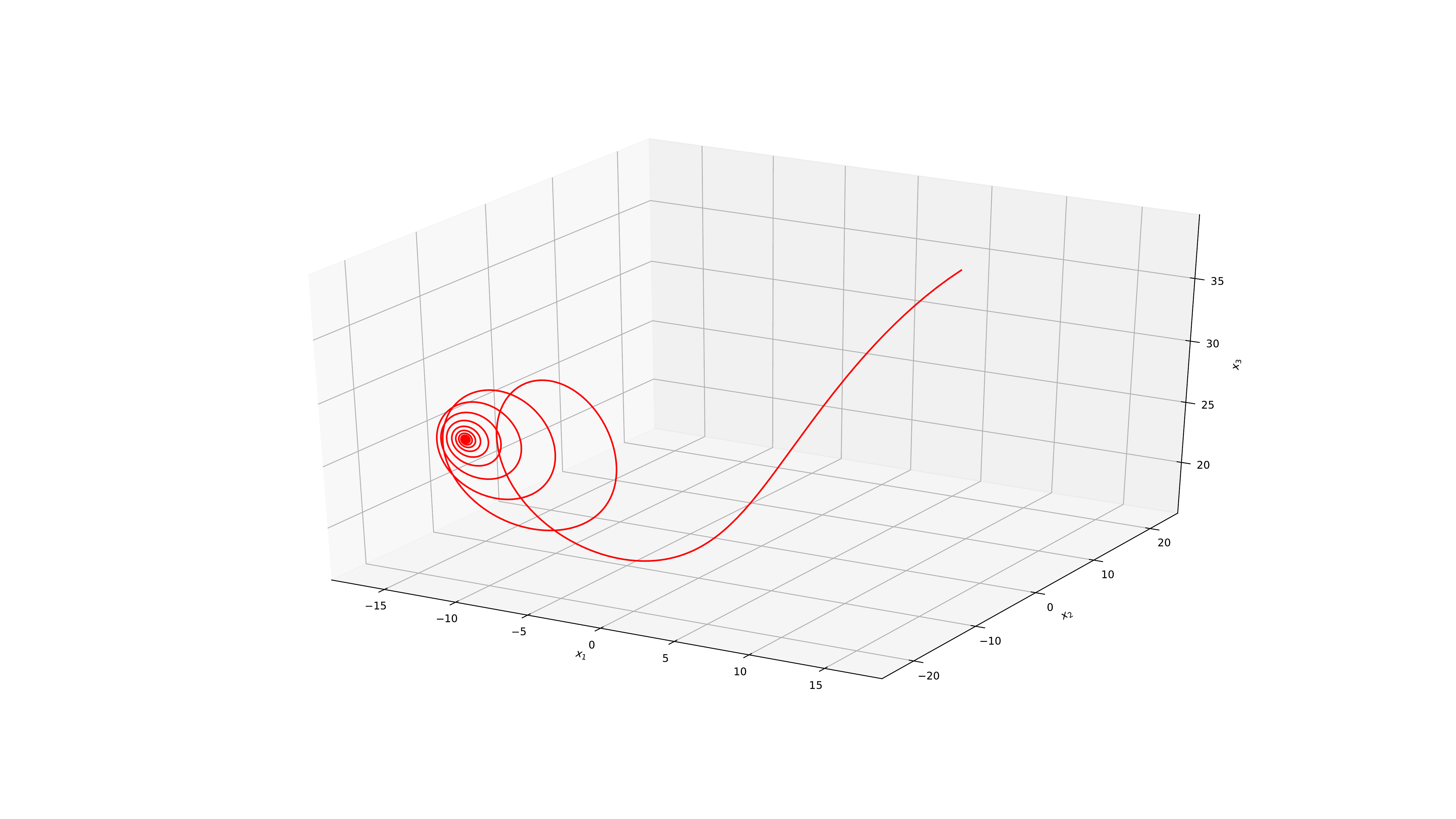}
	\end{subfigure}%

	\begin{subfigure}[b]{0.04\linewidth}
	    \rotatebox[origin=t]{90}{\scriptsize LatentODE}\vspace{0.2\linewidth}
	\end{subfigure}%
	\begin{subfigure}[t]{0.96\linewidth}
		\centering
		\includegraphics[width=\bwidth,clip, trim=100mm 40mm 80mm 40mm]{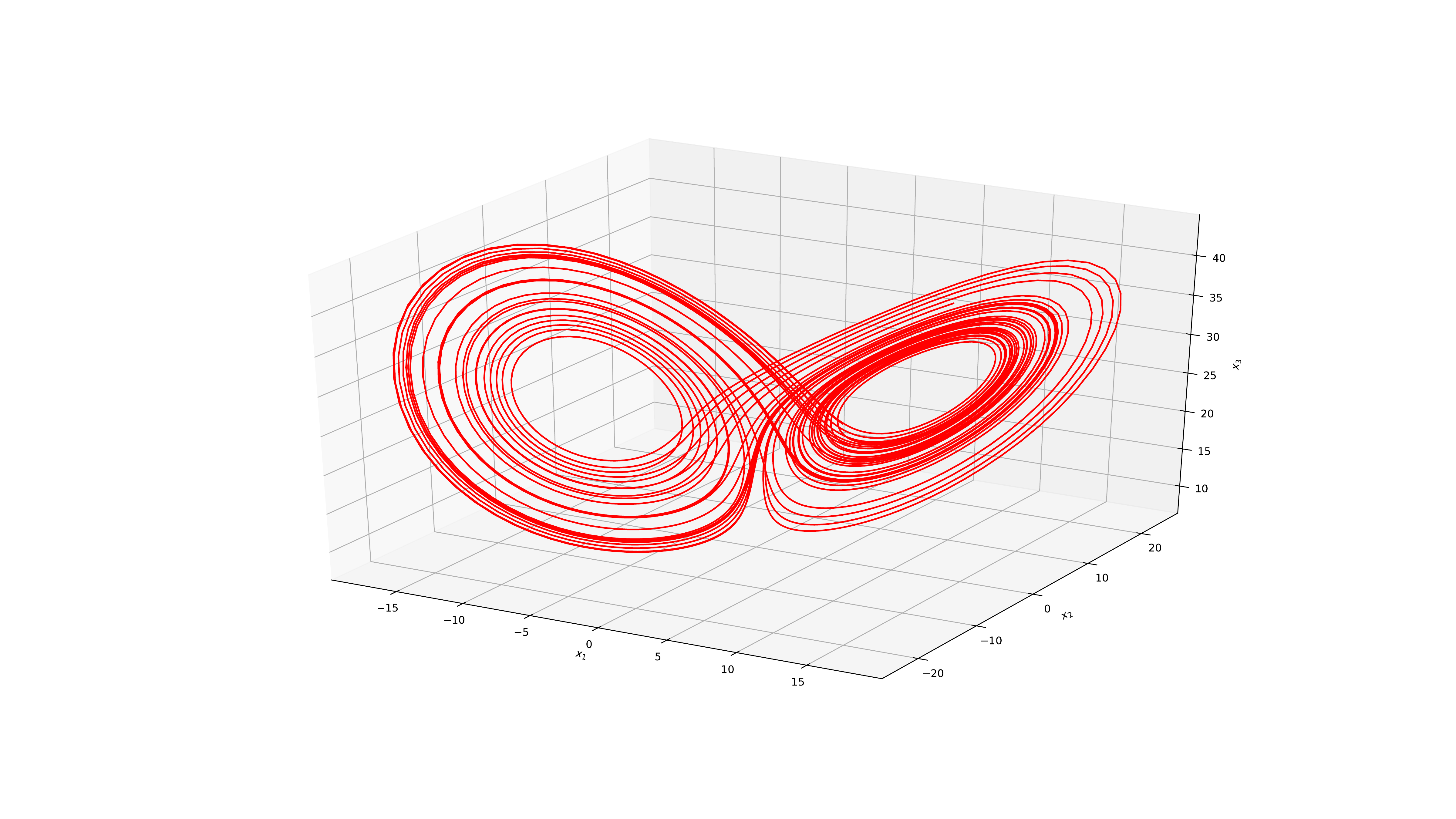}
		\includegraphics[width=\bwidth,clip, trim=100mm 40mm 80mm 40mm]{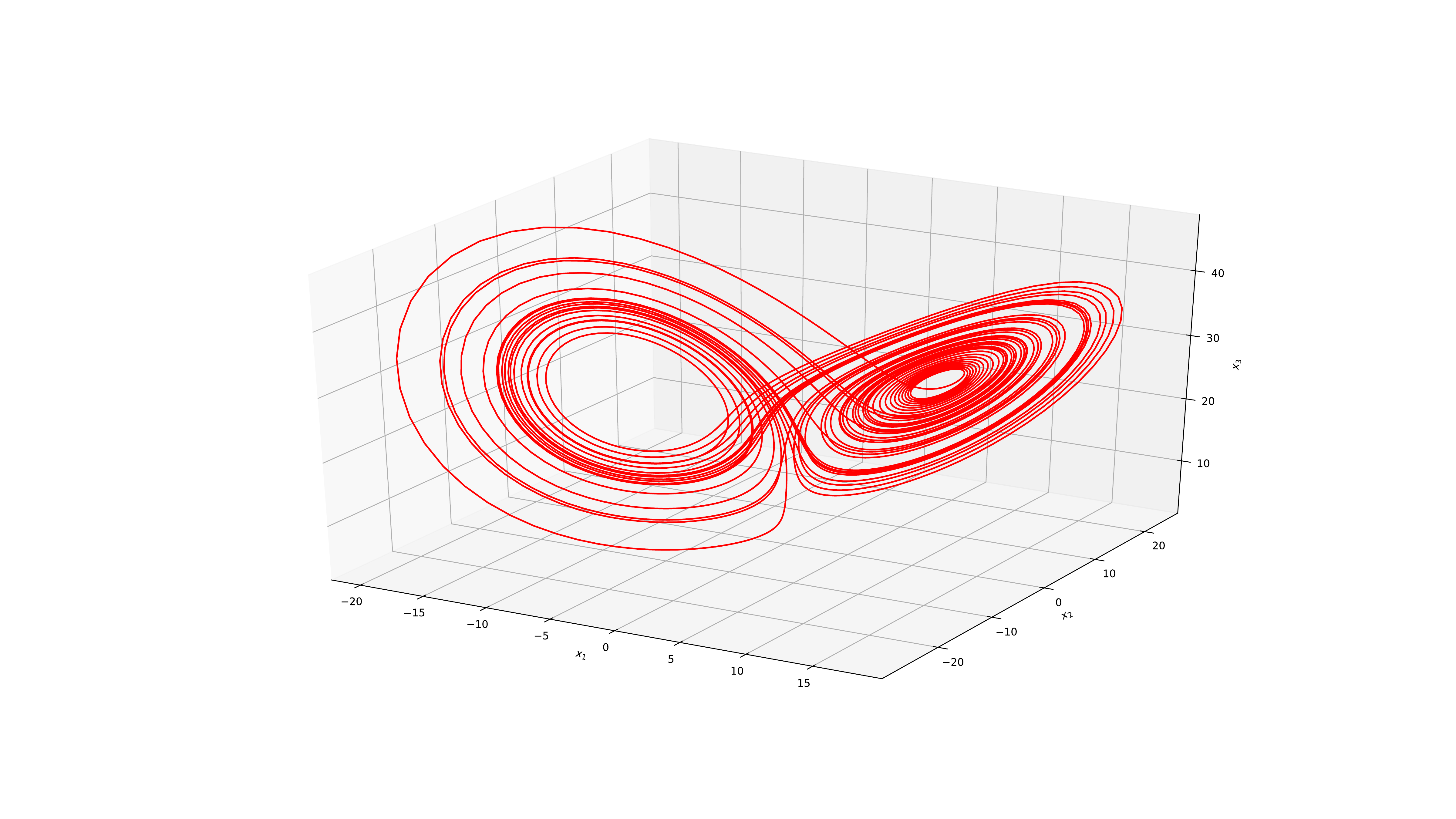}
		\includegraphics[width=\bwidth,clip, trim=100mm 40mm 80mm 40mm]{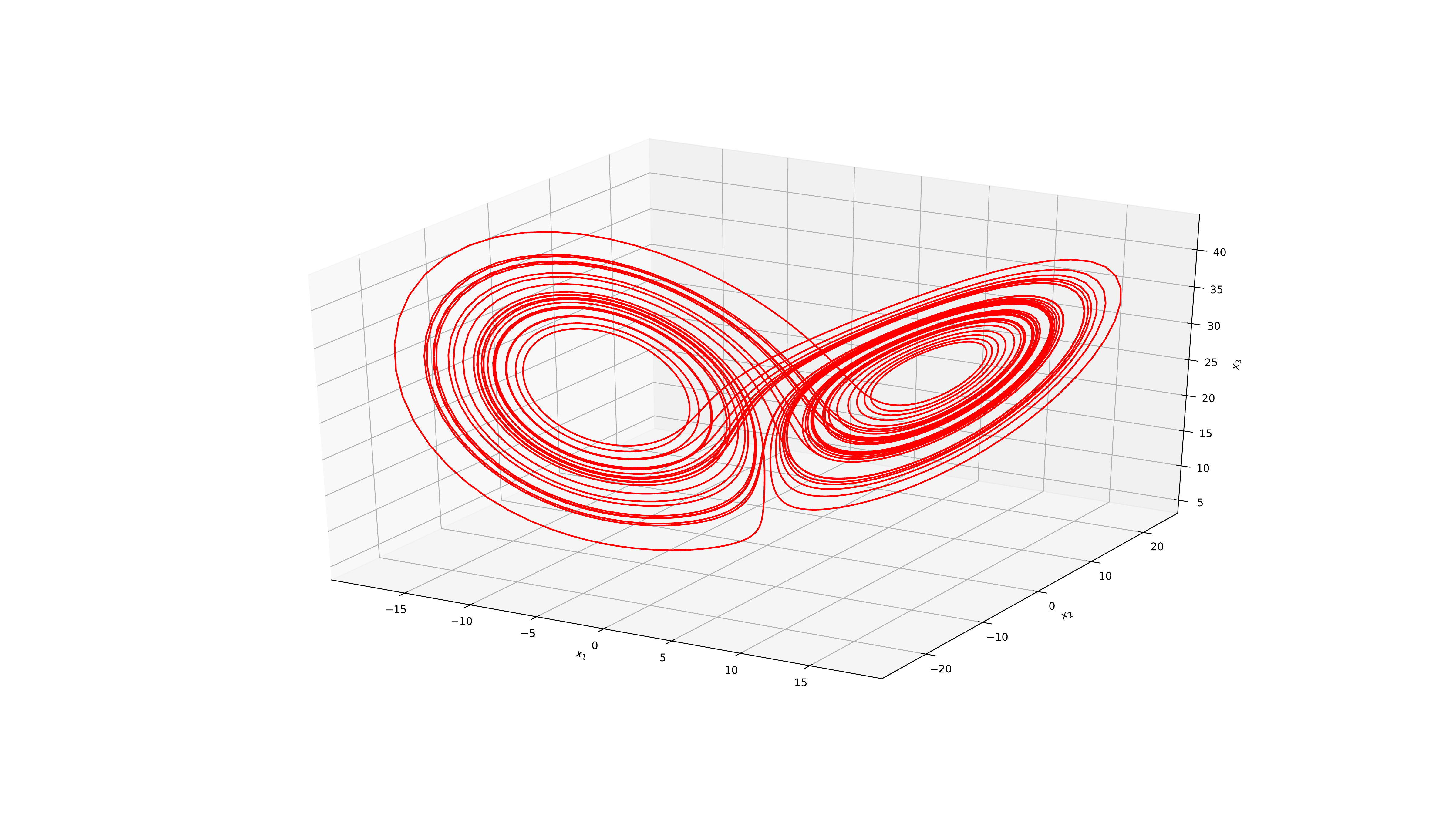}
		\includegraphics[width=\bwidth,clip, trim=100mm 40mm 80mm 40mm]{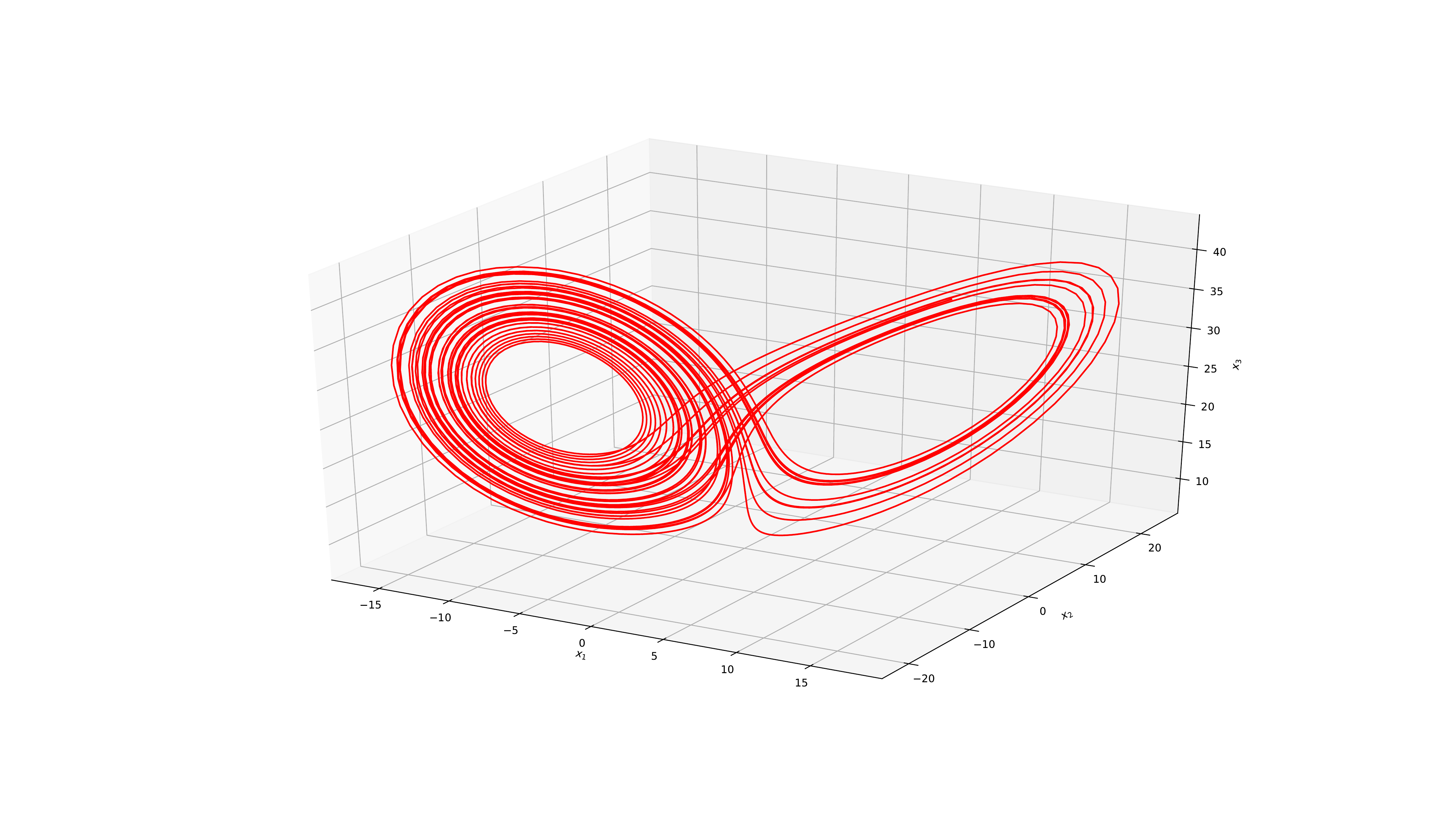}
	\end{subfigure}%
	
	\begin{subfigure}[b]{0.04\linewidth}
	    \rotatebox[origin=t]{90}{\scriptsize     EnKS}\vspace{0.9\linewidth}
	\end{subfigure}%
	\begin{subfigure}[t]{0.96\linewidth}
		\centering
		\includegraphics[width=\bwidth,clip, trim=100mm 40mm 80mm 40mm]{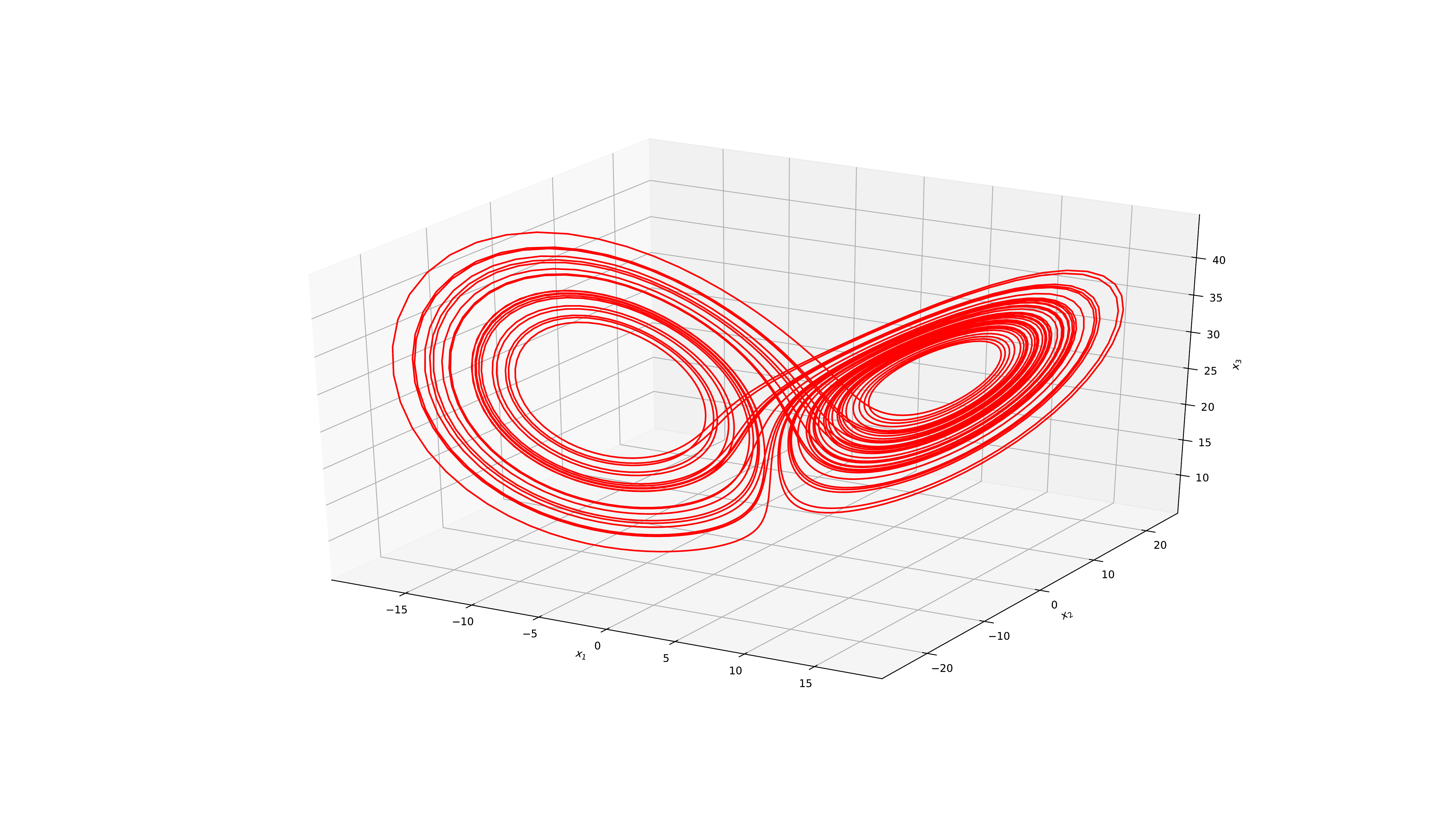}
		\includegraphics[width=\bwidth,clip, trim=100mm 40mm 80mm 40mm]{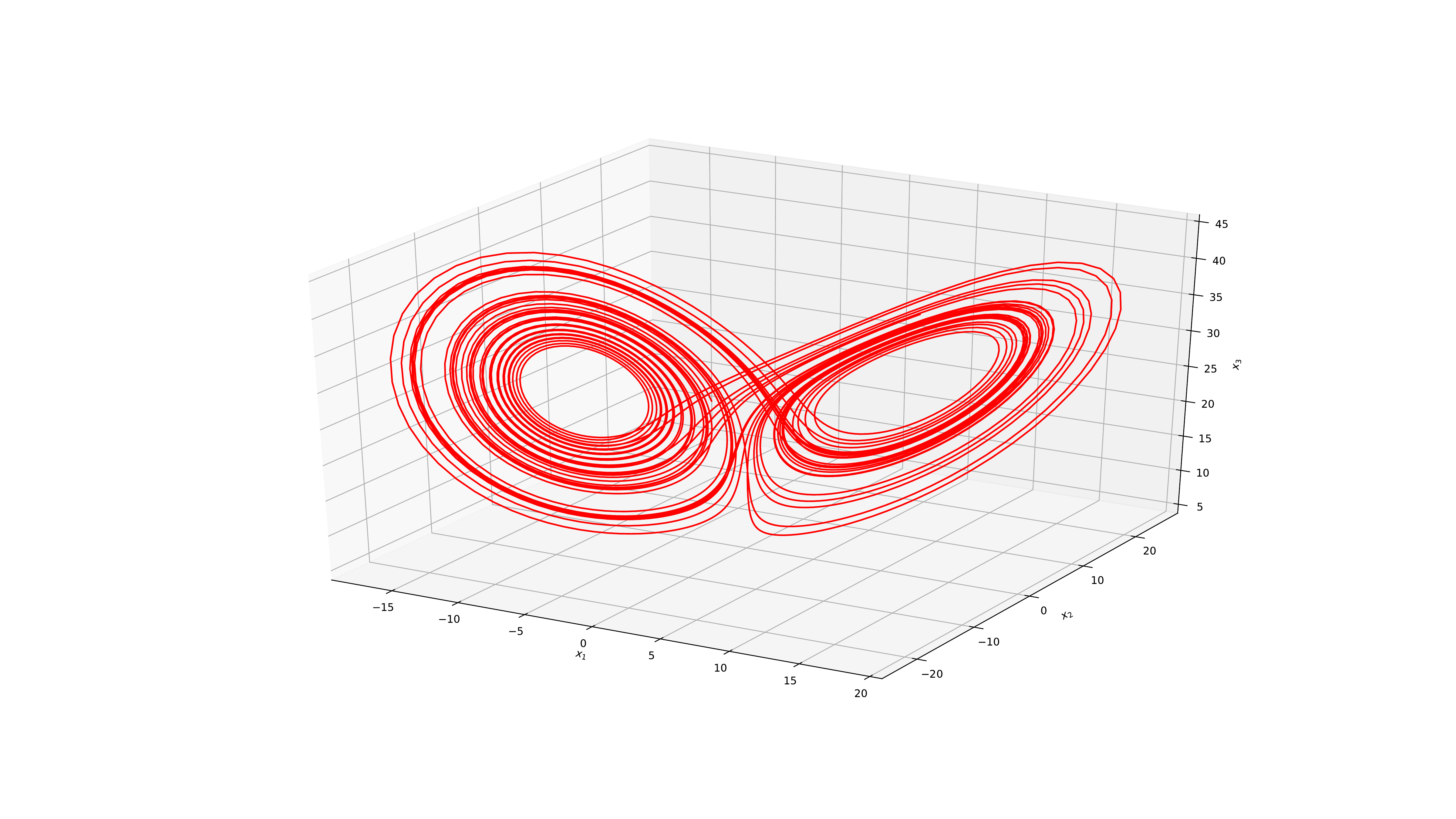}
		\includegraphics[width=\bwidth,clip, trim=100mm 40mm 80mm 40mm]{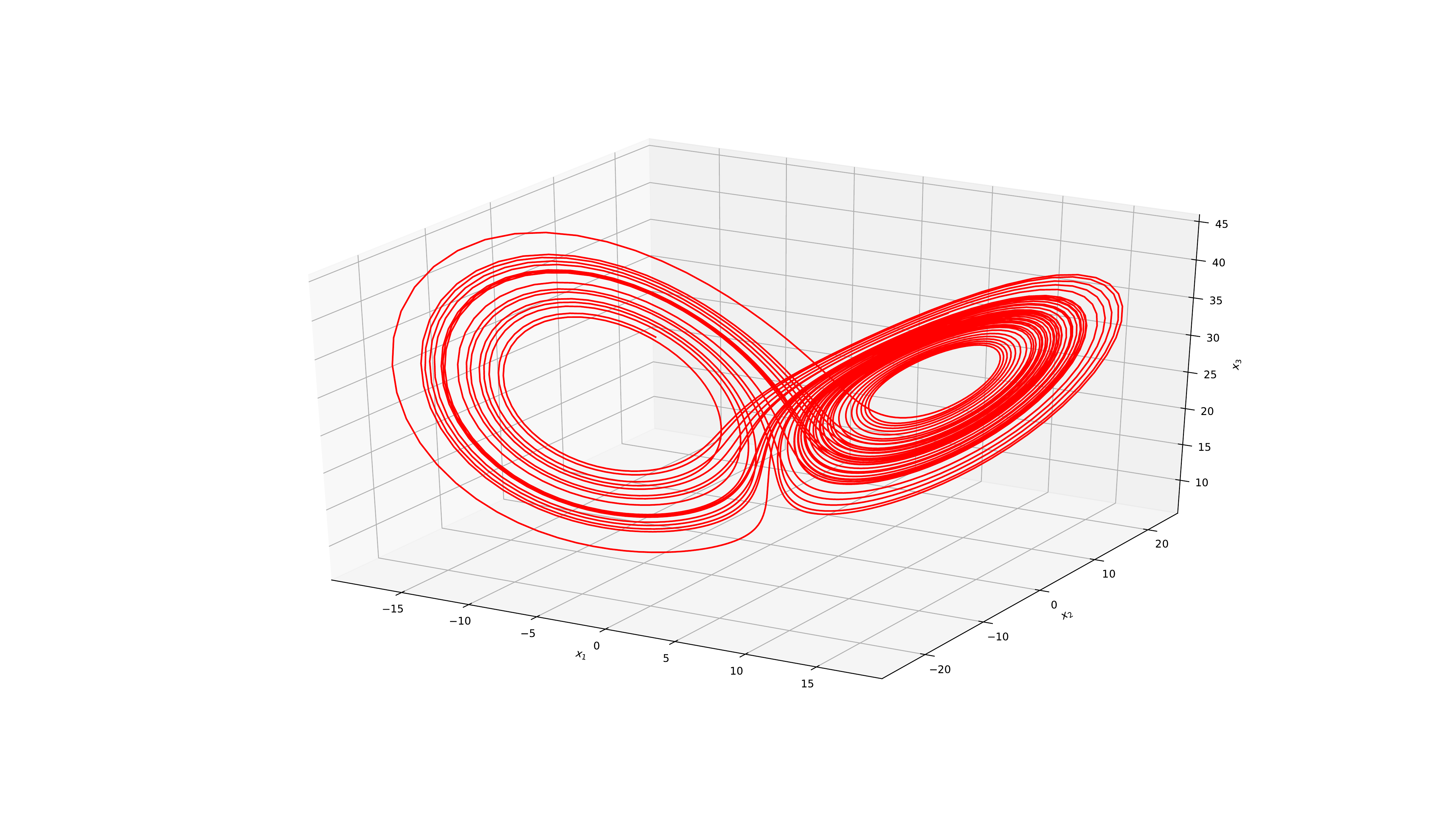}
		\includegraphics[width=\bwidth,clip, trim=100mm 40mm 80mm 40mm]{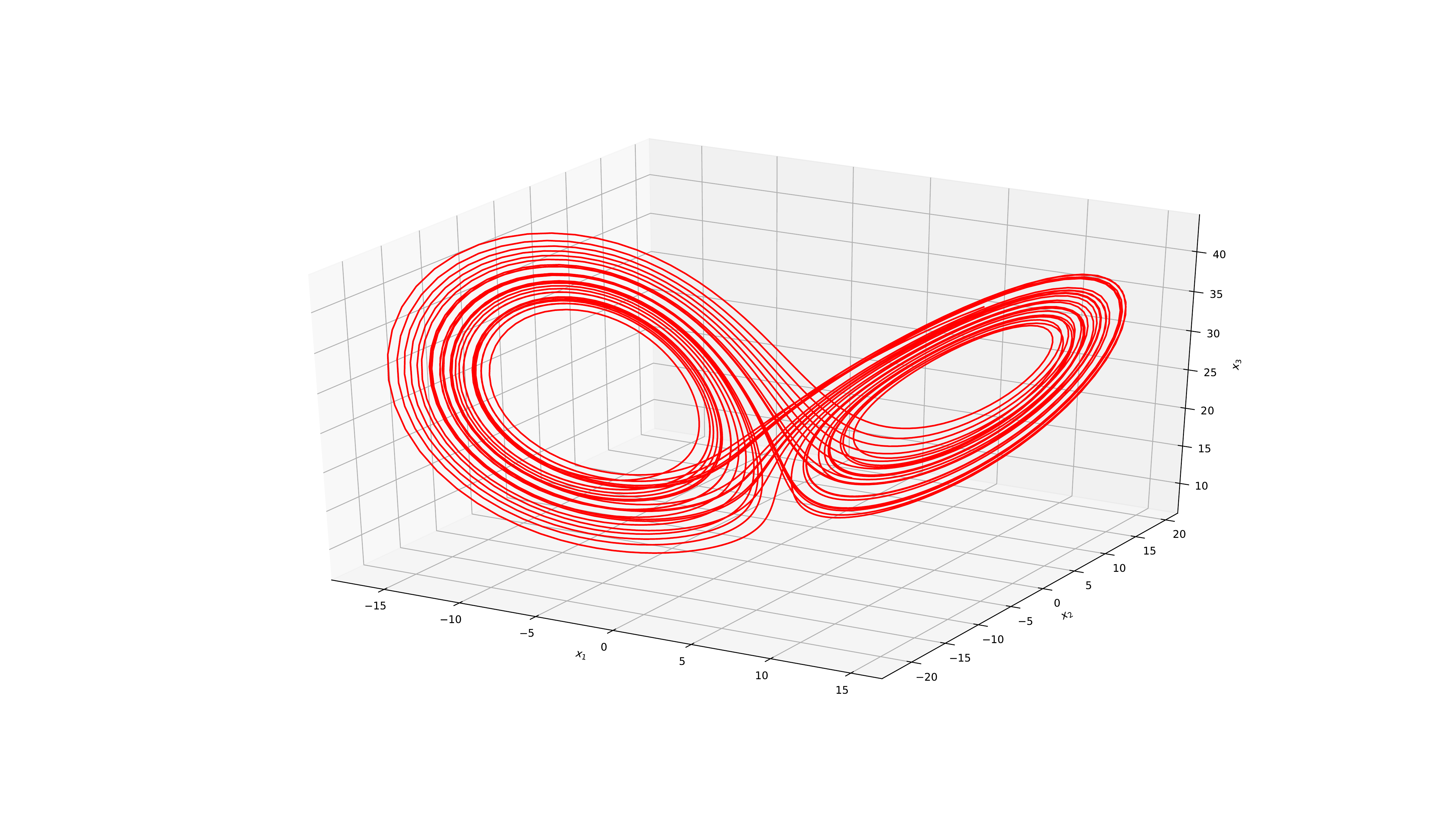}
	\end{subfigure}%
	
	\begin{subfigure}[b]{0.04\linewidth}
	    \rotatebox[origin=t]{90}{\scriptsize Determ}\vspace{0.9\linewidth}
	\end{subfigure}%
	\begin{subfigure}[t]{0.96\linewidth}
		\centering
		\includegraphics[width=\bwidth,clip, trim=100mm 40mm 80mm 40mm]{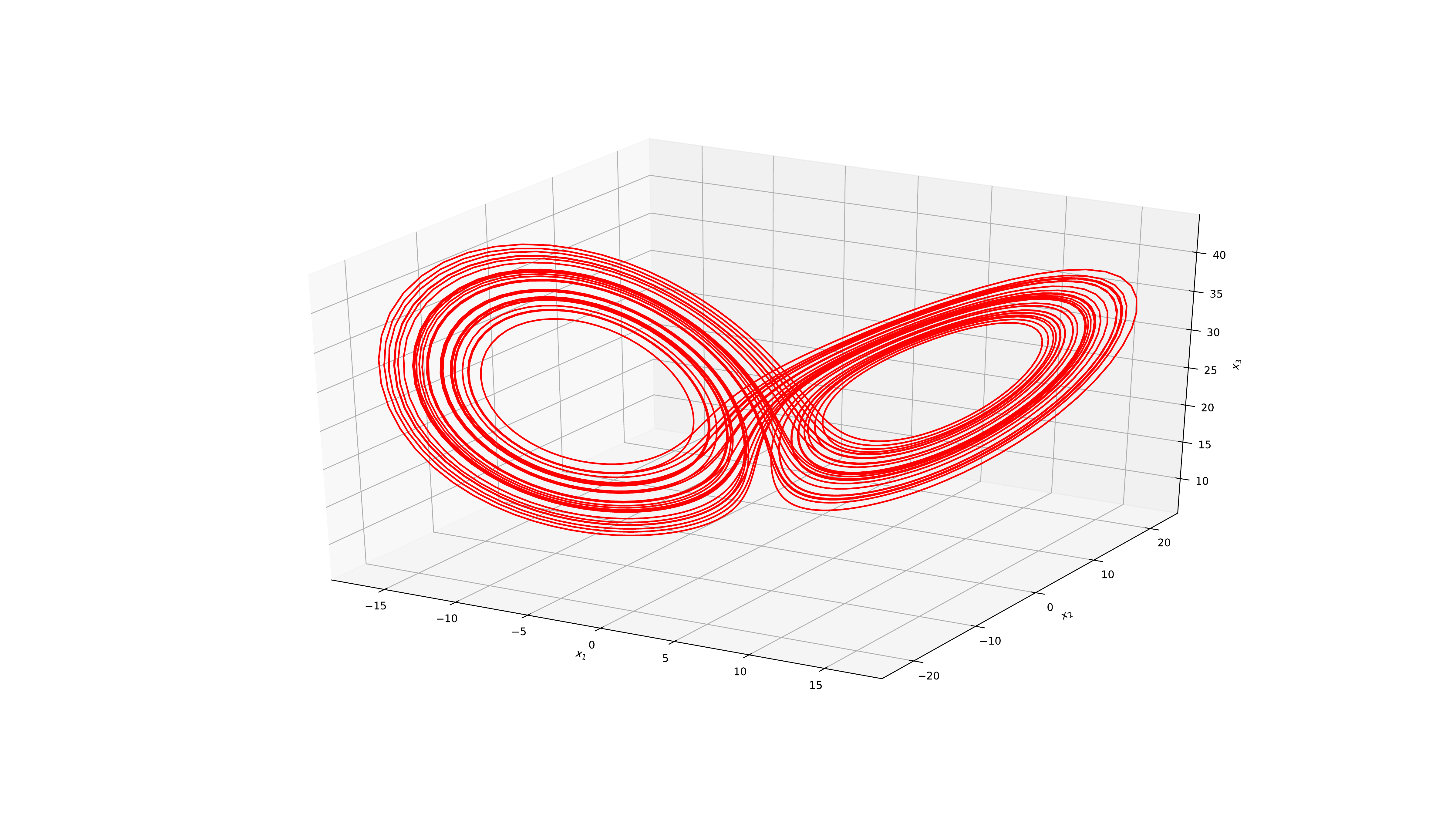}
		\includegraphics[width=\bwidth,clip, trim=100mm 40mm 80mm 40mm]{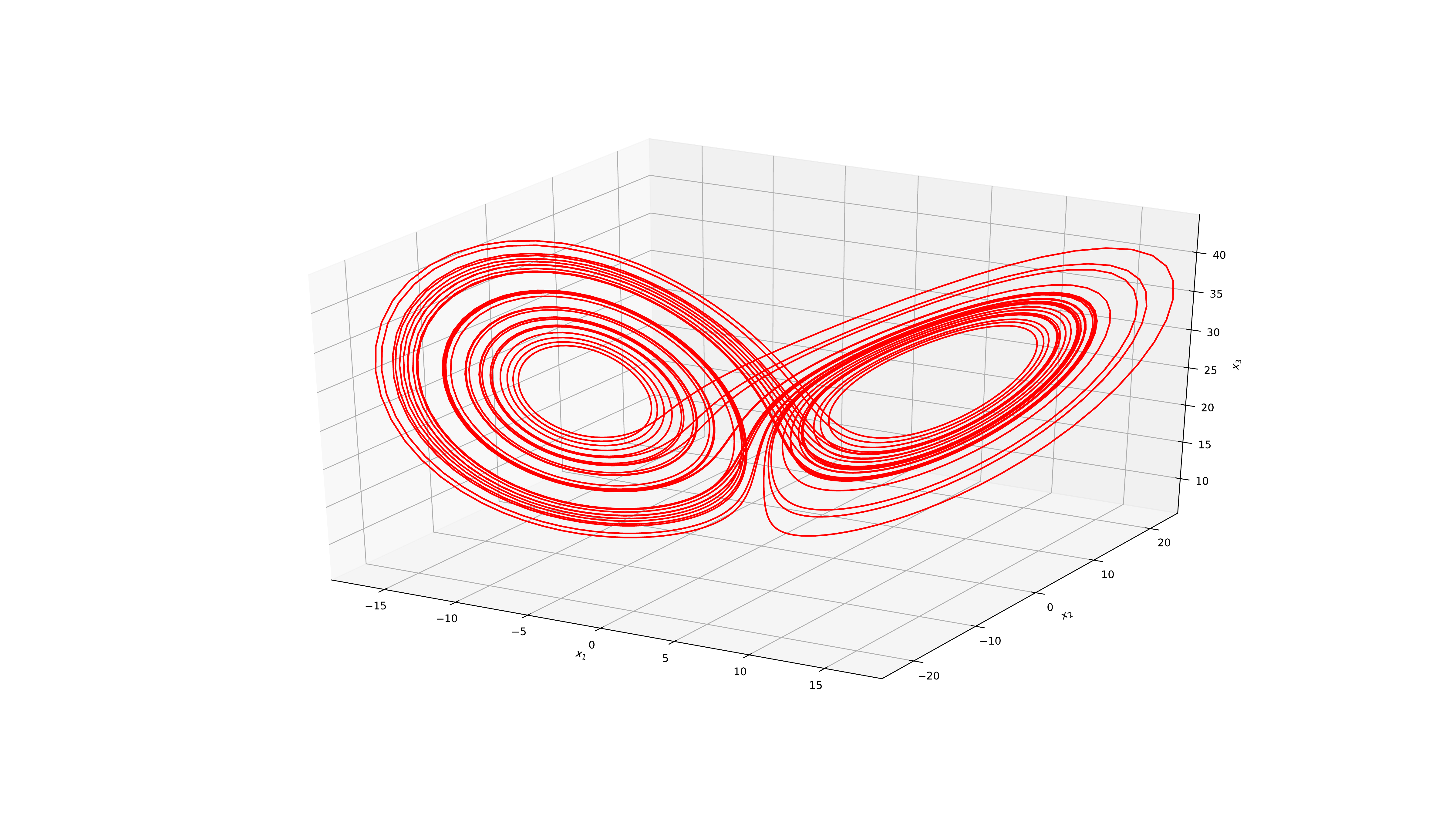}
		\includegraphics[width=\bwidth,clip, trim=100mm 40mm 80mm 40mm]{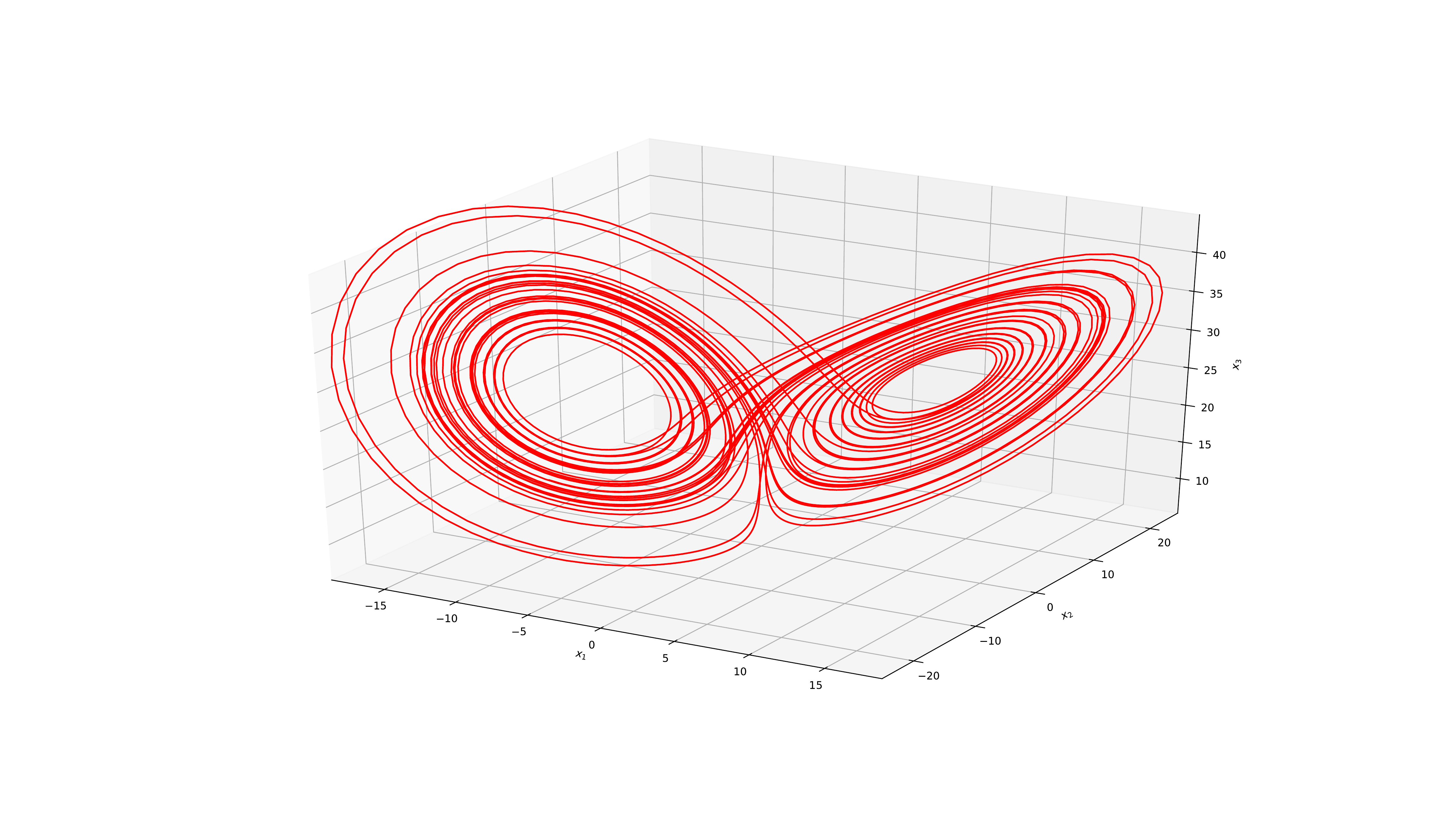}
		\includegraphics[width=\bwidth,clip, trim=100mm 40mm 80mm 40mm]{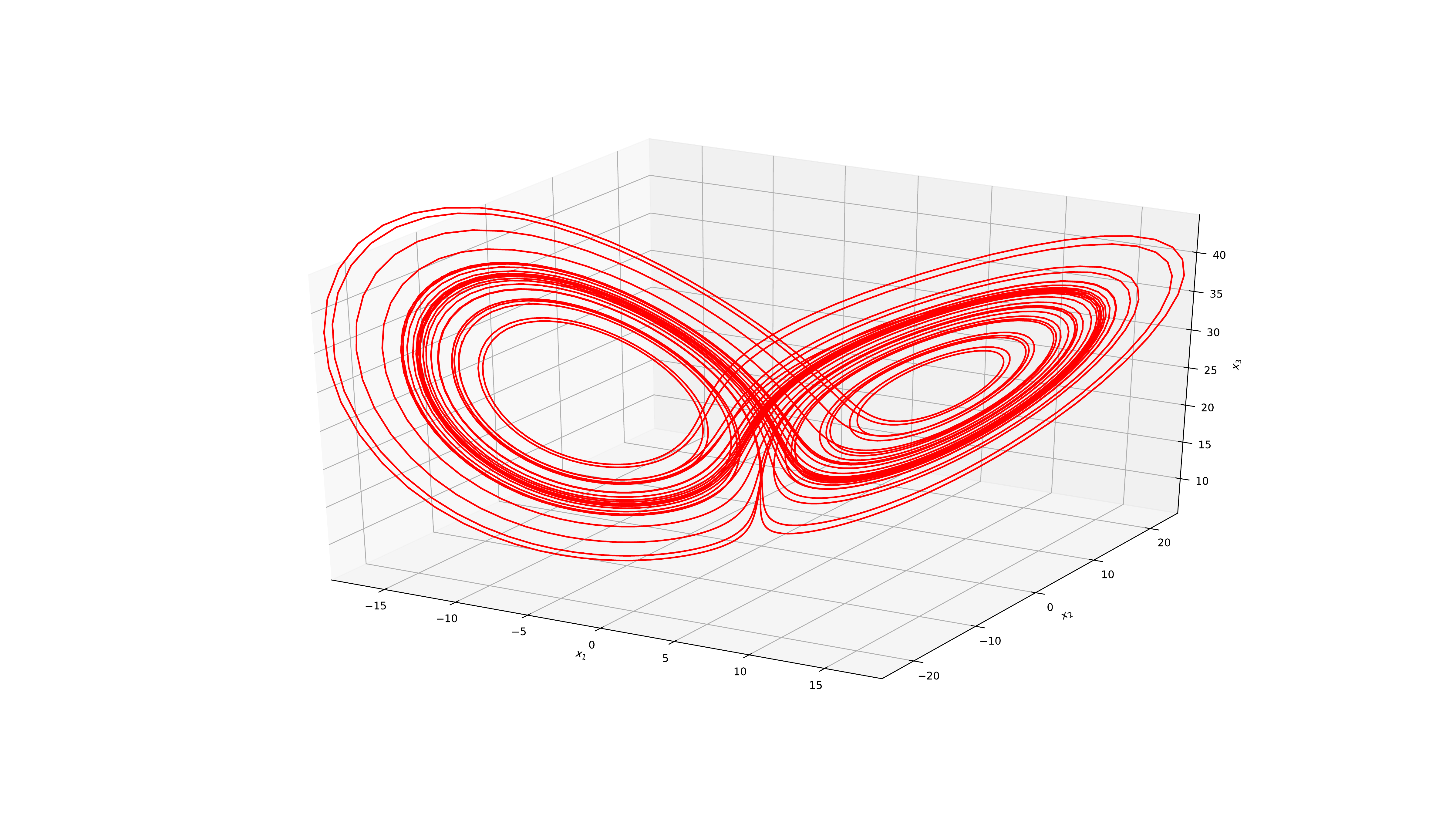}
	\end{subfigure}%
	
	\begin{subfigure}[b]{0.04\linewidth}
	    \rotatebox[origin=t]{90}{\scriptsize MAP}\vspace{0.9\linewidth}
	\end{subfigure}%
	\begin{subfigure}[t]{0.96\linewidth}
		\centering
		\includegraphics[width=\bwidth,clip, trim=100mm 40mm 80mm 40mm]{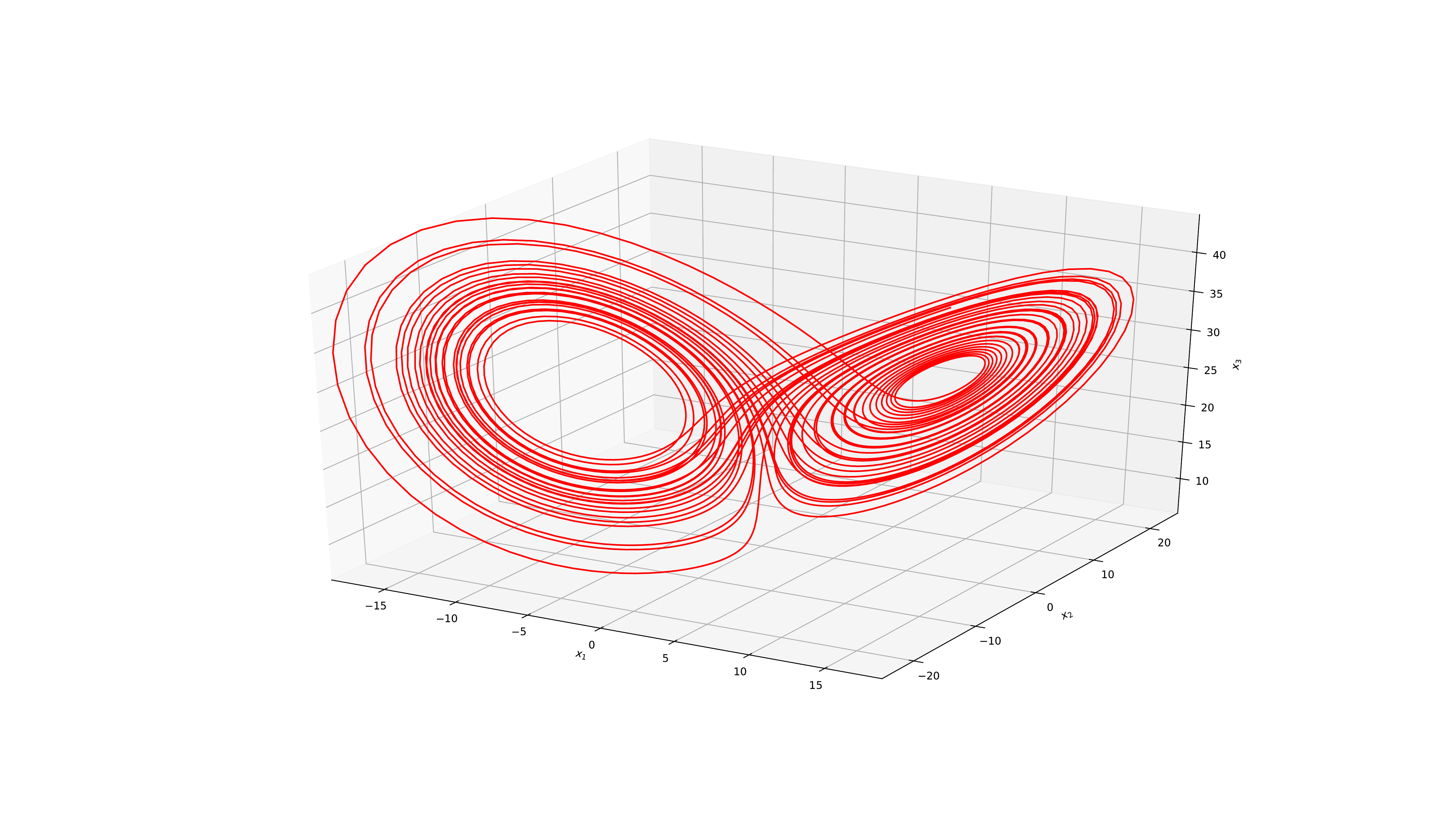}
		\includegraphics[width=\bwidth,clip, trim=100mm 40mm 80mm 40mm]{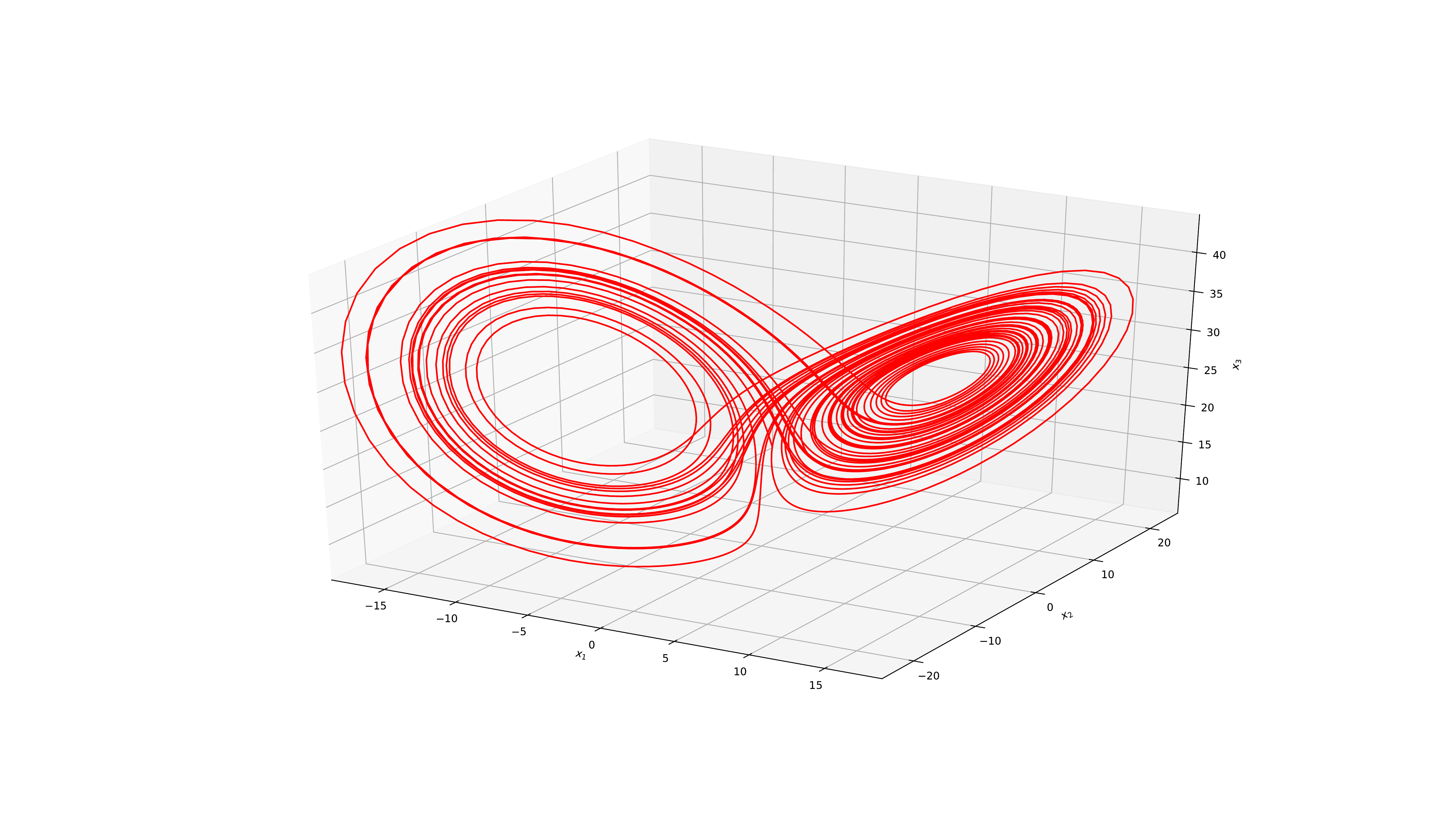}
		\includegraphics[width=\bwidth,clip, trim=100mm 40mm 80mm 40mm]{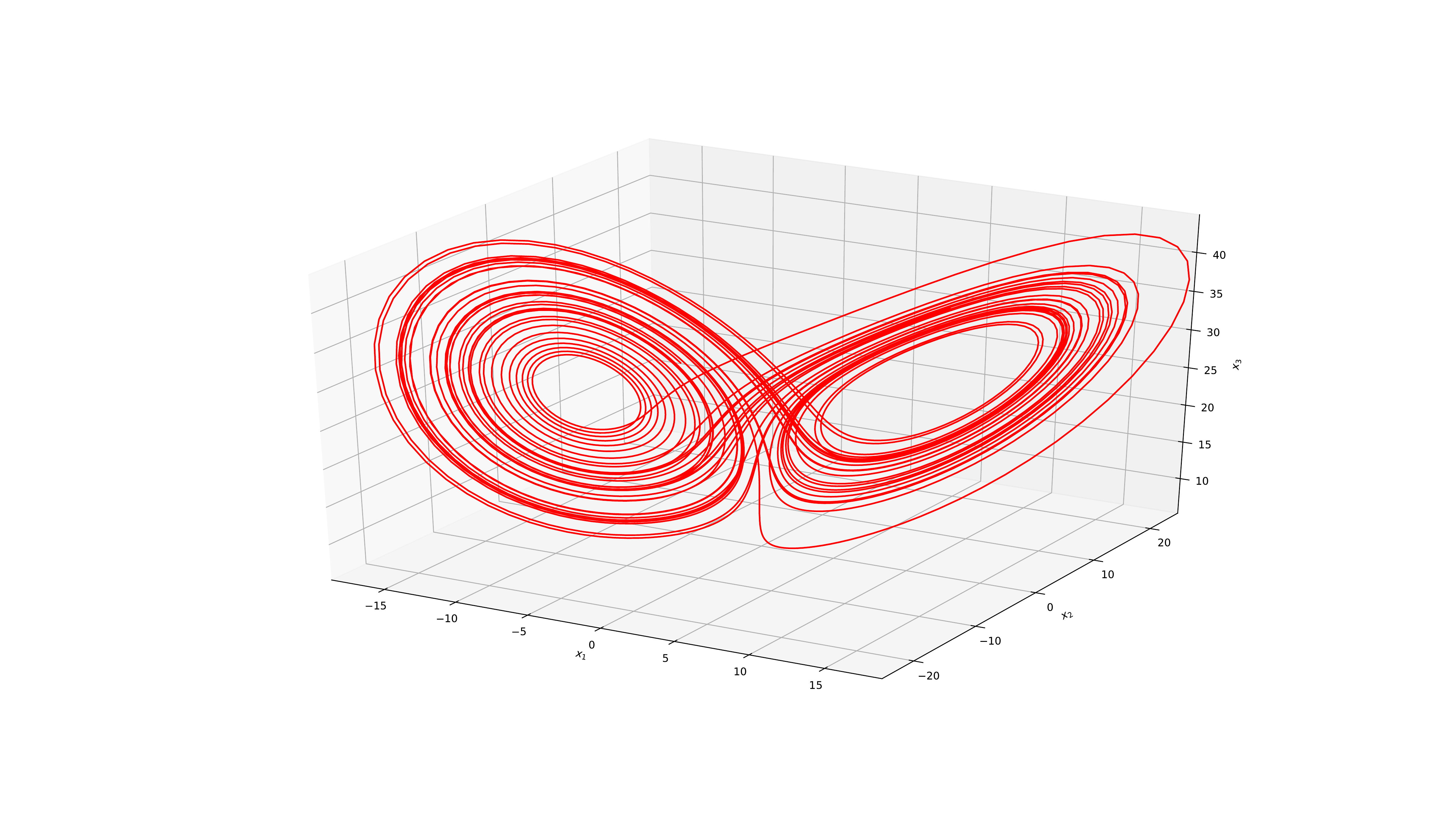}
		\includegraphics[width=\bwidth,clip, trim=100mm 40mm 80mm 40mm]{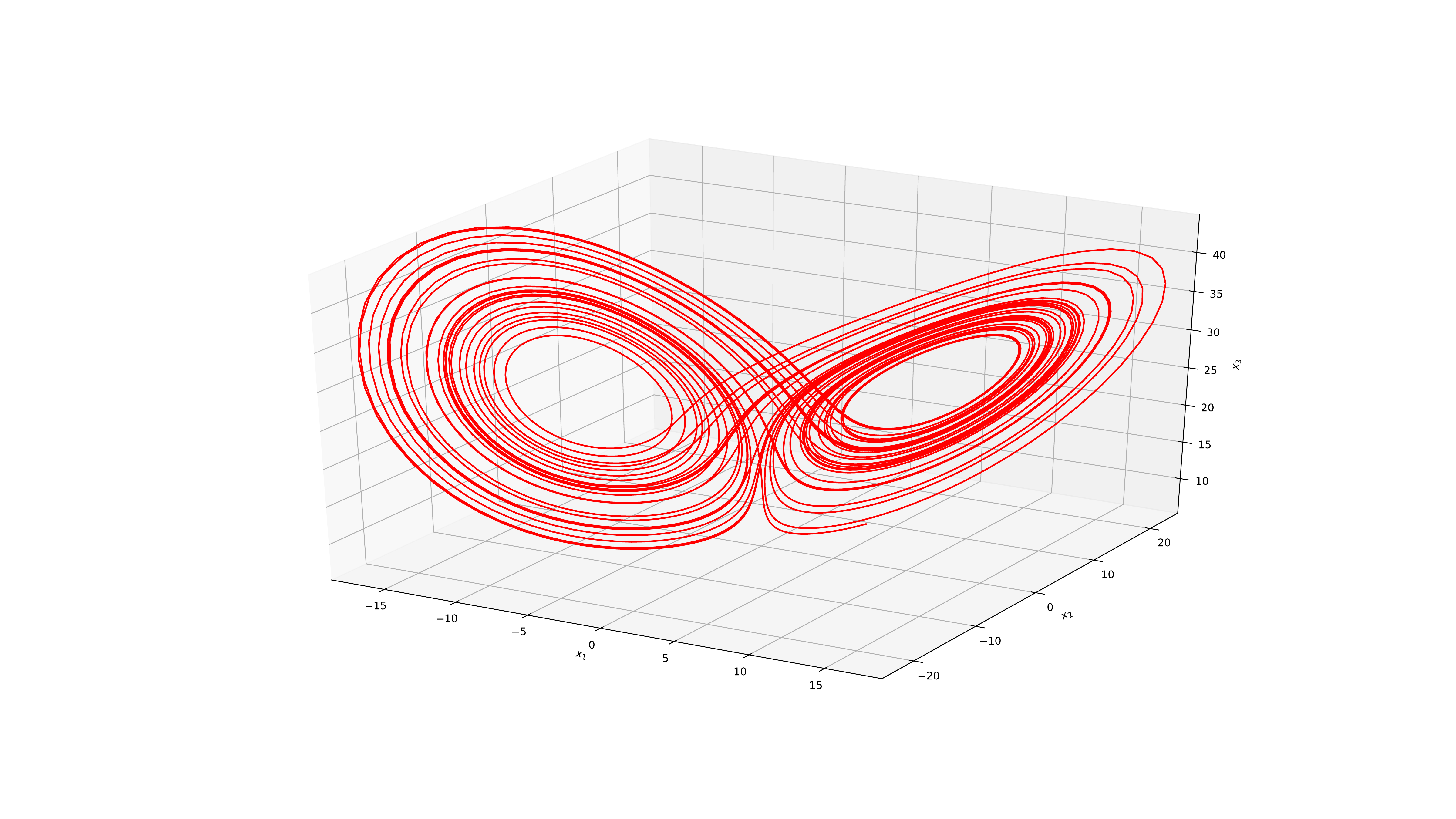}
	\end{subfigure}%
	
	\begin{subfigure}[b]{0.04\linewidth}
	    \rotatebox[origin=t]{90}{\scriptsize Full}\vspace{0.9\linewidth}
	\end{subfigure}%
	\begin{subfigure}[t]{0.96\linewidth}
		\centering
		\includegraphics[width=\bwidth,clip, trim=100mm 40mm 80mm 40mm]{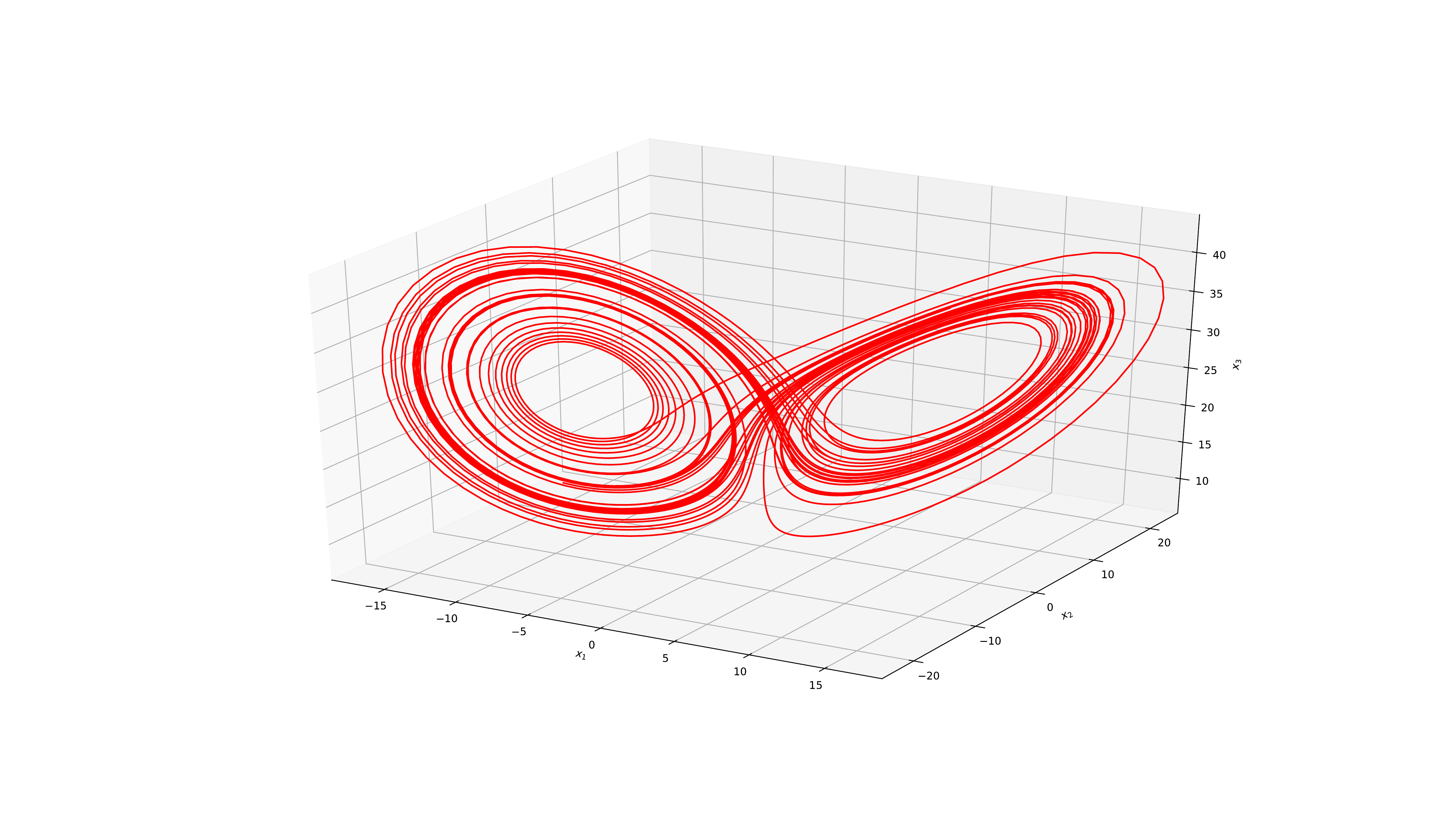}
		\includegraphics[width=\bwidth,clip, trim=100mm 40mm 80mm 40mm]{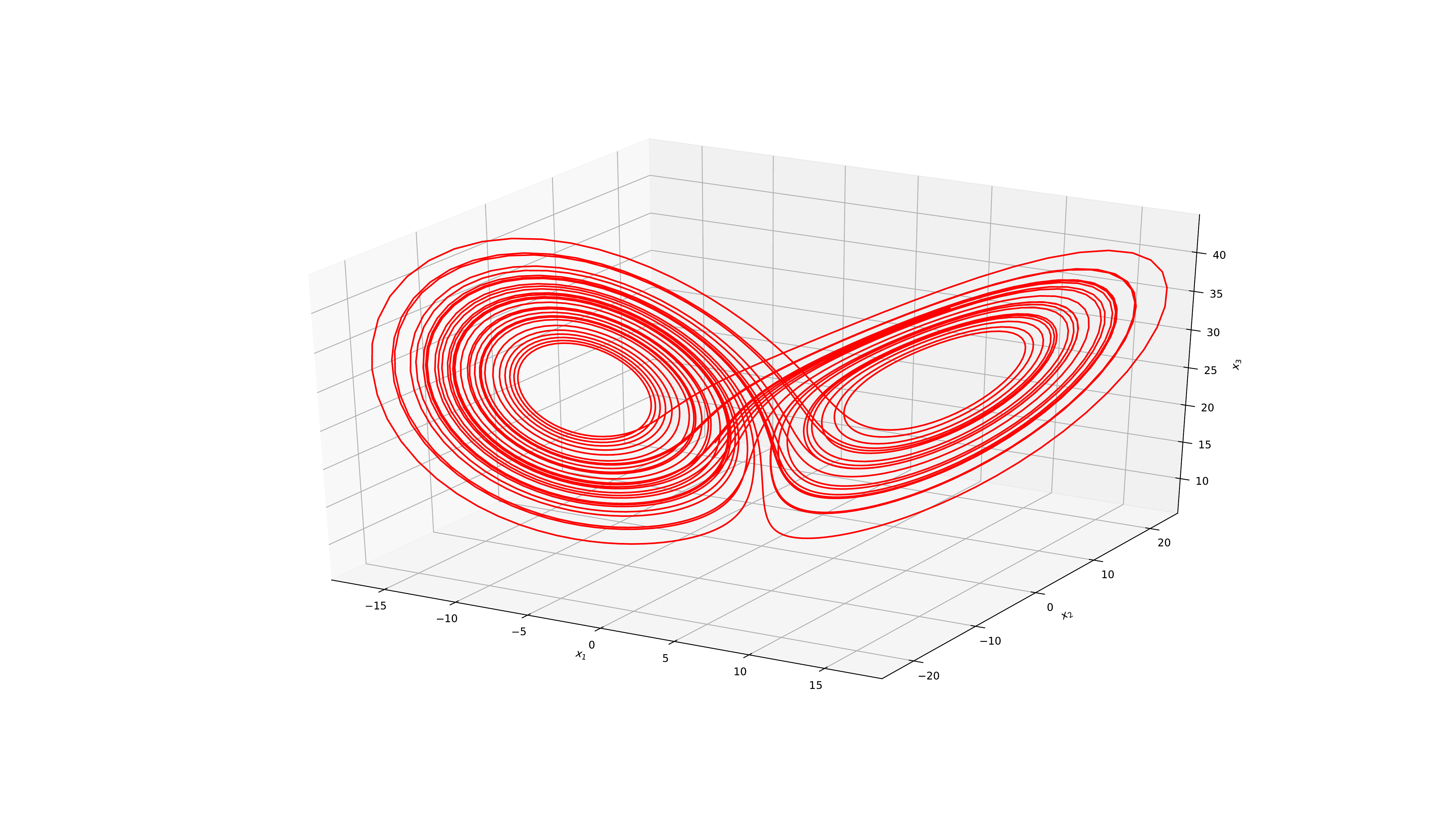}
		\includegraphics[width=\bwidth,clip, trim=100mm 40mm 80mm 40mm]{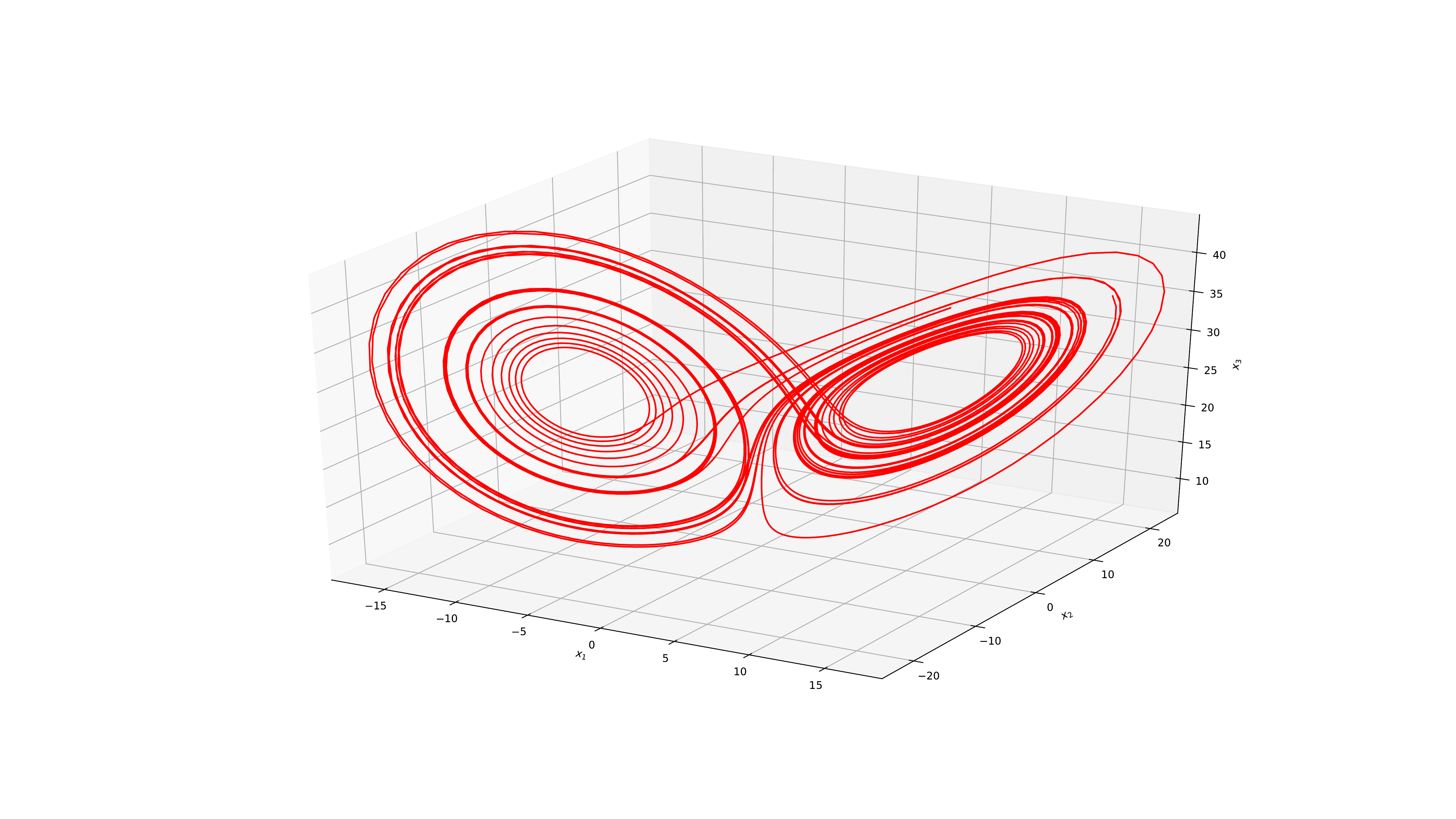}
		\includegraphics[width=\bwidth,clip, trim=100mm 40mm 80mm 40mm]{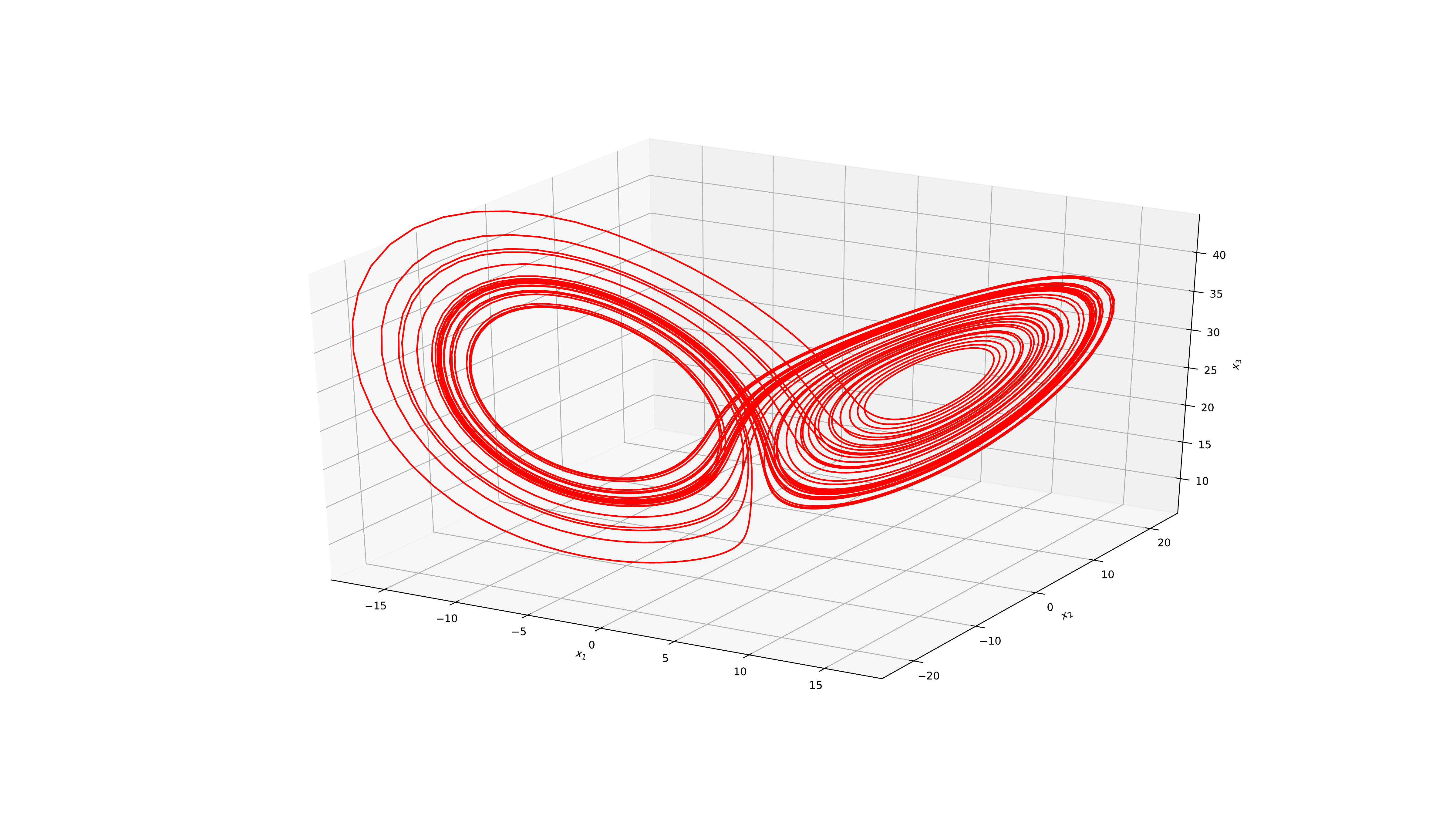}
	\end{subfigure}%
	
	\caption{Attactors generated by models trained on noisy data.}
    \label{fig:attactors_lorenz63_noisy}
\end{figure}
\endgroup

\begin{figure}
    \centering
    \includegraphics[width=\linewidth, clip, trim=0mm 0mm 0mm 0mm]{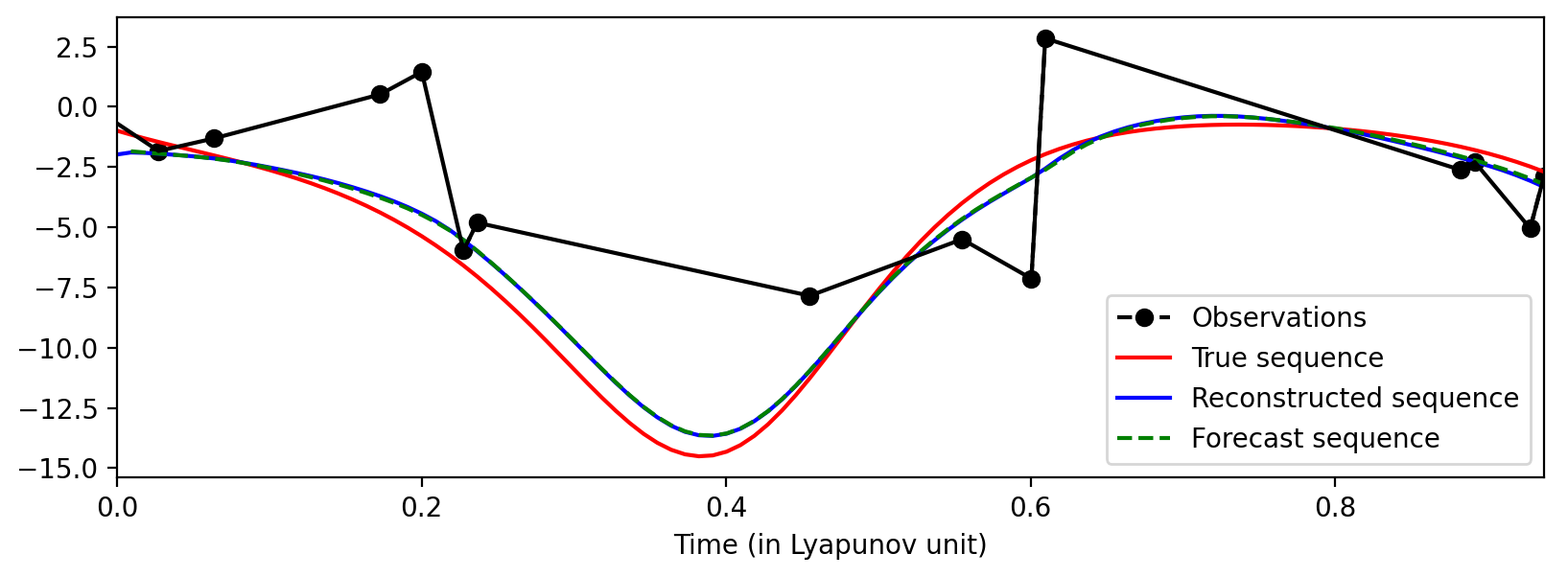}
    \caption{An example of the the first dimension of the L63 system reconstructed by the inference module of DAODEN\_determ trained on noisy and partial data. The observations are noisy ($r=33\%$) and observed partially with a missing rate of 87.5\%.}
    \label{fig:smoother_partial}
\end{figure}

In real life applications, we cannot always measure a process regularly with a high sampling frequency. Hence, we address here the problem of learning dynamical systems from not only noisy but also partial observations\footnote{The term ``partial'' in this context means the observations are not complete at every time step. Some components of the observations may be missing, in both spatial and temporal dimensions; however, all the components of the states of the system are seen at least once. For the cases where some components of the systems are never observed, please refer to \cite{ayed_learning_2019, ouala_learning_2020}.}. Specifically, we consider a case study where the noisy L63 data are sampled partially, both in time and in space, with a missing rate of 87.5\% (see Fig. \ref{fig:smoother_partial}). For this configuration, baseline schemes do not apply. We report in Table. \ref{tab:L63_partial} and Fig. \ref{fig:attactors_lorenz63_partial} the performance of the different configurations of the proposed framework. If the noise level is not significantly high ($r$=33.3\% or $r$=66.7\%), all the models are able to capture the dynamical characteristics of the data. When the noise level is small, BiNN\_EnKS tends to perform better than DAODEN. However, when the data are awash with noise, BiNN\_EnKS does not work well anymore. On the other hand, DAODEN models, especially DAODEN\_full work well in these cases. This may come from the capacity of LSTM architectures to capture long-term correlations in data.  

{}

 \begingroup
\begin{table*}[]
    \caption {Performance of models trained on noisy and partial L63 data. The data are observed partially, both in time and in space, with a missing rate of 87.5\%.  For each index, the best score is marked in \textbf{bold} and the second best score is marked in \textit{italic}.}
    \label{tab:L63_partial}
    \footnotesize
    \centering
    \begin{tabular}{ll*{4}c}
    \toprule
    \multicolumn{2}{c}{\multirow{2}{*}{Model}} & \multicolumn{4}{c}{$r$} \\
    & & 8.5\% & 16.7\% & 33.3\% & 66.7\%\\
    \midrule 
    
    \midrule 
    \multirow{4}{*}{BiNN\_EnKS}
    & $e_4$      & \textit{0.129±0.081}  & \textbf{0.143±0.065} & 0.350±0.204  &\textit{ 0.973±0.649} \\
    & $rec$      & \textbf{0.721±0.204}  & \textbf{1.062±0.401} &  2.342±1.622 & 6.675±1.410 \\
    & $\pi_{0.5}$      & 1.873±1.034  & \textbf{2.146±1.048} &  \textit{1.616±1.042} & \textit{0.290±0.153}\\
    &  $\lambda_1$  & 0.801±0.016  & 0.782±0.012 & 0.304±0.147  & -1.588±0.009 \\

    \midrule 
    \multirow{4}{*}{DAODEN\_determ}
    & $e_4$      & 0.135±0.082  & 0.170±0.105 &  \textit{0.290±0.202} & 25.034±19.821\\
    & $rec$      & 1.300±1.525  & 1.448±1.332 &  1.985±1.474 & 4.222±2.191\\
    & $\pi_{0.5}$      & 2.399±1.360  & \textit{2.140±1.110} &  1.441±0.823 & 0.022±0.087\\
    &  $\lambda_1$  & 0.905±0.014 &  0.888±0.013 & 0.809±0.018 & -0.011±0.014\\

    \midrule 
    \multirow{4}{*}{DAODEN\_MAP}
    & $e_4$      & 0.175±0.119  & 0.325±0.235 &  0.459±0.343 & 9.105±7.136\\
    & $rec$      & 1.352±0.997  & 1.705±1.434 &  \textit{1.972±1.247} & \textit{3.704±1.180}\\
    & $\pi_{0.5}$      & \textbf{2.628±1.448}  & 1.706±1.125 &  1.505±0.949 & 0.064±0.216\\
    &  $\lambda_1$  & 0.894±0.010 &  0.844±0.016 & 0.736±0.017 & 0.453±0.030\\

    \midrule 
    \multirow{4}{*}{DAODEN\_full}
    & $e_4$      & \textbf{0.089±0.062}  & \textit{0.158±0.104} &  \textbf{0.162±0.104} & \textbf{0.254±0.142}\\
    & $rec$      & \textit{1.052±0.612}  & \textit{1.268±0.718} &  \textbf{1.685±0.928} & \textbf{2.725±1.356}\\
    & $\pi_{0.5}$      & \textit{2.590±1.193}  & 1.943±0.904 &  \textbf{1.984±0.949} & \textbf{1.347±1.014}\\
    &  $\lambda_1$  & 0.892±0.011 & 0.846±0.013 & 0.859±0.013 & 0.720±0.019\\


    
    \bottomrule
    \end{tabular}
    \normalsize
\end{table*}
\endgroup

\newcommand{\cwidth}{0.24\linewidth}%
\newcommand{\twidth}{0.05\linewidth}%
\begin{figure}
    \centering
	\begin{subfigure}[t]{0.04\linewidth}
		\hfill
	    \vspace{-2.5mm}
		\caption*{}
	\end{subfigure}%
	\begin{subfigure}[t]{0.96\linewidth}
		\hspace{2.5mm} $r=8.5\%$ \hspace{\twidth} $r=16.7\%$ \hspace{\twidth} $r=33.3\%$ \hspace{\twidth} $r=66.7\%$ \hfill
	\end{subfigure}%
	\vspace{-2mm}
	\begin{subfigure}[b]{0.04\linewidth}
	    \rotatebox[origin=t]{90}{\scriptsize EnKS}\vspace{0.7\linewidth}
	\end{subfigure}%
	\begin{subfigure}[t]{0.96\linewidth}
	    \centering
		\includegraphics[width=\cwidth,clip, trim=100mm 40mm 80mm 40mm]{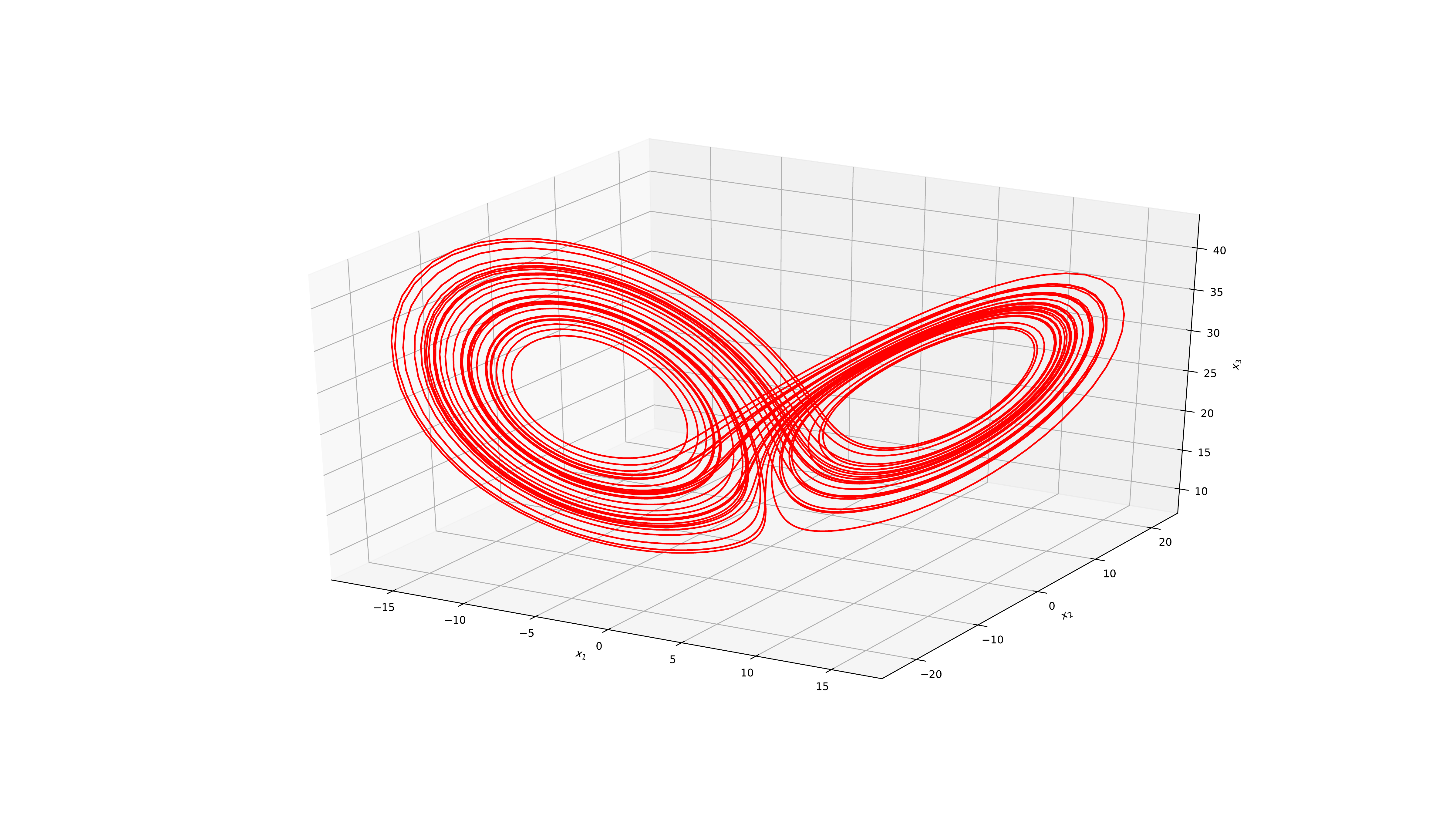}
		\includegraphics[width=\cwidth,clip, trim=100mm 40mm 80mm 40mm]{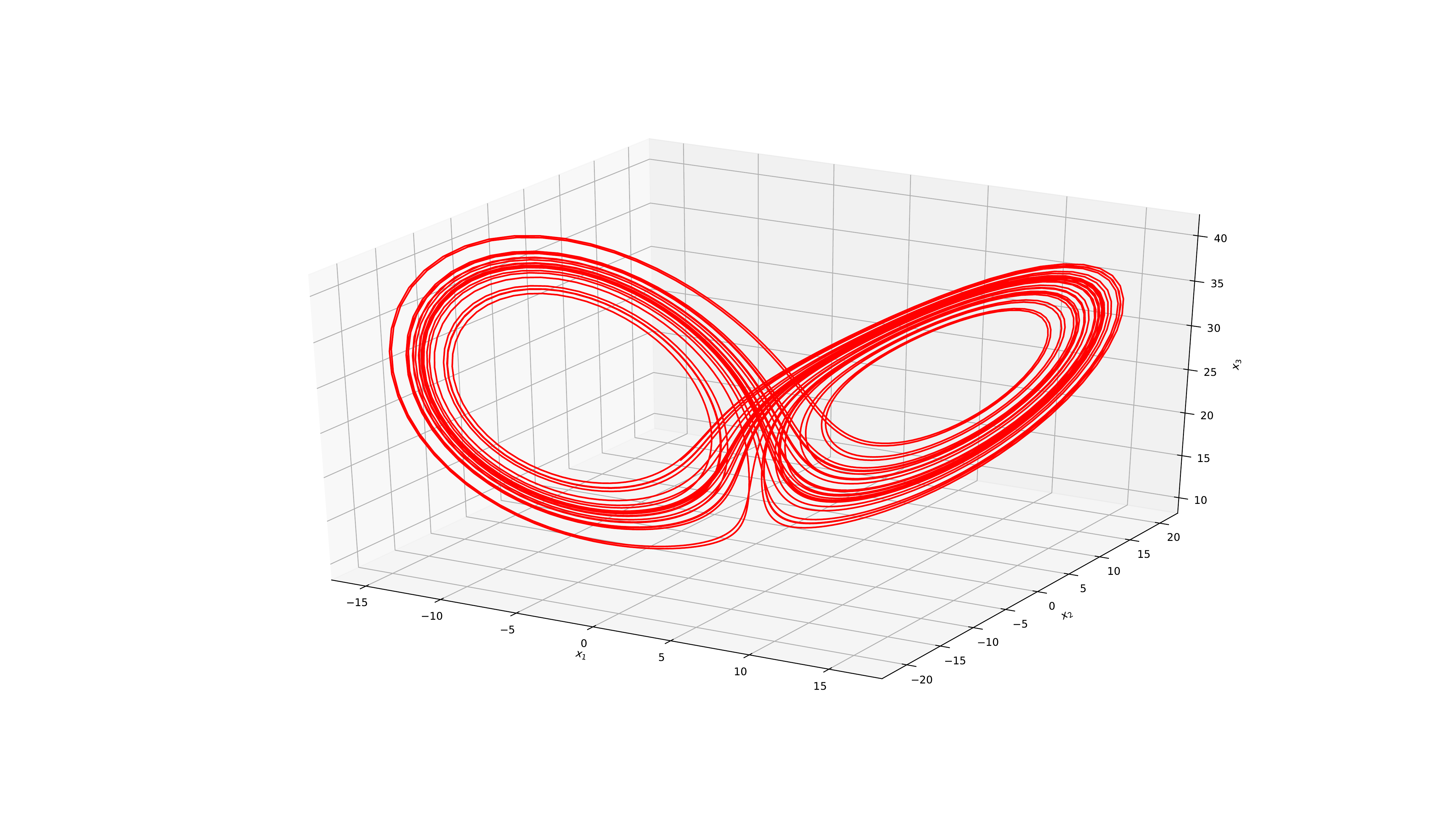}
		\includegraphics[width=\cwidth,clip, trim=100mm 40mm 80mm
		40mm]{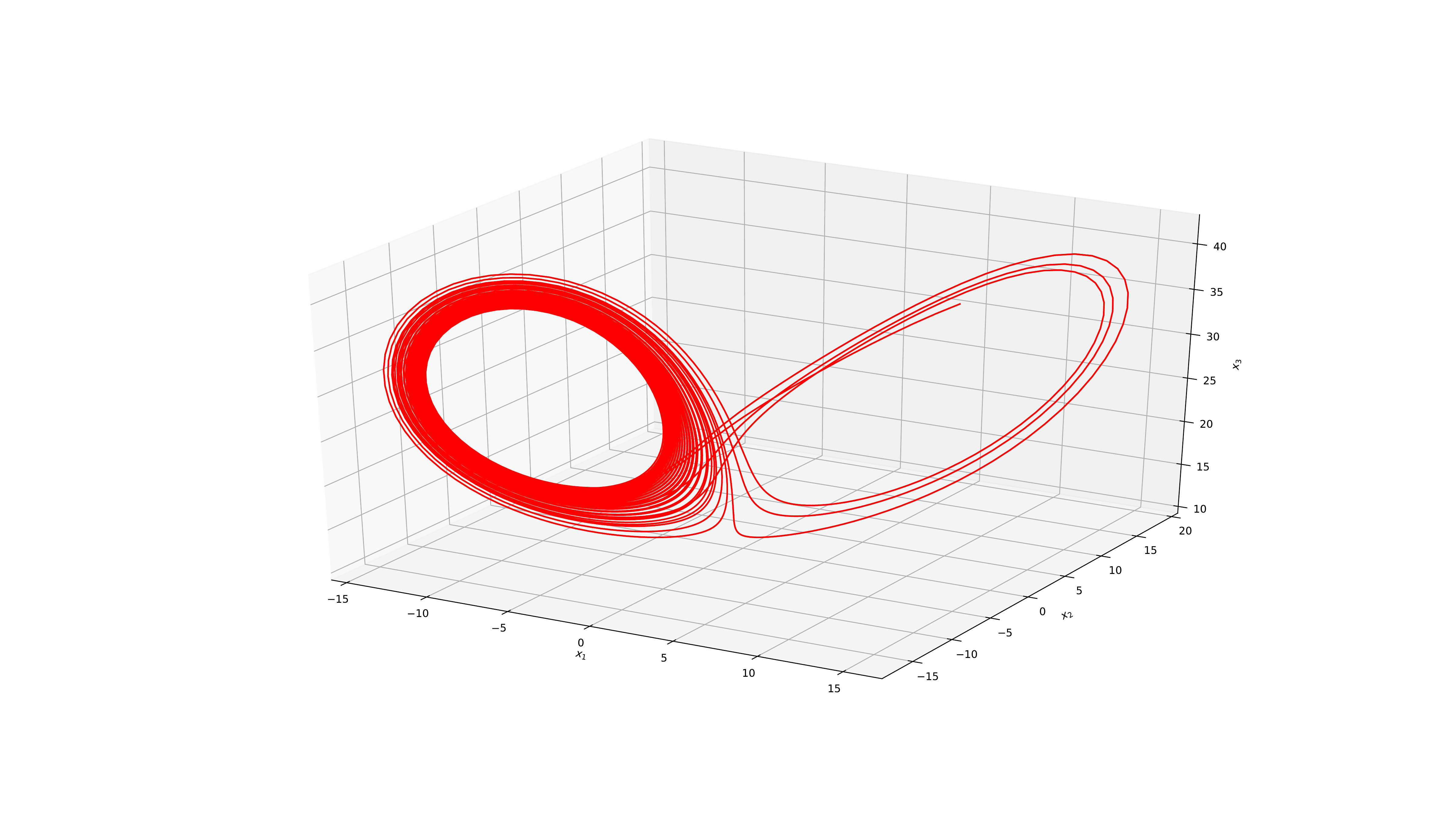}
		\includegraphics[width=\cwidth,clip, trim=100mm 40mm 80mm 40mm]{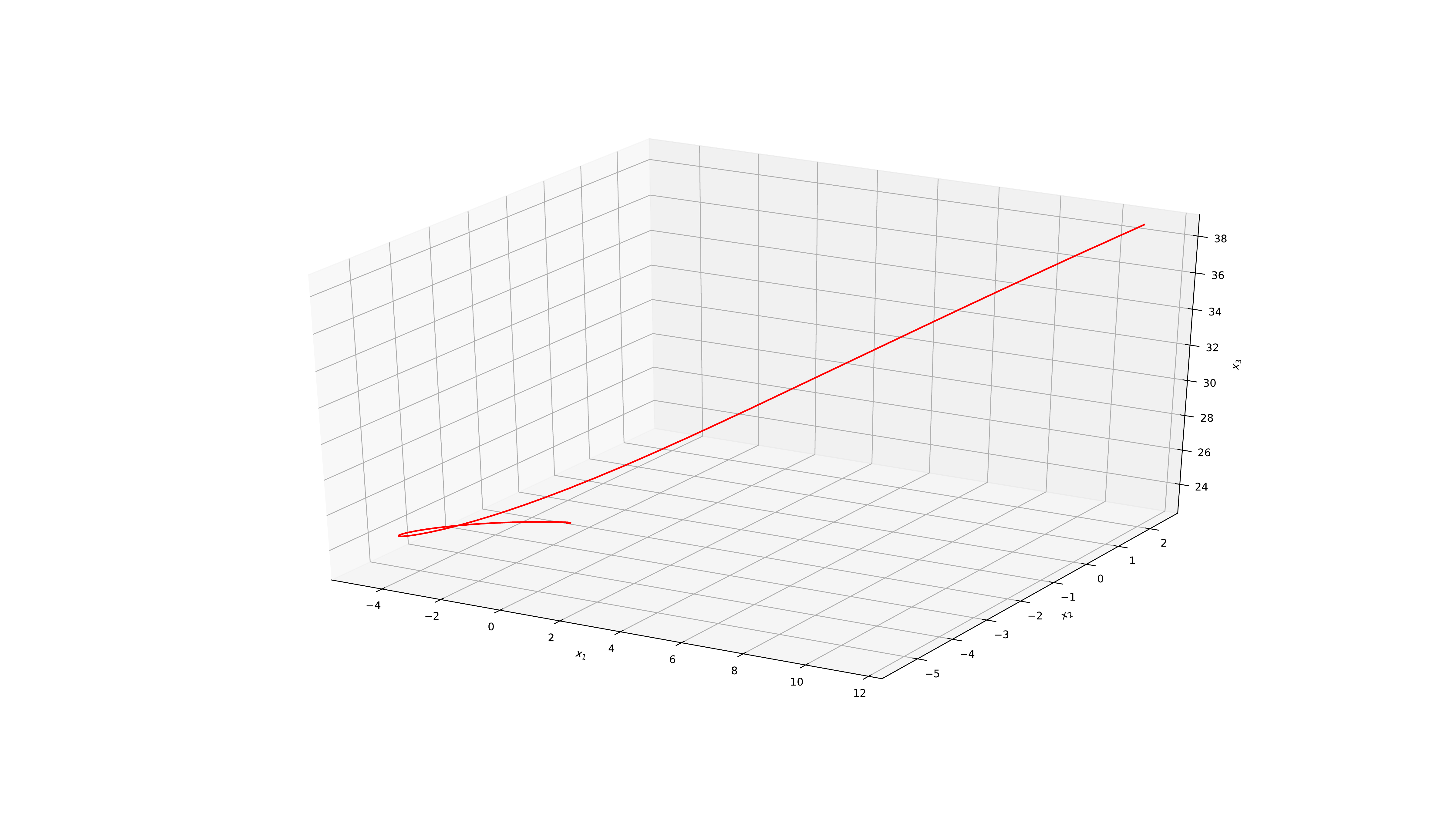}
	\end{subfigure}%
	
	\begin{subfigure}[b]{0.04\linewidth}
	    \rotatebox[origin=t]{90}{\scriptsize Determ}\vspace{0.6\linewidth}
	\end{subfigure}%
	\begin{subfigure}[t]{0.96\linewidth}
		\centering
		\includegraphics[width=\cwidth,clip, trim=100mm 40mm 80mm 40mm]{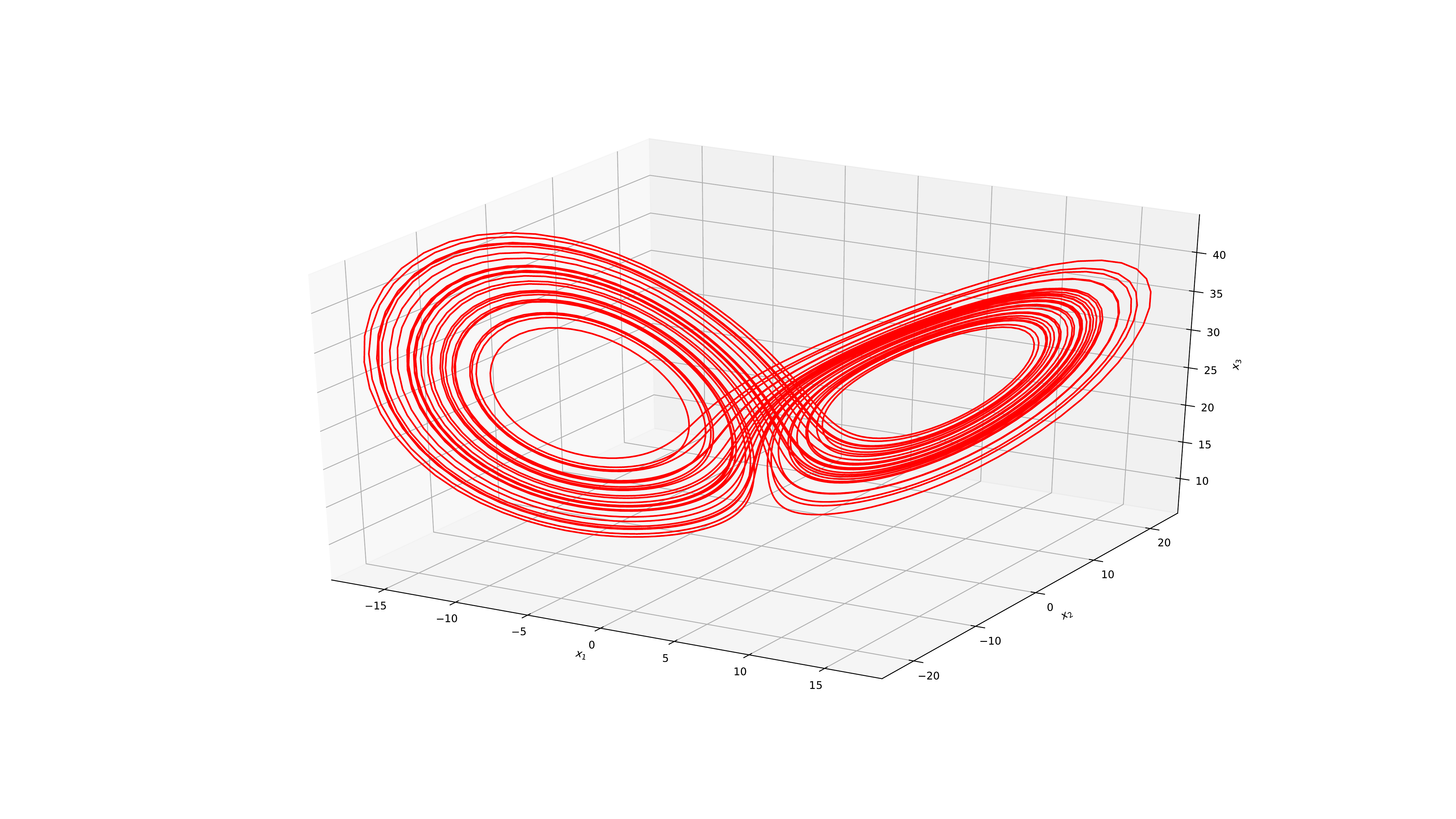}
		\includegraphics[width=\cwidth,clip, trim=100mm 40mm 80mm 40mm]{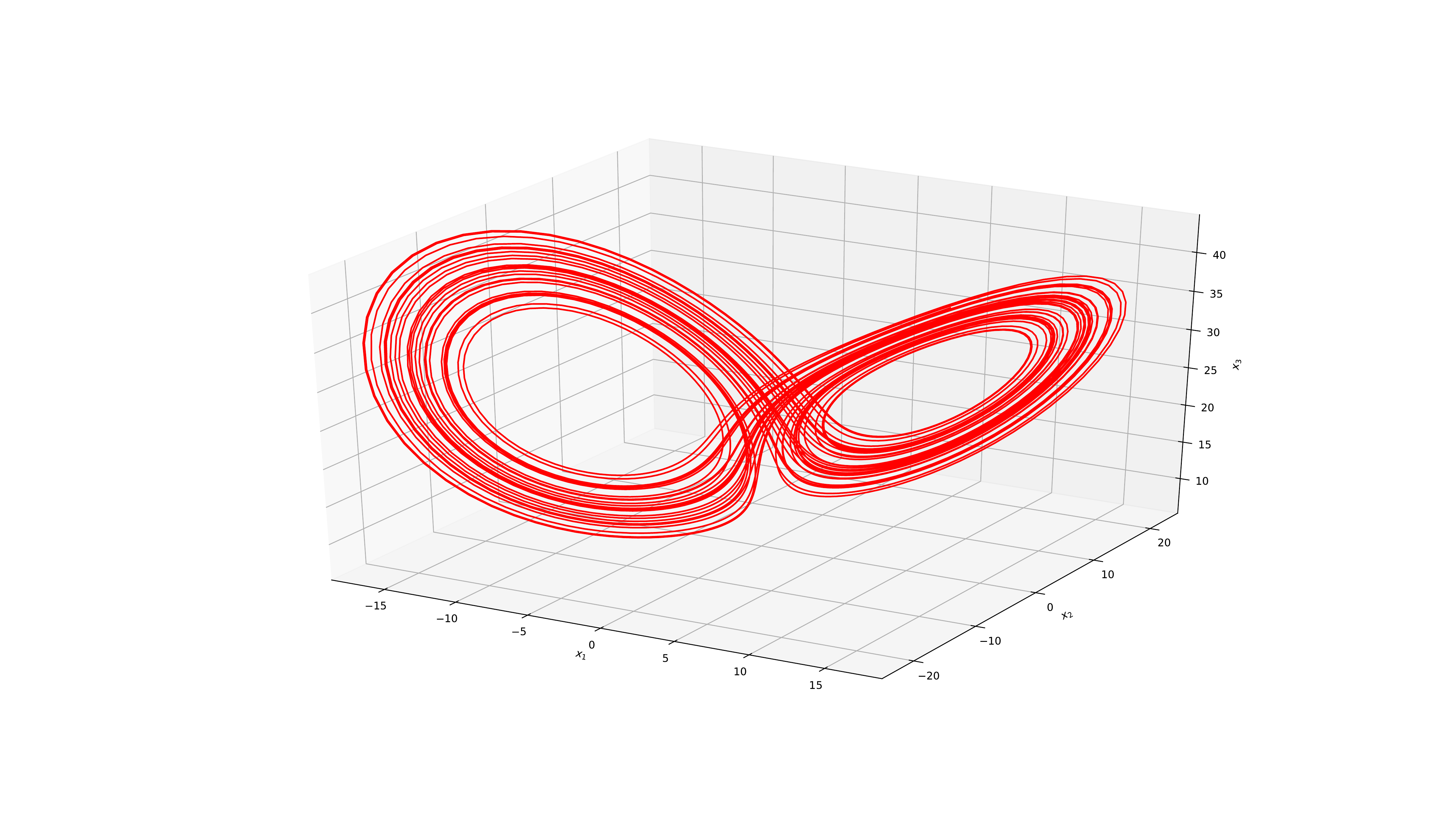}
		\includegraphics[width=\cwidth,clip, trim=100mm 40mm 80mm 40mm]{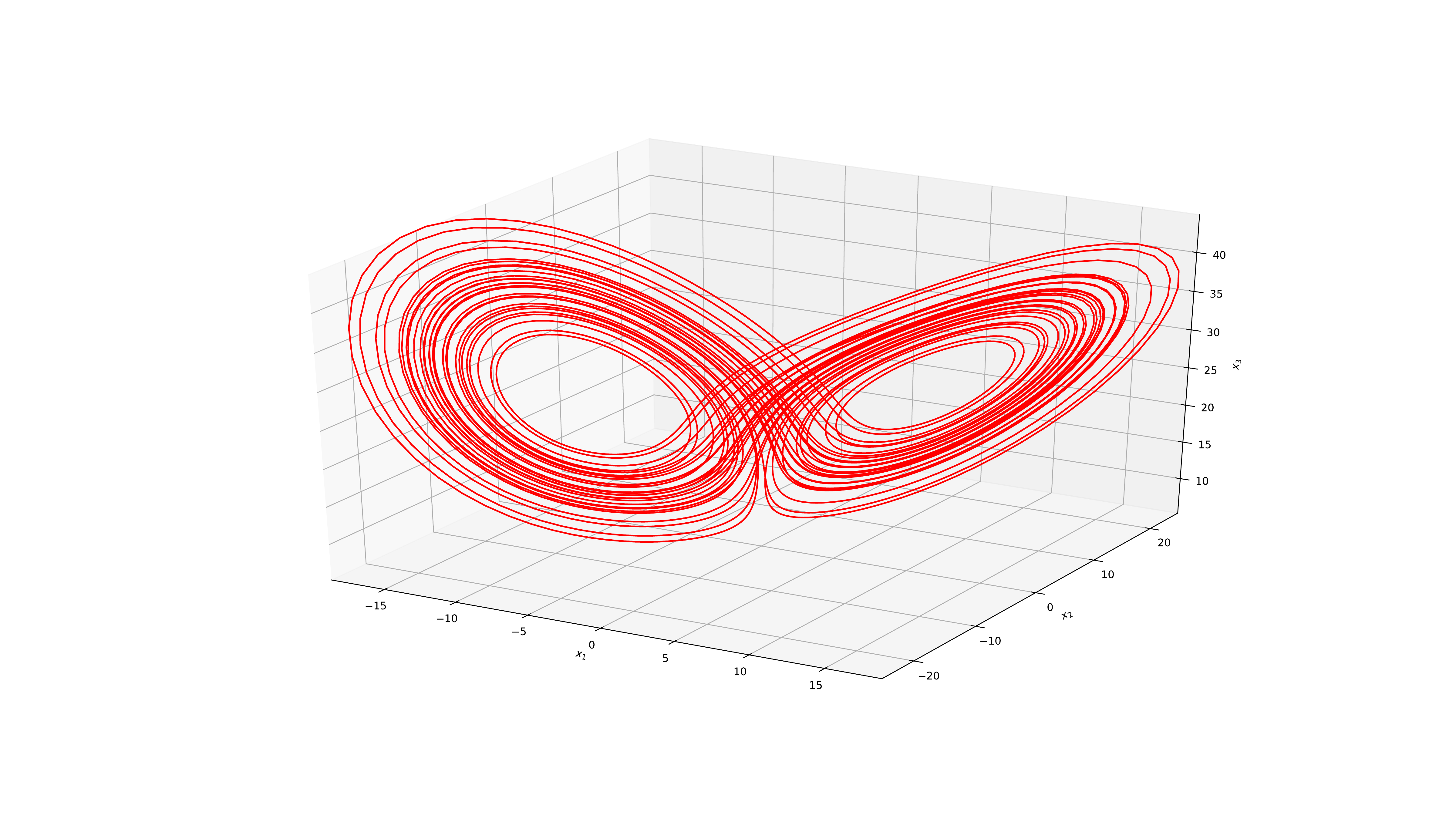}
		\includegraphics[width=\cwidth,clip, trim=100mm 40mm 80mm 40mm]{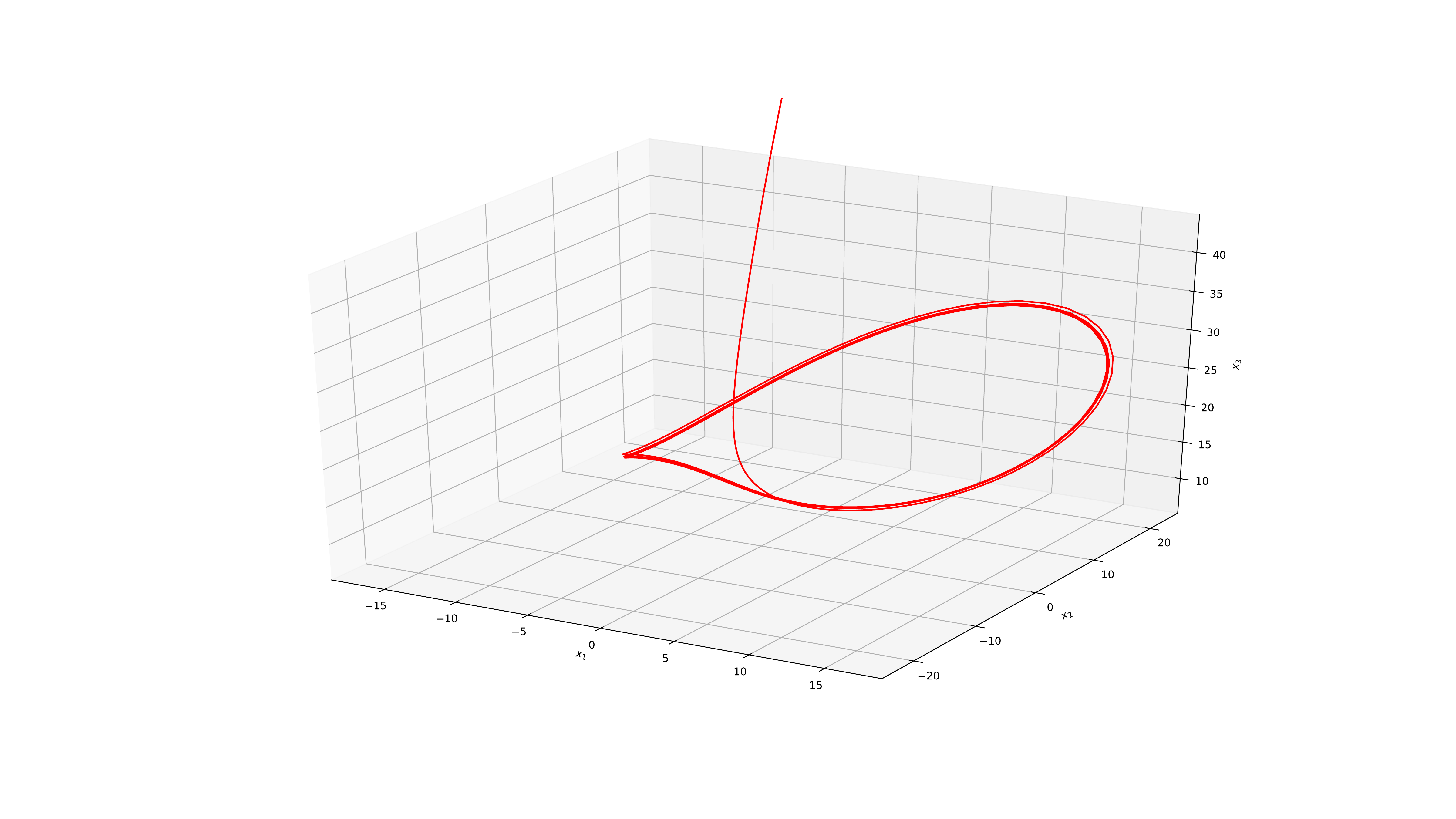}
	\end{subfigure}%
	
	\begin{subfigure}[b]{0.04\linewidth}
	    \rotatebox[origin=t]{90}{\scriptsize MAP}\vspace{1.0\linewidth}
	\end{subfigure}%
	\begin{subfigure}[t]{0.96\linewidth}
	    \centering
		\includegraphics[width=\cwidth,clip, trim=100mm 40mm 80mm 40mm]{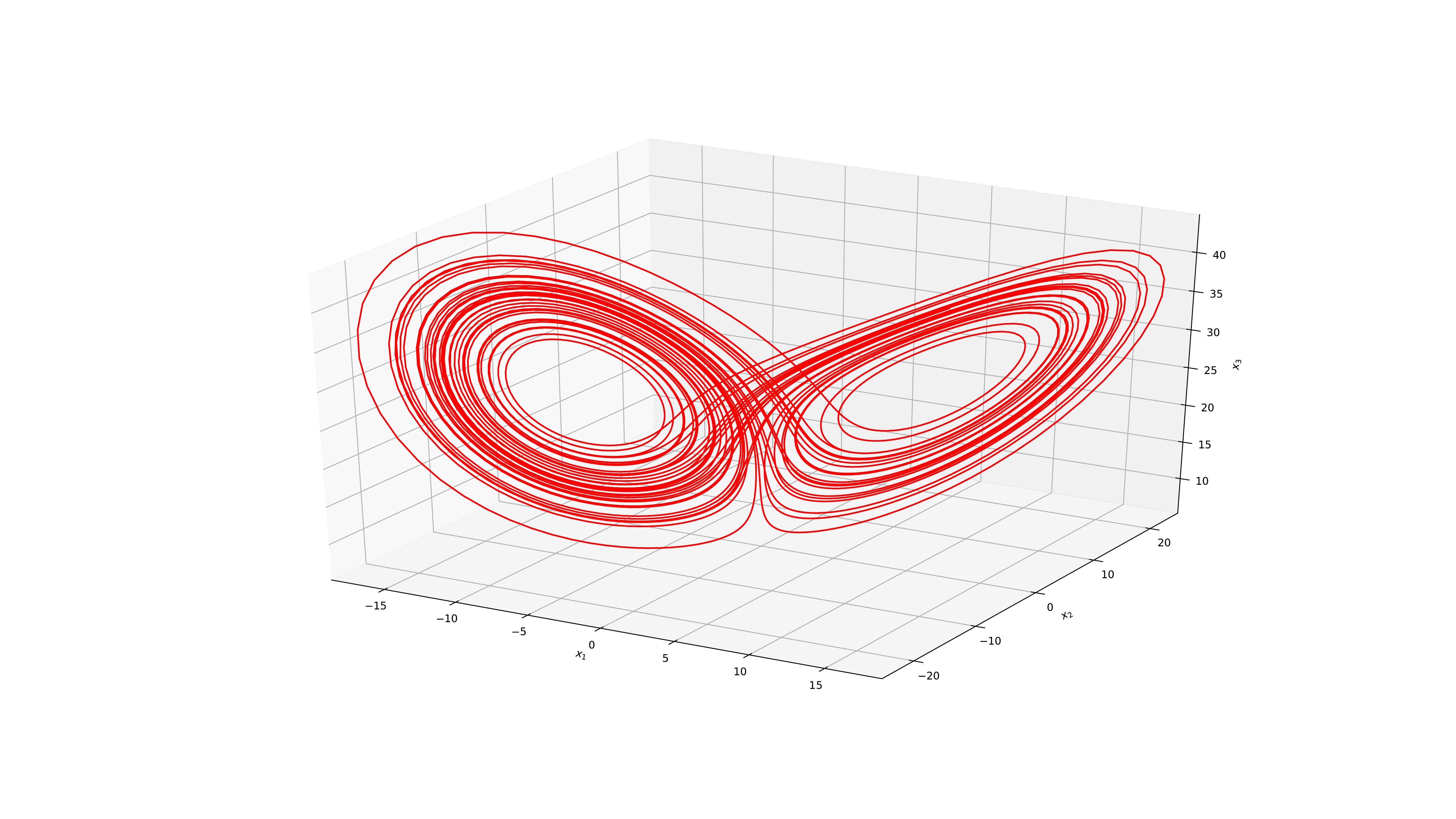}
		\includegraphics[width=\cwidth,clip, trim=100mm 40mm 80mm 40mm]{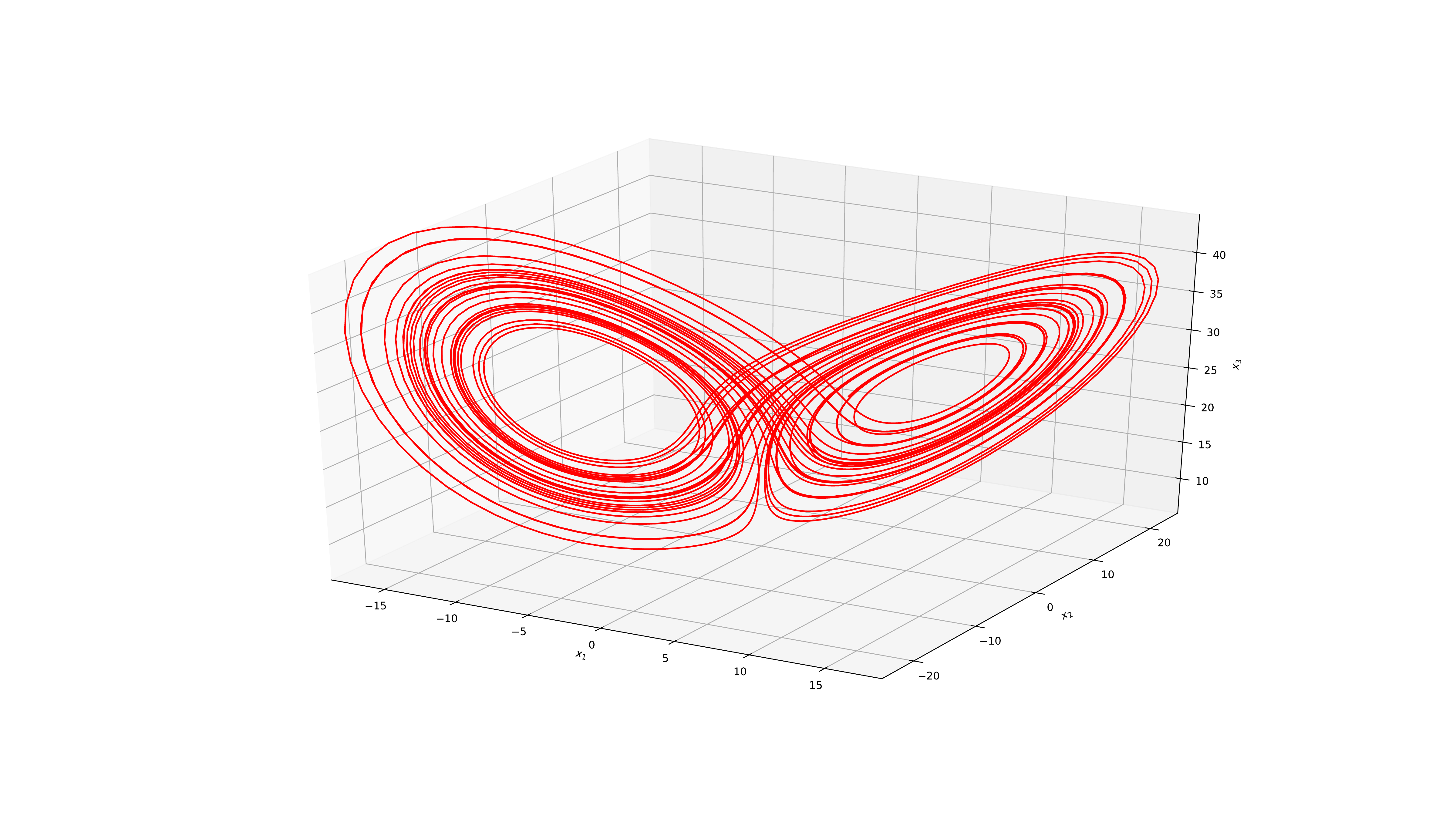}
		\includegraphics[width=\cwidth,clip, trim=100mm 40mm 80mm 40mm]{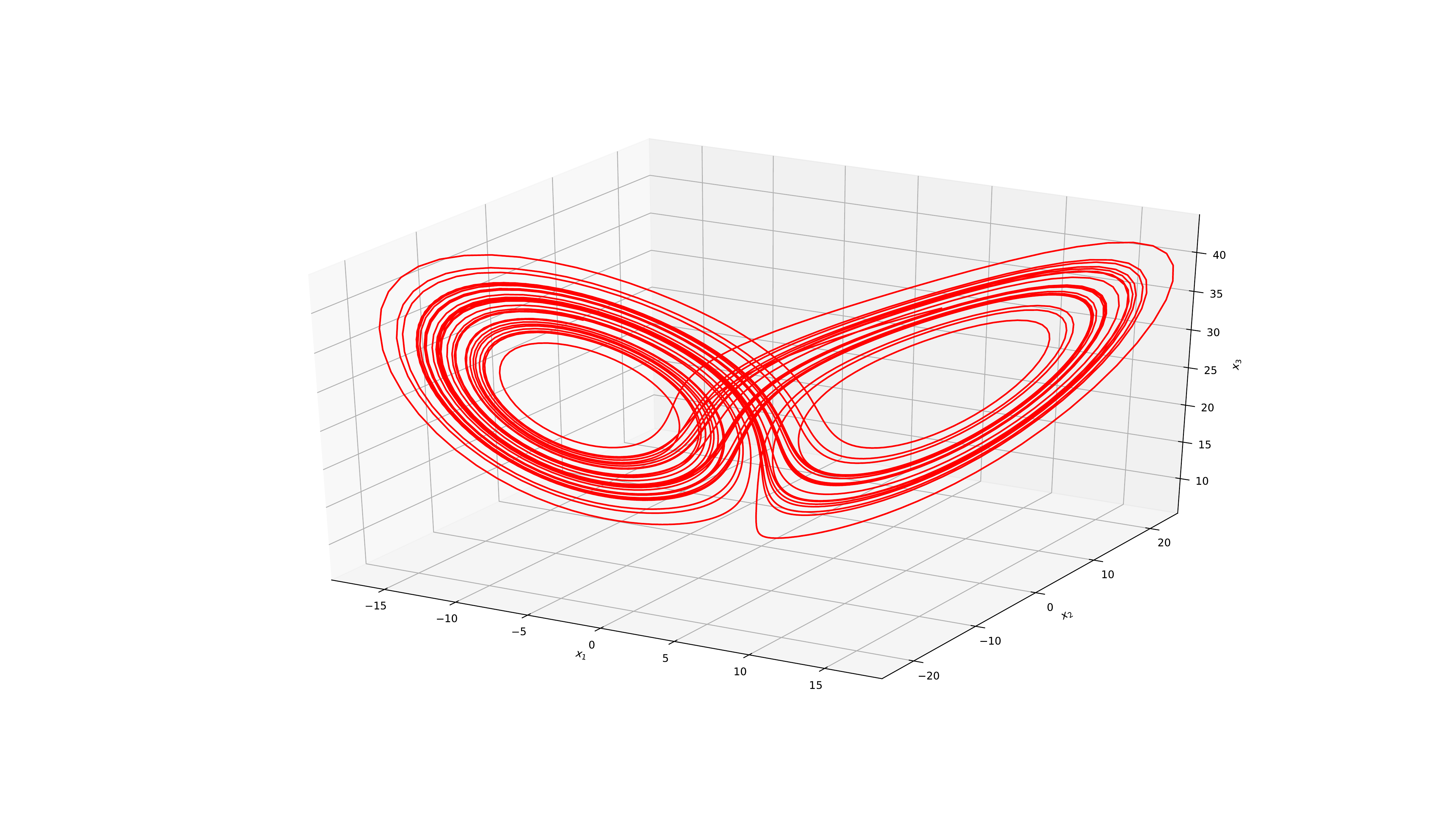}
		\includegraphics[width=\cwidth,clip, trim=100mm 40mm 80mm 40mm]{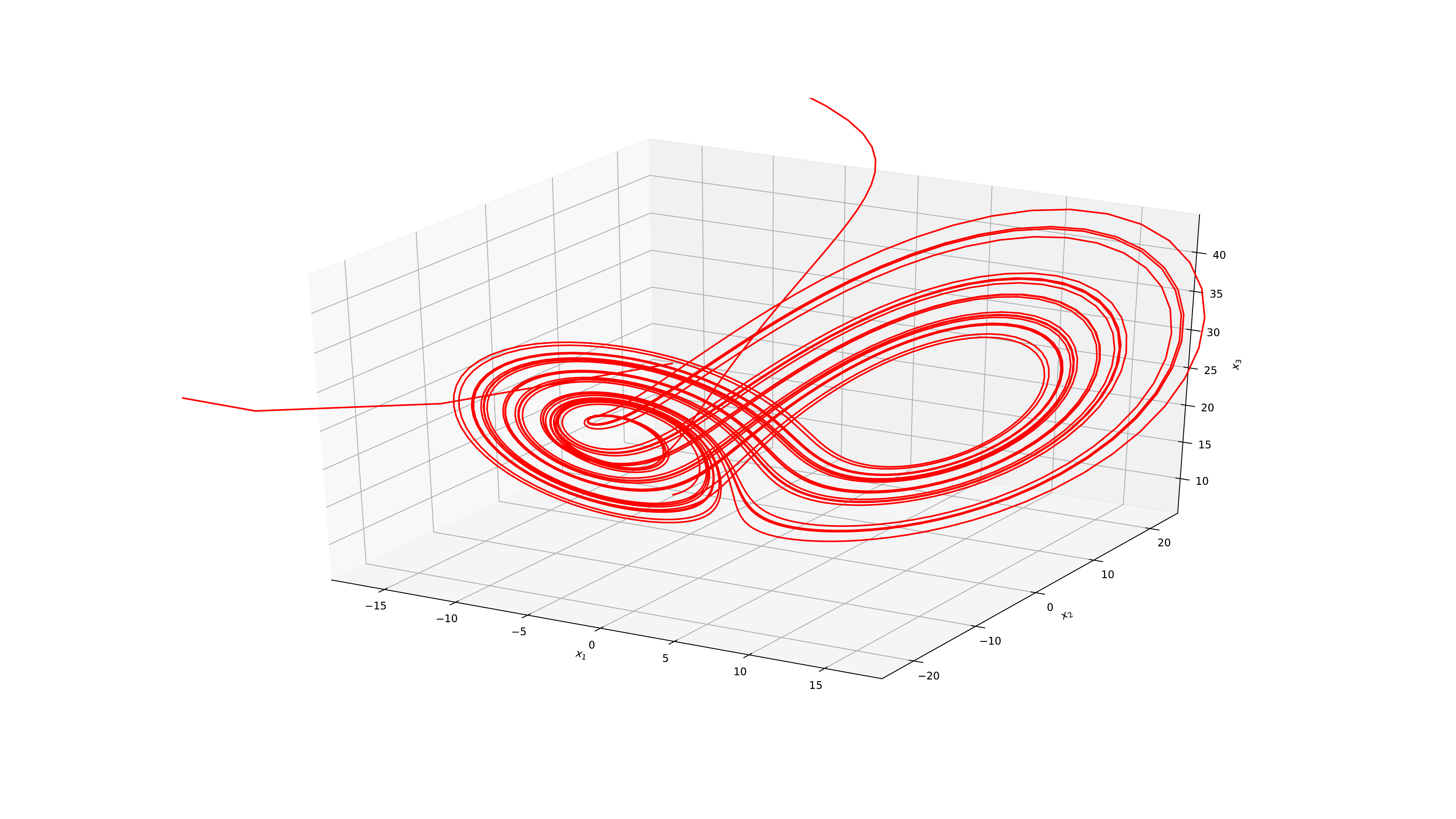}
	\end{subfigure}%
	
	\begin{subfigure}[b]{0.04\linewidth}
	    \rotatebox[origin=t]{90}{\scriptsize Full}\vspace{1.0\linewidth}
	\end{subfigure}%
	\begin{subfigure}[t]{0.96\linewidth}
	    \centering
		\includegraphics[width=\cwidth,clip, trim=100mm 40mm 80mm 40mm]{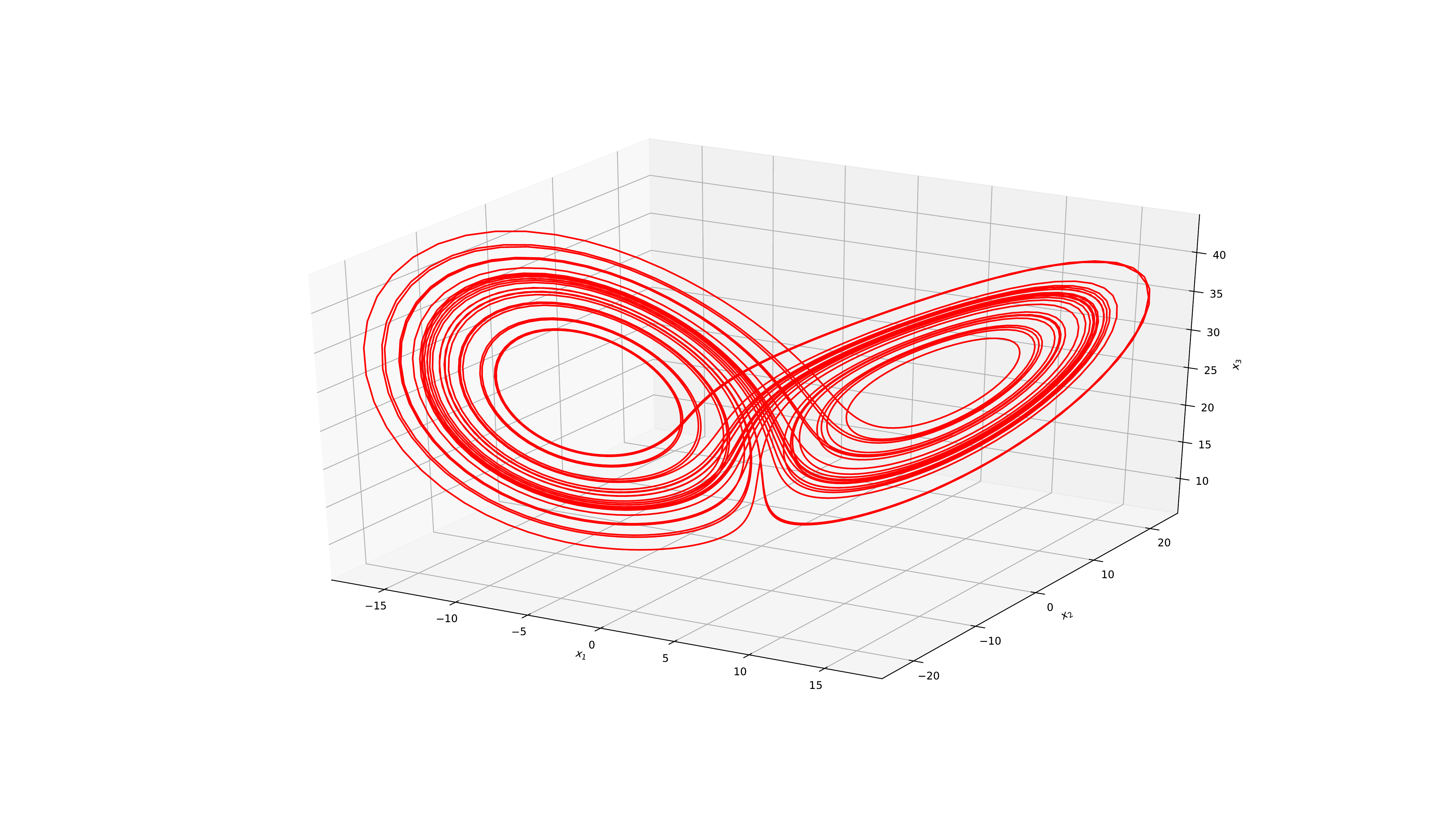}
		\includegraphics[width=\cwidth,clip, trim=100mm 40mm 80mm 40mm]{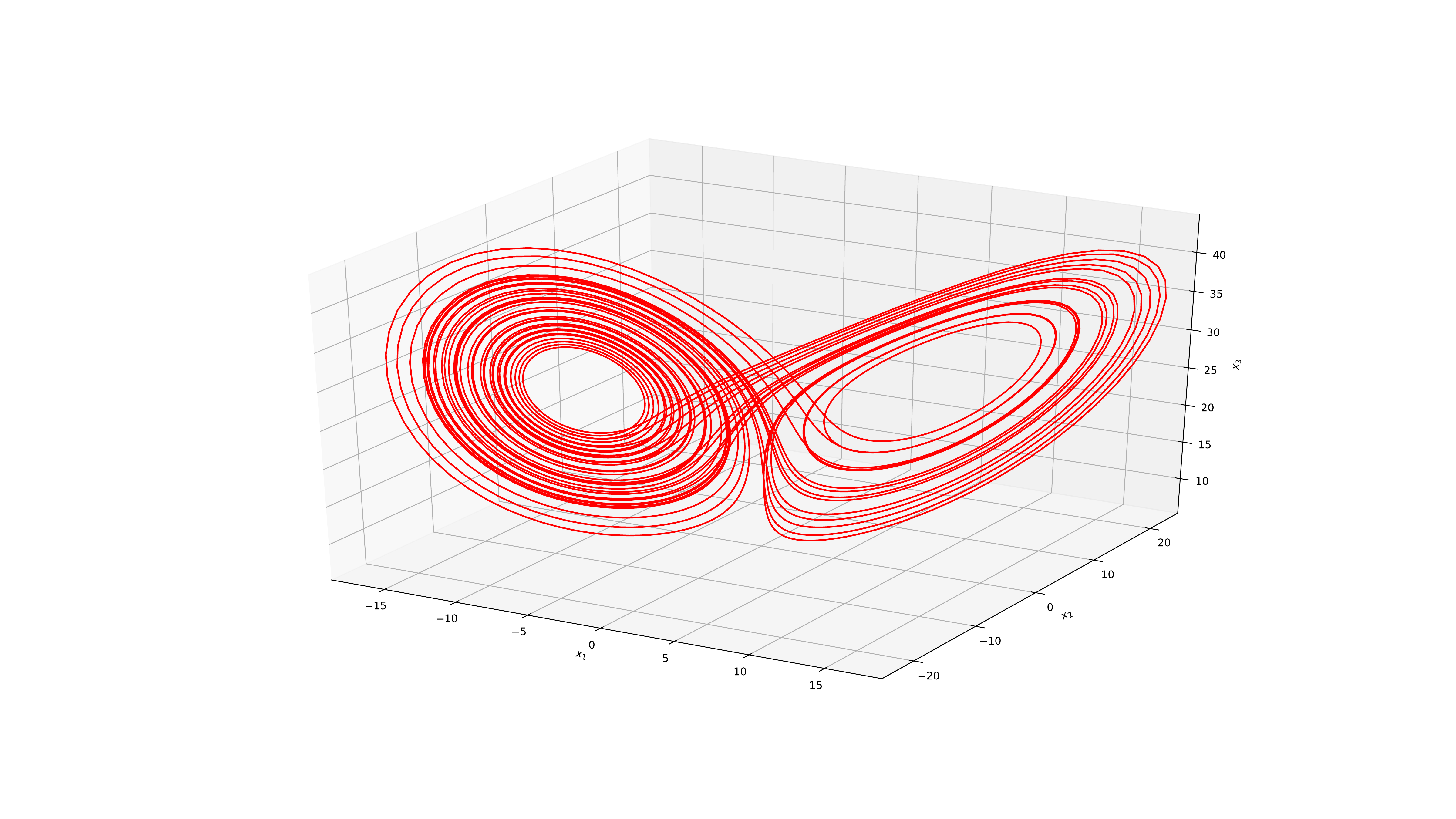}
		\includegraphics[width=\cwidth,clip, trim=100mm 40mm 80mm 40mm]{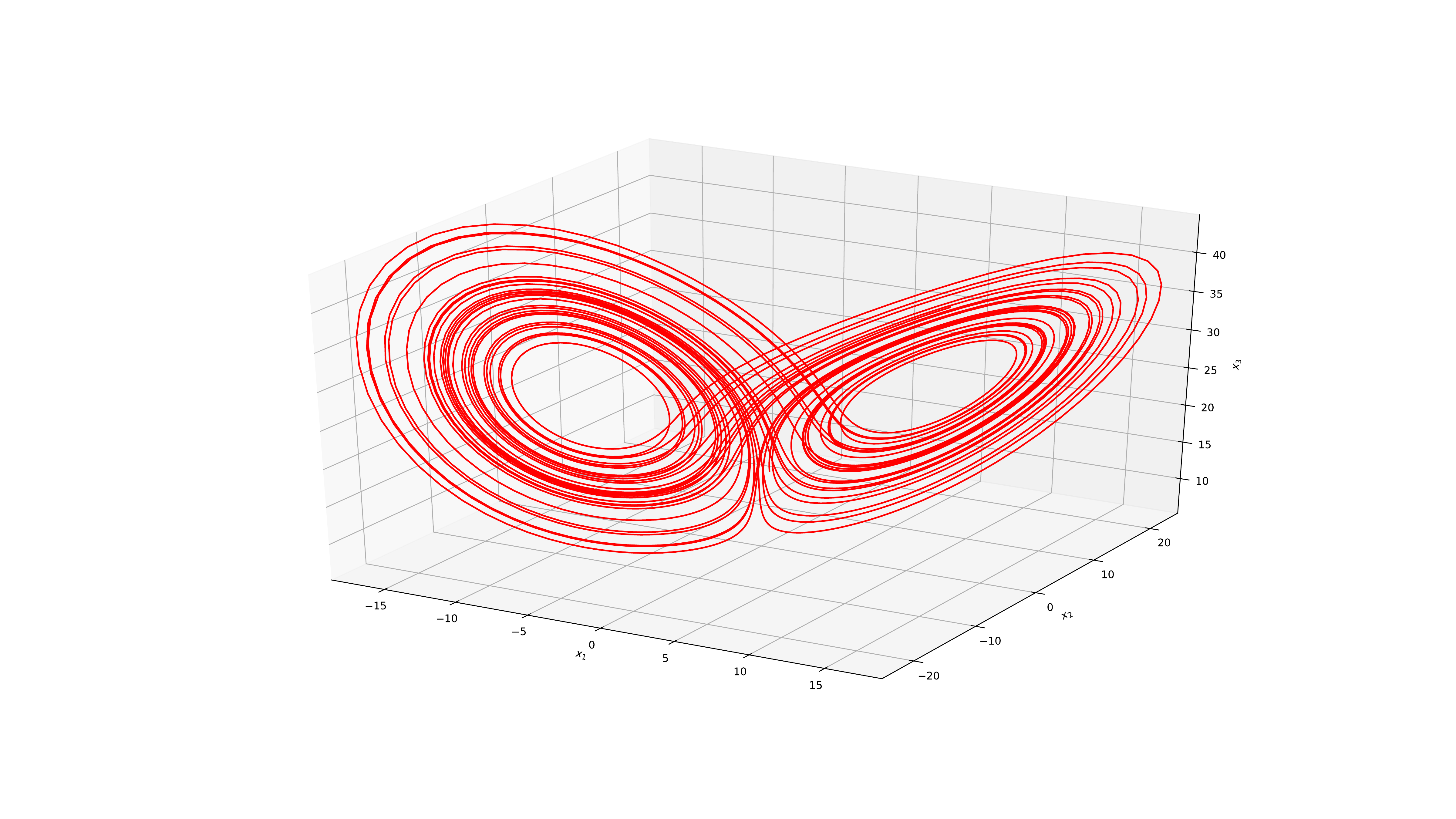}
		\includegraphics[width=\cwidth,clip, trim=100mm 40mm 80mm 40mm]{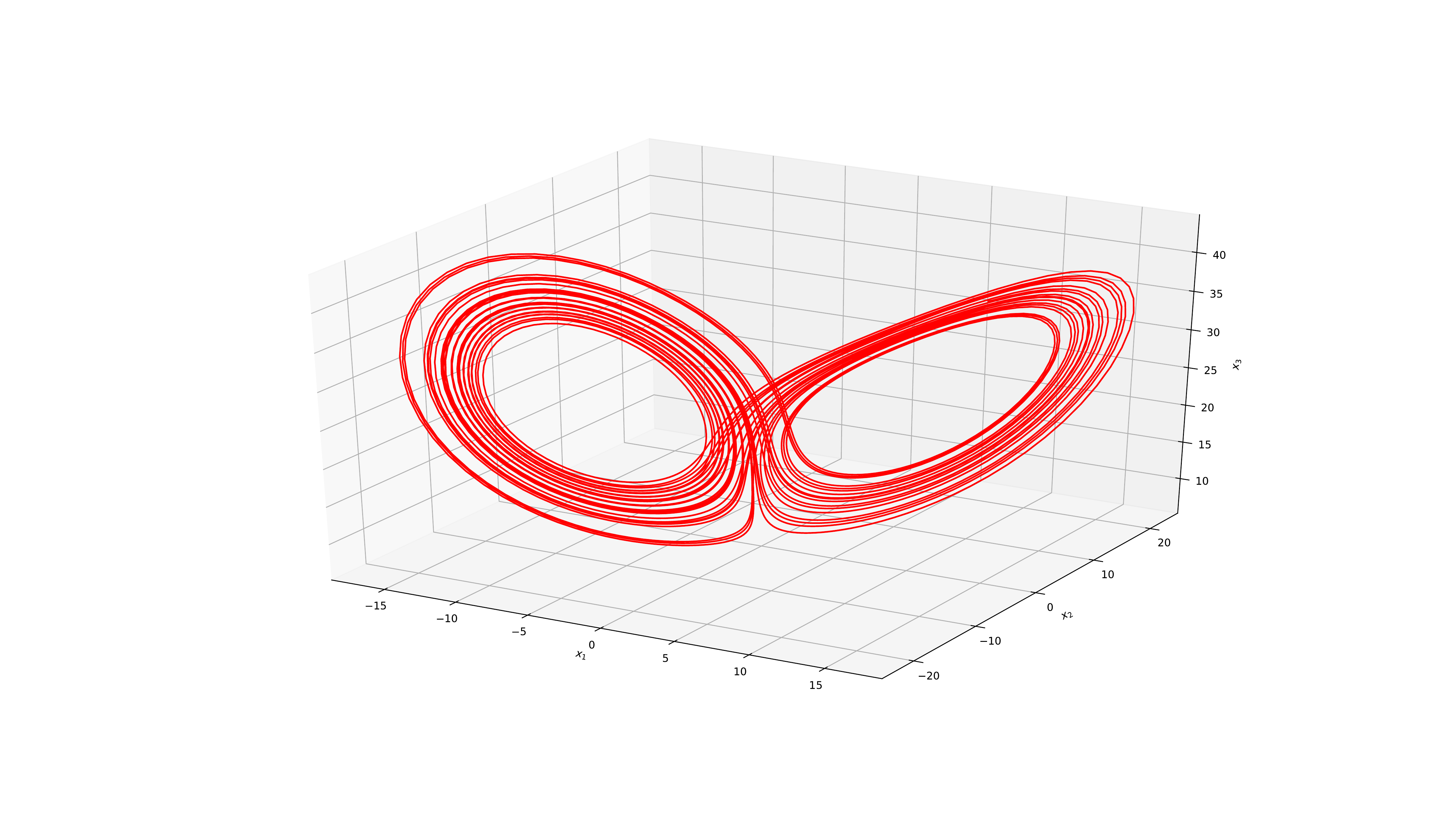}
	\end{subfigure}%
	\caption{Attractors generated by models trained on noisy and partially observed data.}
    \label{fig:attactors_lorenz63_partial}
\end{figure}




\subsection{L96 case-study}
\label{sec:lorenz63_partial}
\begin{figure}
    \centering
    \includegraphics[width=\linewidth,clip, trim=3mm 0mm 23mm 0mm]{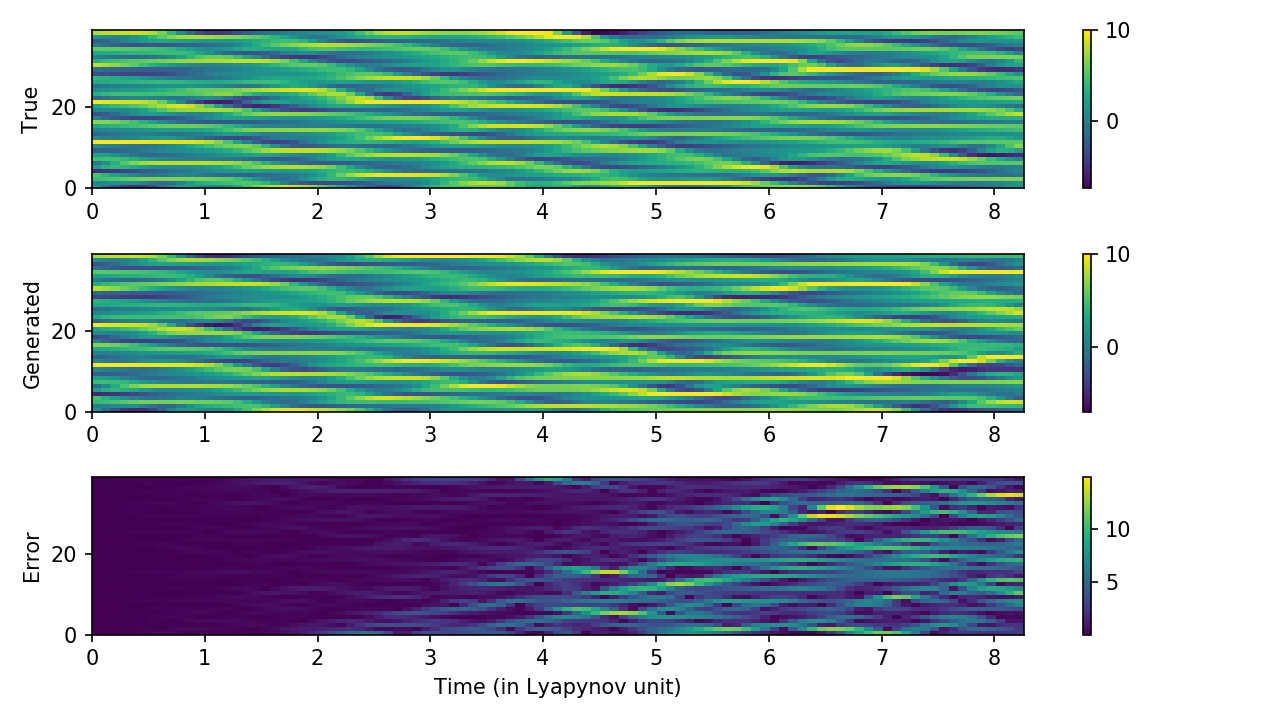}
    \caption{The true L96 sequence (top), the sequence generated by the model trained on noisy data with $r=19.4\%$ (middle) and the error between the true and the generated sequence (bot).}
    \label{fig:lorenz96}
\end{figure}

In this section we present experiments on a L96 system. The objective is to assess how the proposed framework applies in higher-dimensional spaces. We choose the deterministic and the full version of DAODEN as the candidate models. The results of models trained on noisy observations are shown in Table. \ref{tab:L96}. DAODEN models outperform state-of-the-art methods both in terms of short-term prediction and long-term topology. In Fig. \ref{fig:lorenz96} we show the error between the true sequence and the sequence generated by the DAODEN\_determ learnt on noisy observation with $r=19.4\%$. Both sequences have the same starting point. 

\begingroup
\begin{table}[]
    \caption {Performance of models trained on noisy L96 data.  For each index, the best score is marked in \textbf{bold}.}
    \label{tab:L96}
    \footnotesize
    \centering
    \begin{tabular}{ll*{2}c}
    \toprule
    \multicolumn{2}{c}{\multirow{2}{*}{Model}} & \multicolumn{2}{c}{$r$} \\
    & & 19.4\% & 38.8\% \\
    \midrule 

    \midrule 
    \multirow{3}{*}{AnDA}
    & $e_4$      & 0.582±0.106  & 1.140±0.174 \\
    & $\pi_{0.5}$      & 1.491±0.481  & 0.768±0.281 \\
    &  $\lambda_1$  & 53.362±0.734 &  92.733±0.883 \\

    \midrule 
    \multirow{3}{*}{SINDy}
    & $e4$      & 0.309±0.048  & 0.767±0.117 \\
    & $\pi_{0.5}$      & 0.628±0.166  & 0.150±0.047 \\
    &  $\lambda_1$  & 1.444±0.048 &  1.316±0.045 \\

    \midrule 
    \multirow{3}{*}{BiNN}
    & $e_4$      & 0.310±0.046  & 0.788±0.112 \\
    & $\pi_{0.5}$      & 2.503±0.565  & 1.111±0.274 \\
    &  $\lambda_1$  & 1.409±0.019 &  1.041±0.016 \\    
    
    \midrule 
    \multirow{3}{*}{DAODEN\_determ}
    & $e_4$      & \textbf{0.048±0.006}  & 0.157±0.022 \\
    & $\pi_{0.5}$      & \textbf{4.790±0.960}  & \textbf{3.178±0.779} \\
    &  $\lambda_1$  & 1.624±0.022 &  1.601±0.023 \\

    \midrule 
    \multirow{3}{*}{DAODEN\_full}
    & $e_4$      & 0.067±0.014  & \textbf{0.145±0.030} \\
    & $\pi_{0.5}$      & 4.076±1.084  & 3.146±0.962 \\
    &  $\lambda_1$  & 1.543±0.026 &  1.348±0.020 \\
    
    \bottomrule
    \end{tabular}
    \normalsize
\end{table}
\endgroup


{}

\subsection{L63s case-study}
\label{subsec:lorenz63s}
\begingroup
\begin{figure}
    \centering
	\begin{subfigure}[b]{0.04\linewidth}
	    \rotatebox[origin=t]{90}{\scriptsize True}\vspace{0.9\linewidth}
	\end{subfigure}%
	\begin{subfigure}[t]{0.96\linewidth}
	    \centering
		\includegraphics[width=\cwidth,clip, trim=70mm 10mm 40mm 10mm]{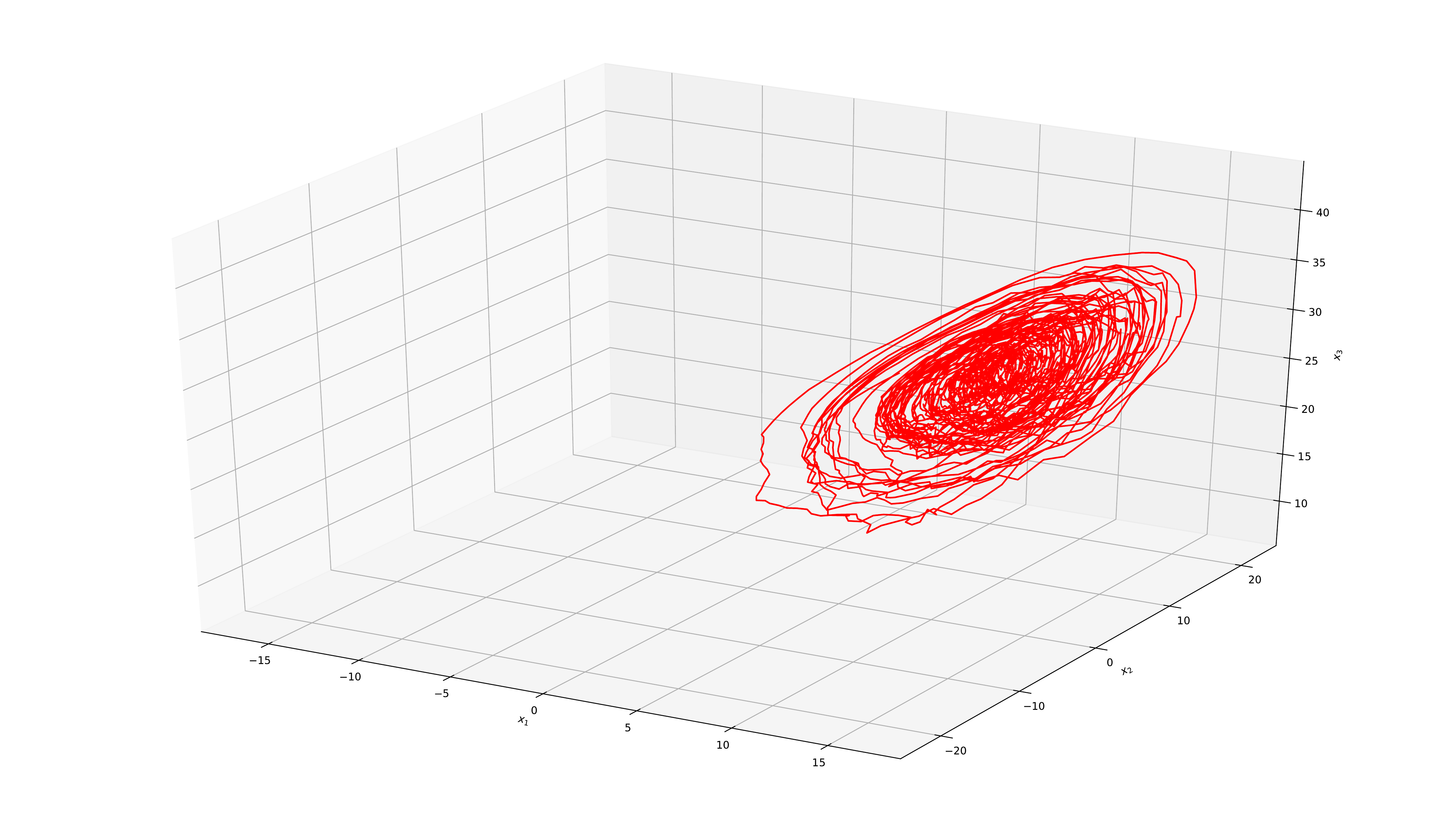}
		\includegraphics[width=\cwidth,clip, trim=70mm 10mm 40mm 10mm]{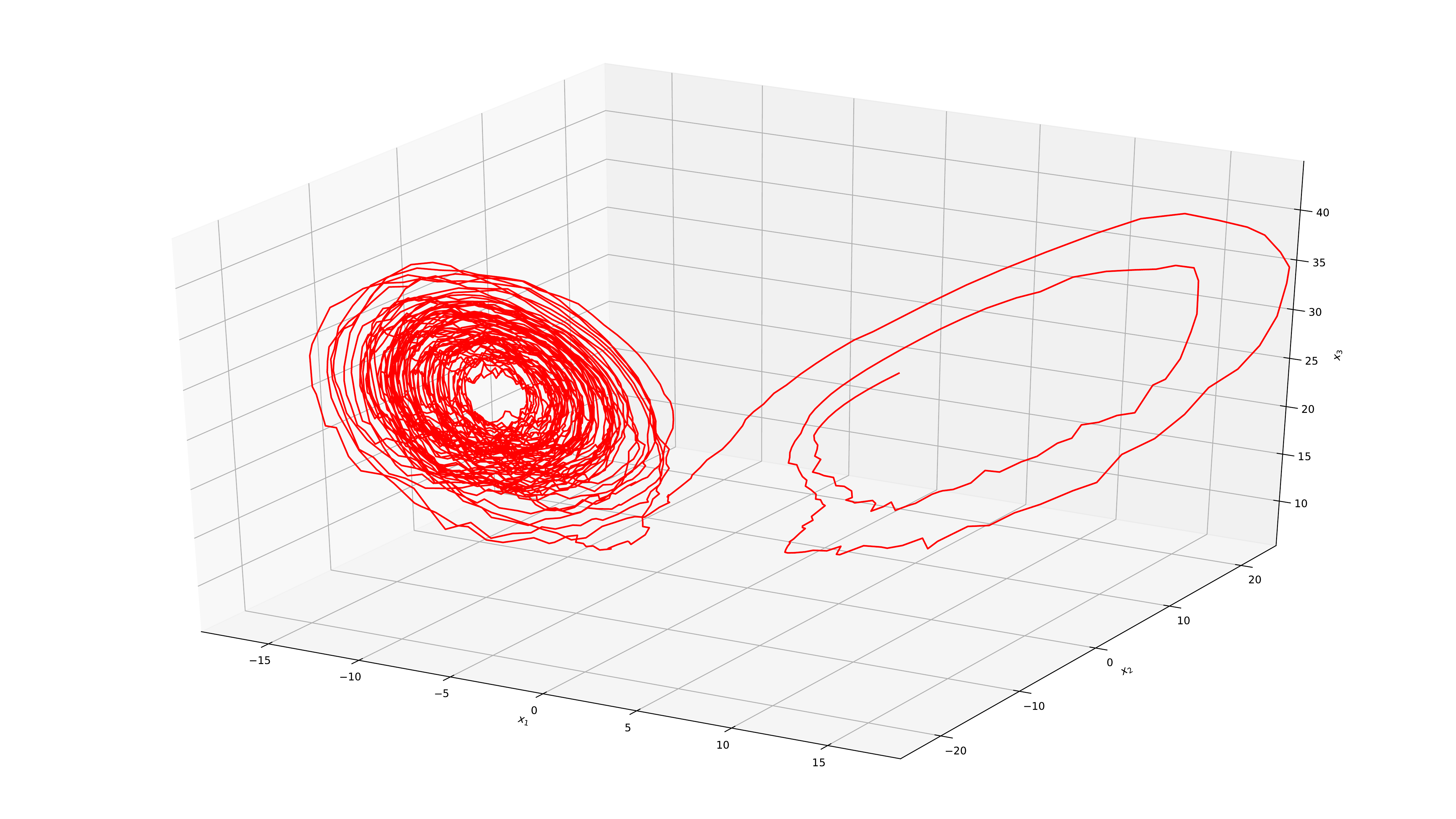}
		\includegraphics[width=\cwidth,clip, trim=70mm 10mm 40mm 10mm]{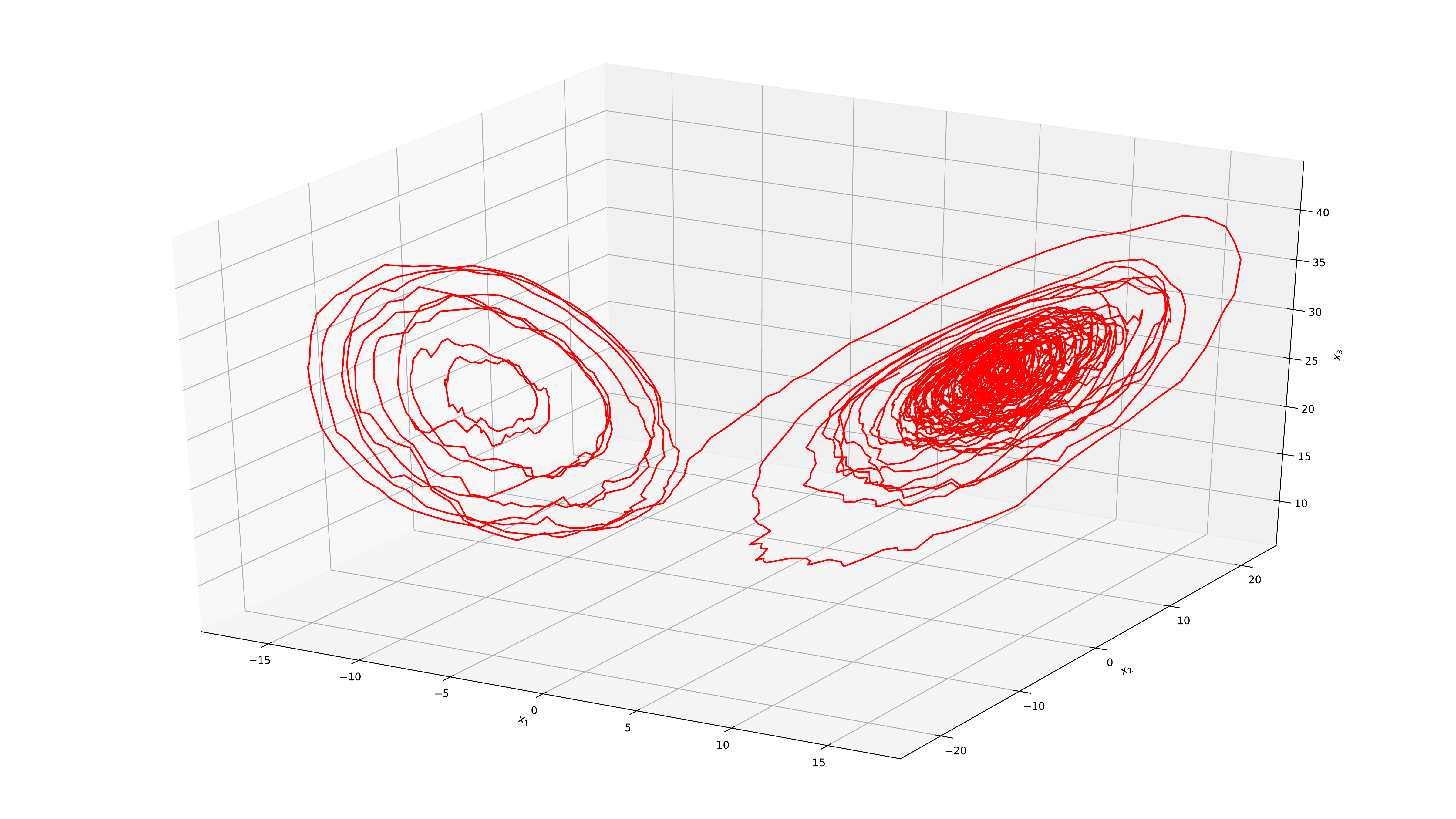}
		\includegraphics[width=\cwidth,clip, trim=70mm 10mm 40mm 10mm]{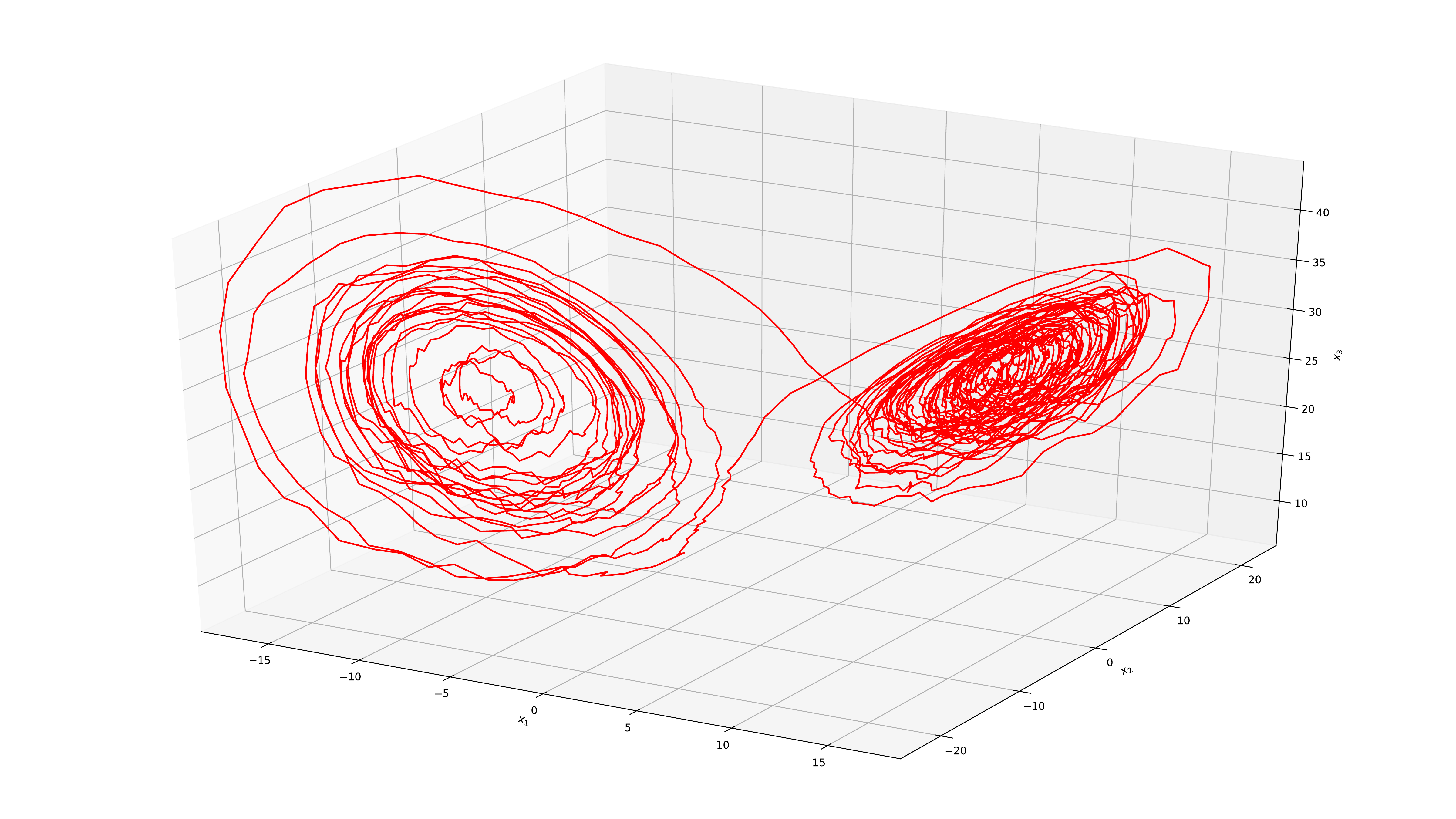}
	\end{subfigure}%

	\begin{subfigure}[b]{0.04\linewidth}
	    \rotatebox[origin=t]{90}{\scriptsize Determ}\vspace{0.8\linewidth}
	\end{subfigure}%
	\begin{subfigure}[t]{0.96\linewidth}
	    \centering
		\includegraphics[width=\cwidth,clip, trim=70mm 10mm 40mm 10mm]{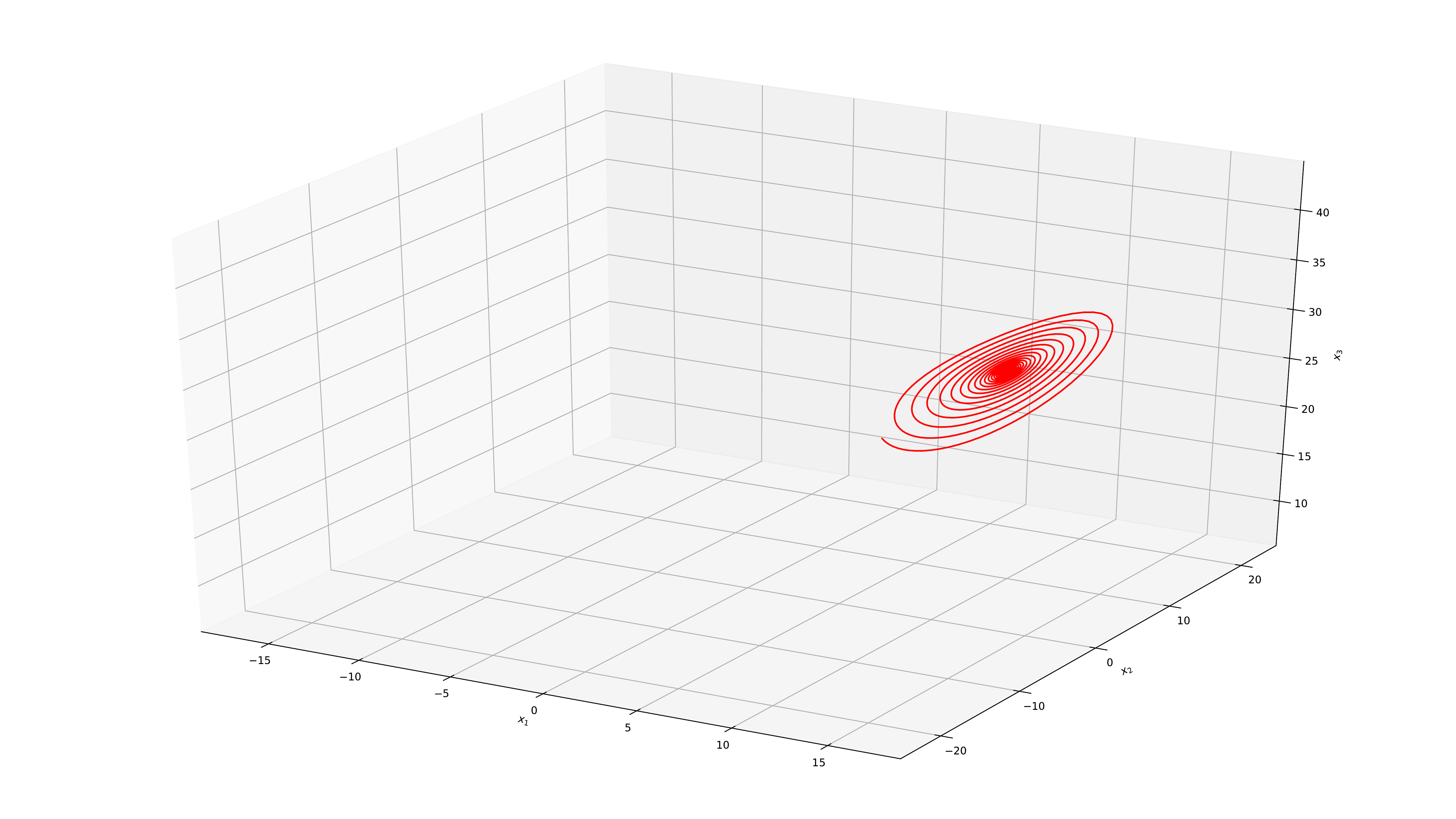}
		\includegraphics[width=\cwidth,clip, trim=70mm 10mm 40mm 10mm]{figures/lorenz63_stoch/L63s_determ.pdf}
		\includegraphics[width=\cwidth,clip, trim=70mm 10mm 40mm 10mm]{figures/lorenz63_stoch/L63s_determ.pdf}
		\includegraphics[width=\cwidth,clip, trim=70mm 10mm 40mm 10mm]{figures/lorenz63_stoch/L63s_determ.pdf}
	\end{subfigure}%
	
	\begin{subfigure}[b]{0.04\linewidth}
	    \rotatebox[origin=t]{90}{\scriptsize Full}\vspace{1.5\linewidth}
	\end{subfigure}%
	\begin{subfigure}[t]{0.96\linewidth}
	    \centering
		\includegraphics[width=\cwidth,clip, trim=70mm 10mm 40mm 10mm]{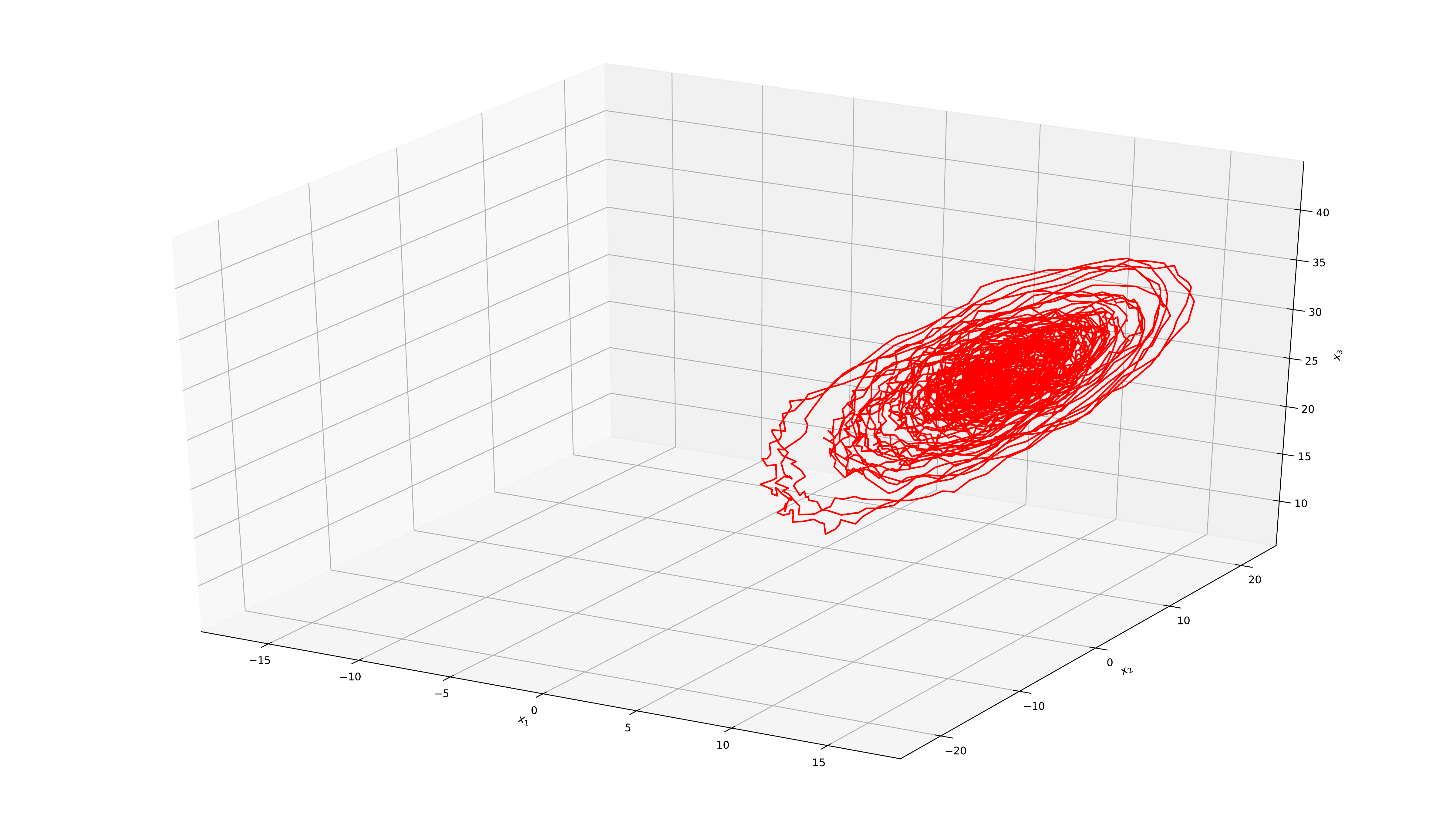}
		\includegraphics[width=\cwidth,clip, trim=70mm 10mm 40mm 10mm]{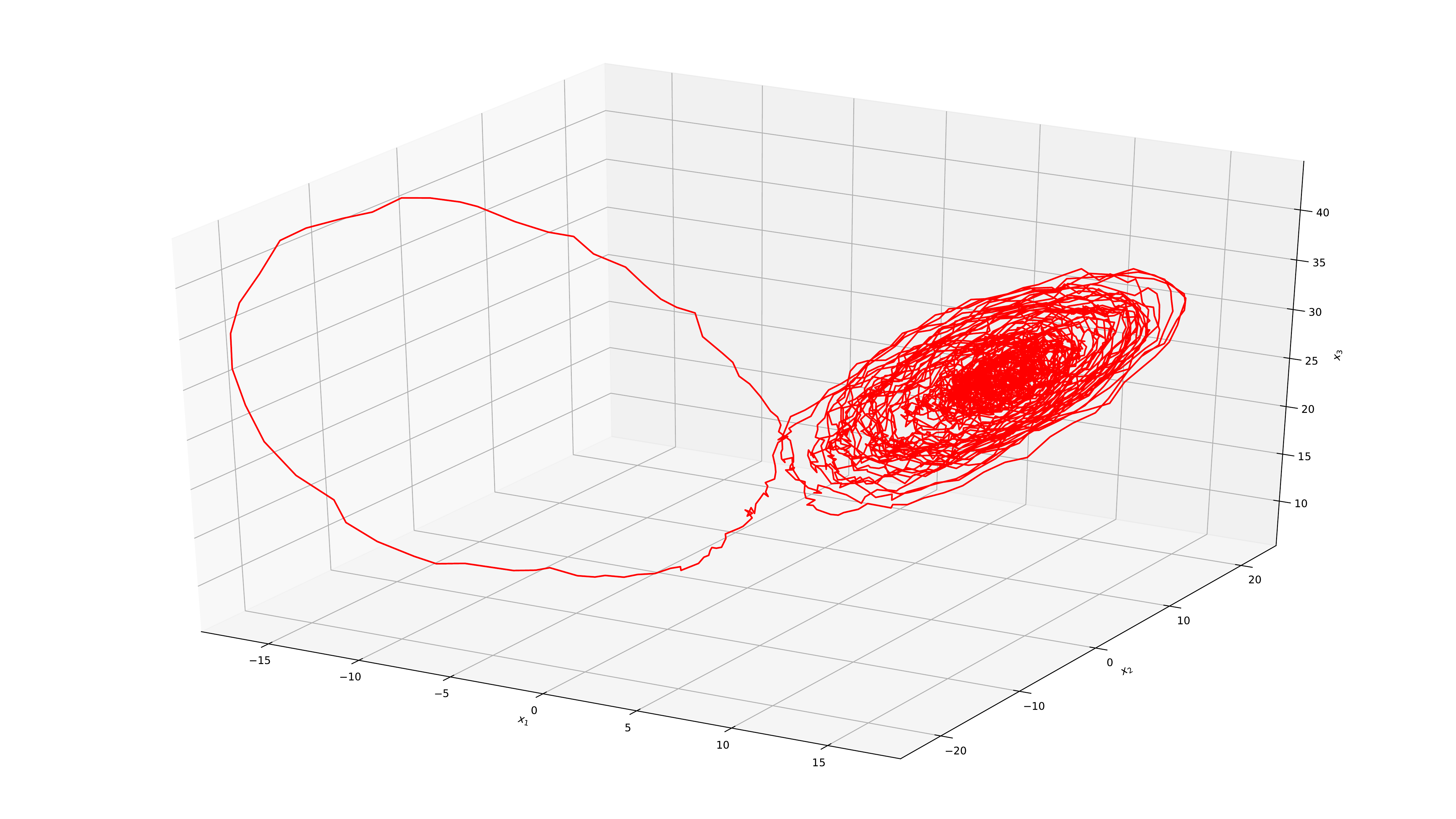}
		\includegraphics[width=\cwidth,clip, trim=70mm 10mm 40mm 10mm]{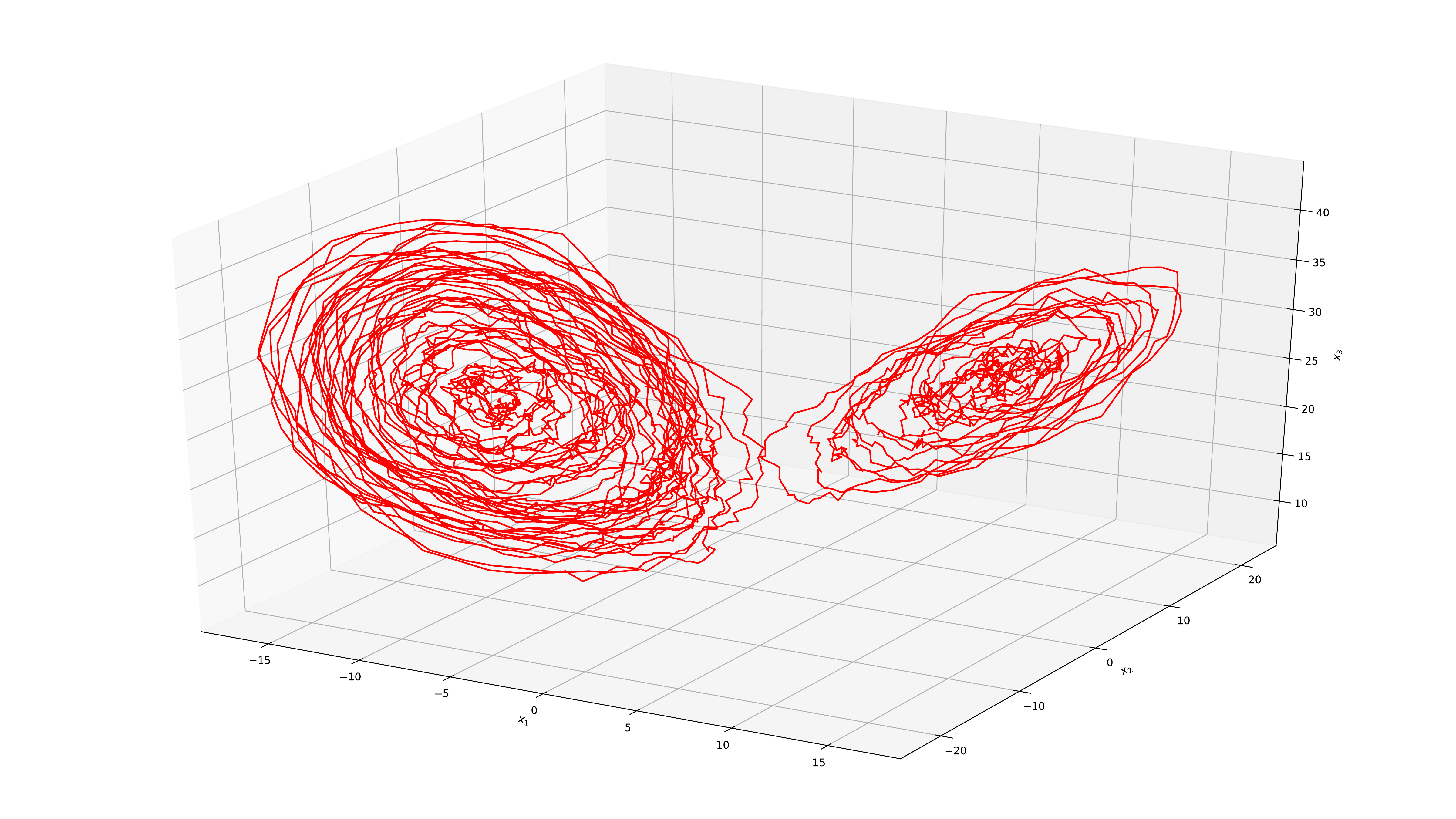}
		\includegraphics[width=\cwidth,clip, trim=70mm 10mm 40mm 10mm]{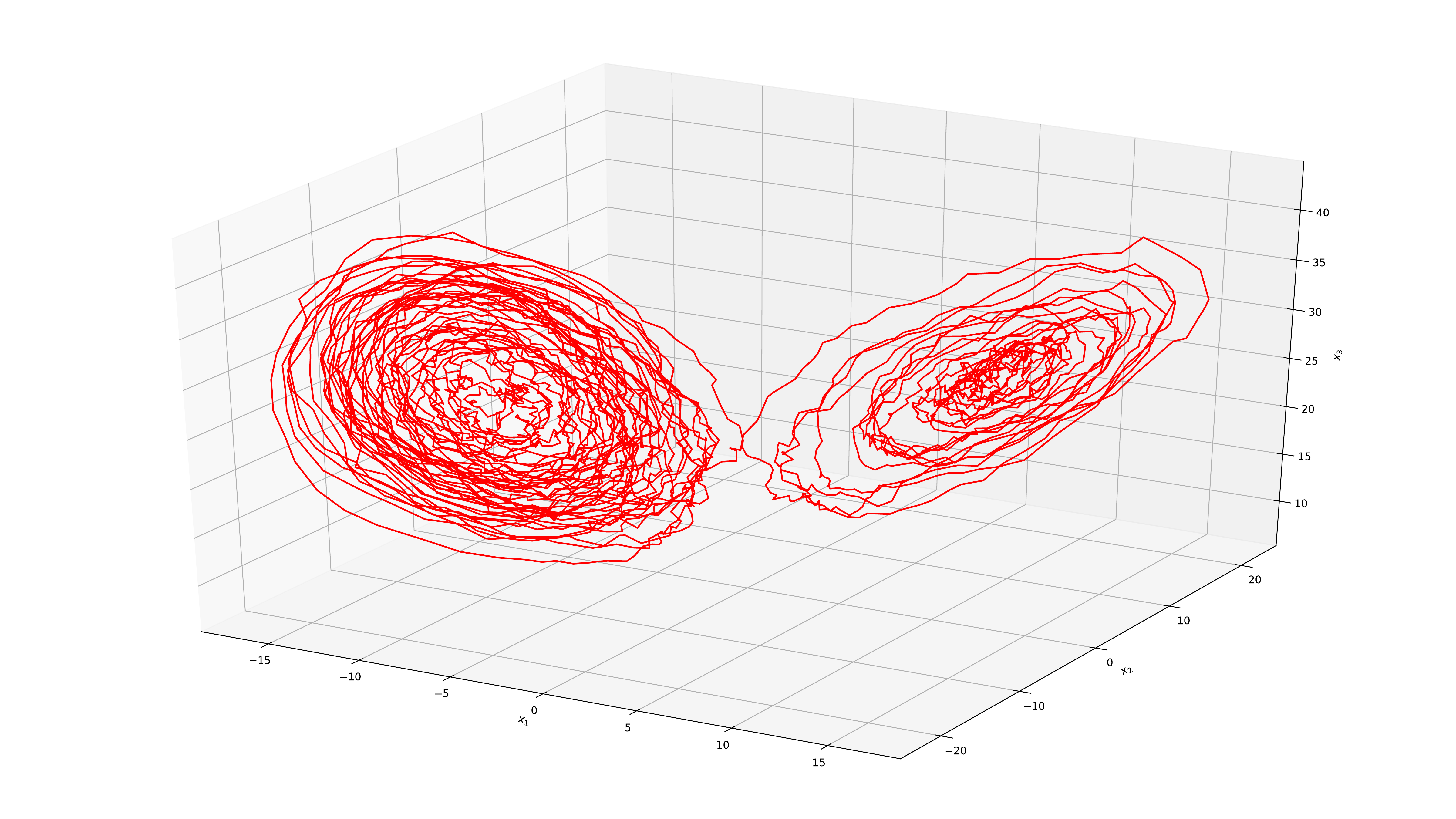}
	\end{subfigure}%
	\caption{Several attractors generated by the true L63s models (top), by DAODEN\_determ (middle)  and by DAODEN\_full (bottom). The true L63s and DAODEN\_full system are stochastic, hence each runtime we obtain a different sequence, even with the same initial condition. The models were trained on noisy observations with $r=33.3\%$.}
    \label{fig:attactors_l63s}
\end{figure}
\endgroup

Whereas most related work is designed for ODE only, (\ie the governing equations are deterministic), the proposed framework accounts for stochastic perturbations, hence it can apply to Stochastic Differential Systems (SDEs).
Using the stochastic Lorenz-63 system (L63s) presented in \cite{chapron_large-scale_2018}, we illustrate in this experiment the ability of DAODEN\_full scheme to infer stochastic governing equations from noisy observation data. We may recall that DAODEN\_full scheme embeds a parametric form of the covariance of perturbation $\vect{\omega}_k$ given by (\ref{eq:dyn_x}). Note that this parametrisation is consistent with the true parametrisation for L63s \cite{chapron_large-scale_2018}. 

Here, we ran experiments similar to those in Section \ref{subsec:lorenz63_noisy} using L63s datasets with an additive Gaussian noise with $r = 33.3\%$. We then ran the identification of governing equations using both a deterministic parametrisation (\eg, BiNN\_EnKS and DAODEN\_determ) and the fully-stochastic scheme DAODEN\_full. For weak stochastic perturbations, (typically, $\gamma$ larger than 8.0 in Eq. (A.3) in Appendix A), deterministic models like BiNN\_EnKS or DAODEN\_determ can still be able to capture the dynamics of the system (not reported in this paper). However, when $\vect{\omega}_k$ plays an important role in controlling the large-scale statistical characteristics of the system, deterministic models fail, as illustrated in Fig. \ref{fig:attactors_l63s} for L63s dynamics with $\gamma = 5.0$. By contrast, the fully-stochastic model successfully uncovers the stochastic dynamics in both situations. In Fig. \ref{fig:attactors_l63s} top, we depict four different L63s trajectories from the same initial conditions. Due to the stochastic perturbation, the trajectories may strongly differ but all show a wide spreadout within the attractor. When considering a deterministic model (Fig. \ref{fig:attactors_l63s} middle), the four trajectories are strictly similar as there is no stochastic perturbation. Besides, the deterministic model simulates trajectories trapped on one side of the attractor, which cannot reproduce the spread of the true model. As illustrated in Fig. \ref{fig:attactors_l63s} bottom, DAODEN\_full scheme succeed in capturing this stochastic patterns by embedding the stochastic factors of the system in the dispersion matrix $\mathbf{Q}_k$. Using a Monte Carlo technique, as presented in Alg. 1 in Appendix C, to forecast the state of the dynamics, we can obtain sequences with similar characteristics to the true L63s system.

\subsection{Dealing with an unknown observation operator}
\label{subsec:legendre}


In previous experiments, the observation operator $\mathcal{H}$ was known. We may also address the situation where it is unknown. It may for instance refer to reduced-order modelling \cite{lucia_reduced-order_2004}, when one looks for a lower-dimensional representation of a higher-dimensional dynamical system. 

\begin{figure}
    \centering
    \includegraphics[width=\linewidth, clip, trim=0mm 0mm 0mm 0mm]{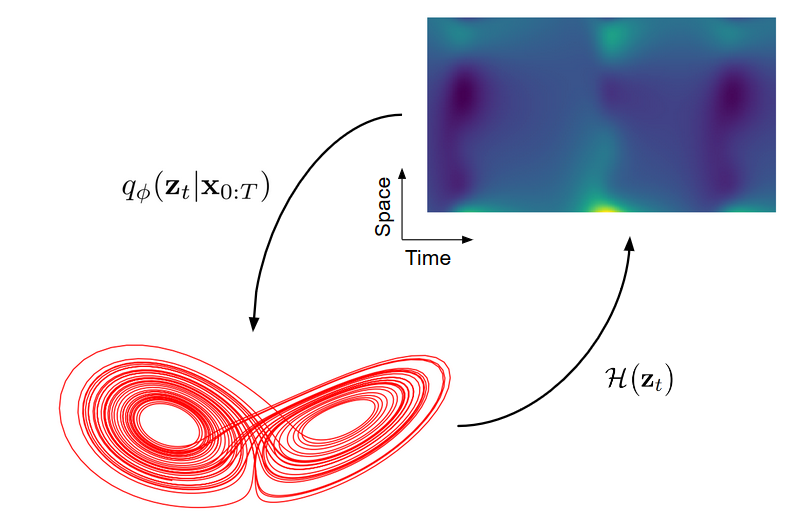}
    \caption{Higher-dimensional Legendre observations governed by lower-dimensional L63 dynamics. Following \cite{champion_data-driven_2019}, the observations (top right) are in a 128-dimensional space, while L63 dynamics (bottom left) are in a 3-dimensional space. The observation operator involves a non-linear mapping according to Legendre polynomials \cite{champion_data-driven_2019}.}
    \label{fig:legendre_obs}
\end{figure}

As case-study, we consider an experimental setting with Lorenz-63 dynamics similar to  \cite{champion_data-driven_2019}. The 128-dimensional observation space derives from a 3-dimensional space, where the system is governed by L63 ODE, according to a polynomial of $\vect{z}_t$ and $\vect{z}^3_t$ with six spatial modes of Legendre coefficients (for details, see \cite{champion_data-driven_2019}). Whereas noise-free cases were considered in \cite{champion_data-driven_2019}, we report here experiments with a Gaussian additive noise with $r$=19.4\%. Fig. \ref{fig:legendre_obs} shows the observations in a high-dimensional space. The inference scheme in \cite{champion_data-driven_2019} is a NN-based encoder, this architecture does not take into account the sequential correlations in the data, hence when the observations are noisy, it can not apply (because $p(\vect{z}_t|\vect{x}_t)$ is intractable). Moreover, \cite{champion_data-driven_2019} supposes that the time derivative $\frac{d\vect{x}_{t}}{dt}$ is observed. This assumption may not be true for many real-life systems. Our model, on the other hand, uses a state-space assimilation formulation. The inference scheme in our model is a sequential model, and we do not need the time derivative of the data, though it could be accounted for in the observation model.

{}

\begin{figure}
    \centering
    \includegraphics[width=\linewidth, clip, trim=20mm 0mm 0mm 20mm]{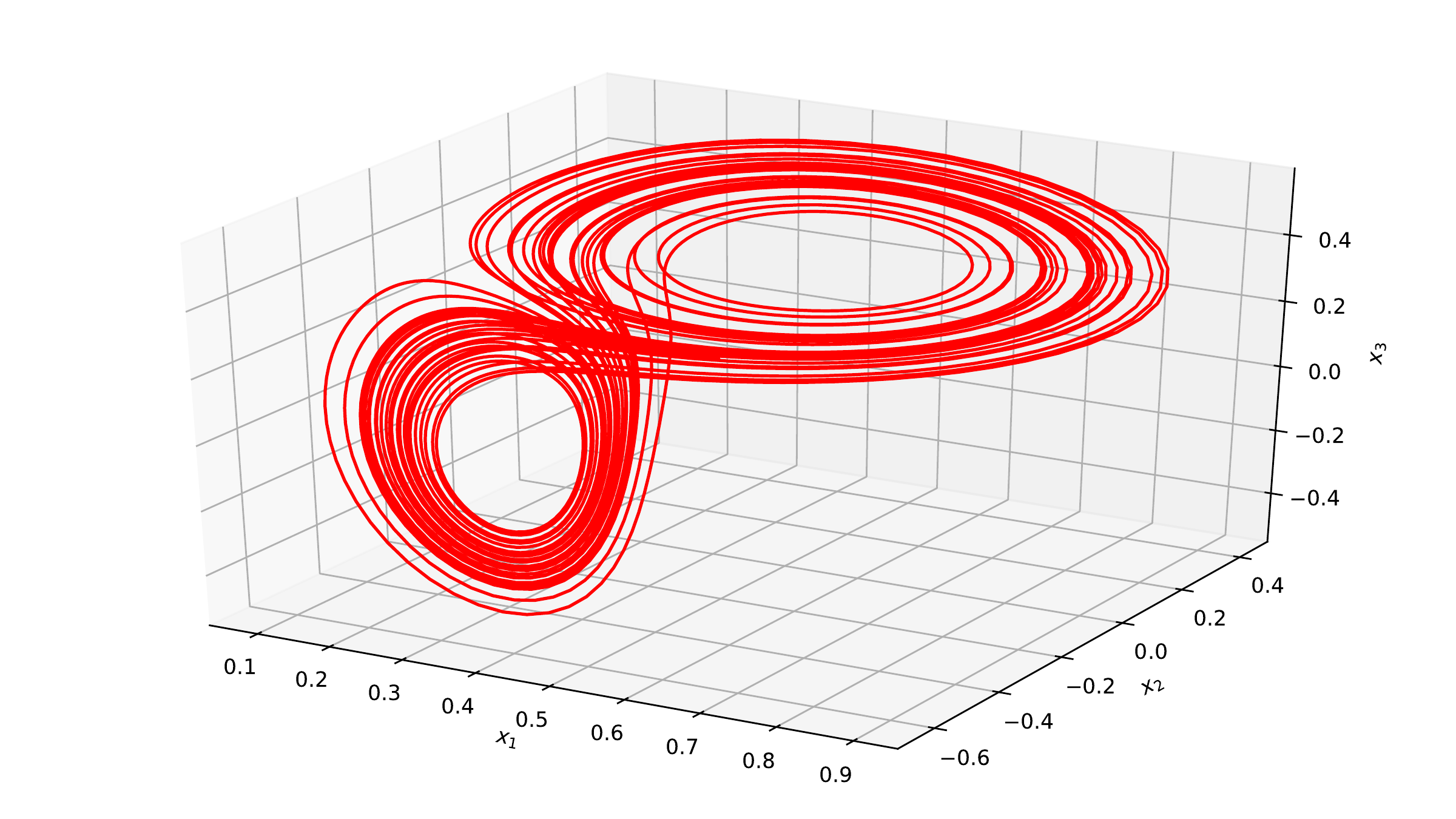}
    \caption{Low-dimensional attractor generated by the proposed model trained from noisy higher-dimensional Legendre observations of L63 dynamics. This attractor recovers the topology of L63 dynamics. We let the reader refer to the main text for details on this experiment.}
    \label{fig:legendre_attractor}
\end{figure}

The unknown observation operator $\mathcal{H}$ was parameterised by the same MLP architecture as the one used in \cite{champion_data-driven_2019}. We run this experiment with DAODEN\_determ. Fig. \ref{fig:legendre_attractor} shows that the proposed framework successfully captures the low-dimensional attractor of the observed high-dimensional observation sequences. This is further supported by the first Lyapunov exponent of the learnt model $\lambda_1 = 0.92$, which is close to the true value (0.91). Because there are several possible solutions for this problem (any affine transformation of the original L63 is a solution), the coordinates of the learnt system are different, however, the topology is well captured. 

{}
\section{Conclusion}
\label{sec:conclusions}


This paper introduces a novel deep learning scheme for the identification of governing equations of a given system from noisy and partial observation series. We combine a Bayesian formulation of the data assimilation with state-of-the-art deep learning architectures. Compared with related work \cite{brajard_combining_2019, bocquet_bayesian_2020}, we account for stochastic dynamics rather than only deterministic ones and derive an end-to-end architecture using a variational deep learning model, which fully conforms to the state-space formulation considered in data assimilation. Through numerical experiments for chaotic and stochastic dynamics, we have demonstrated that we can extend the observation configurations where we can recover hidden governing dynamics from noisy and partial data w.r.t. the state-of-the-art, including for high-dimensional systems governed by lower-dimensional dynamics.  

Beyond the generalisation of previous work through a variational Bayesian formulation, 
the proposed framework involves two key contributions w.r.t. state-of-the-art data assimilation schemes. We first show that neural network architectures bring a new means for the parametrisation of both the dynamical model and the inference scheme. Especially, our experiments support the relevance of LSTM-based architectures as alternatives to state-of-the-art data assimilation schemes such as Ensemble Kalman methods \cite{evensen_ensemble_2000}.  
Future work shall further explore these aspects and could benefit from the resulting end-to-end architecture to improve reconstruction performance \cite{fablet_joint_2020}. For deep learning practitioners, our experiments point out that assimilation schemes and random $n$-step-ahead forecasting can be considered as regularisation techniques to prevent overfitting. 
We have also shown that the stochastic implementation of the proposed framework can capture characteristics of stochastic dynamical systems from noisy data. These results open new research avenues for dealing with real dynamical systems, for which the stochastic perturbations often play a significant role in driving long-term patterns. 

\rv{From a practical point of view, the results showed in this paper suggest that although some models might be able to discover the governing equations of an unknown dynamical system when the data are not corrupted, one should incorporate those models with data assimilation schemes to account for that fact that the model may contain error, and the data are not perfect. Other results also support the use of NN-based method for the identification of dynamical systems.}



\def\url#1{}
\bibliographystyle{IEEEtran}
\footnotesize
\small
\normalsize
\bibliography{Zotero}

\newpage
\maketitle
\appendices
\numberwithin{equation}{section}
\section{Dynamical systems}
\label{sec:benchmark_systems}

\subsection{The Lorenz-63 system}
\label{subsec:L63}

The Lorenz-63 system (L63), named after Edward Lorenz, is a 3-dimensional dynamical system that model the atmospheric convection \cite{lorenz_deterministic_1963}. The L63 is governed by the following ODE:

\begin{equation}
\begin{aligned}
    \label{eq:lorenz-63}
    \frac{\mathrm{d}\vect{z}_{t,1}}{\mathrm{d}t} &=\sigma \left (\vect{z}_{t,2}-\vect{z}_{t,1} \right ) 
    \\
    \frac{\mathrm{d}\vect{z}_{t,2}}{\mathrm{d}t} &= \left(\rho - \vect{z}_{t,3} \right) \vect{z}_{t,1}-\vect{z}_{t,2} 
    \\
    \frac{\mathrm{d}\vect{z}_{t,3}}{\mathrm{d}t} &= \vect{z}_{t,1}\vect{z}_{t,2}-\beta \vect{z}_{t,3} 
\end{aligned}
\end{equation}

\begin{figure}[H]
    \centering
    \includegraphics[width=\linewidth,clip, trim=60mm 20mm 30mm 25mm]{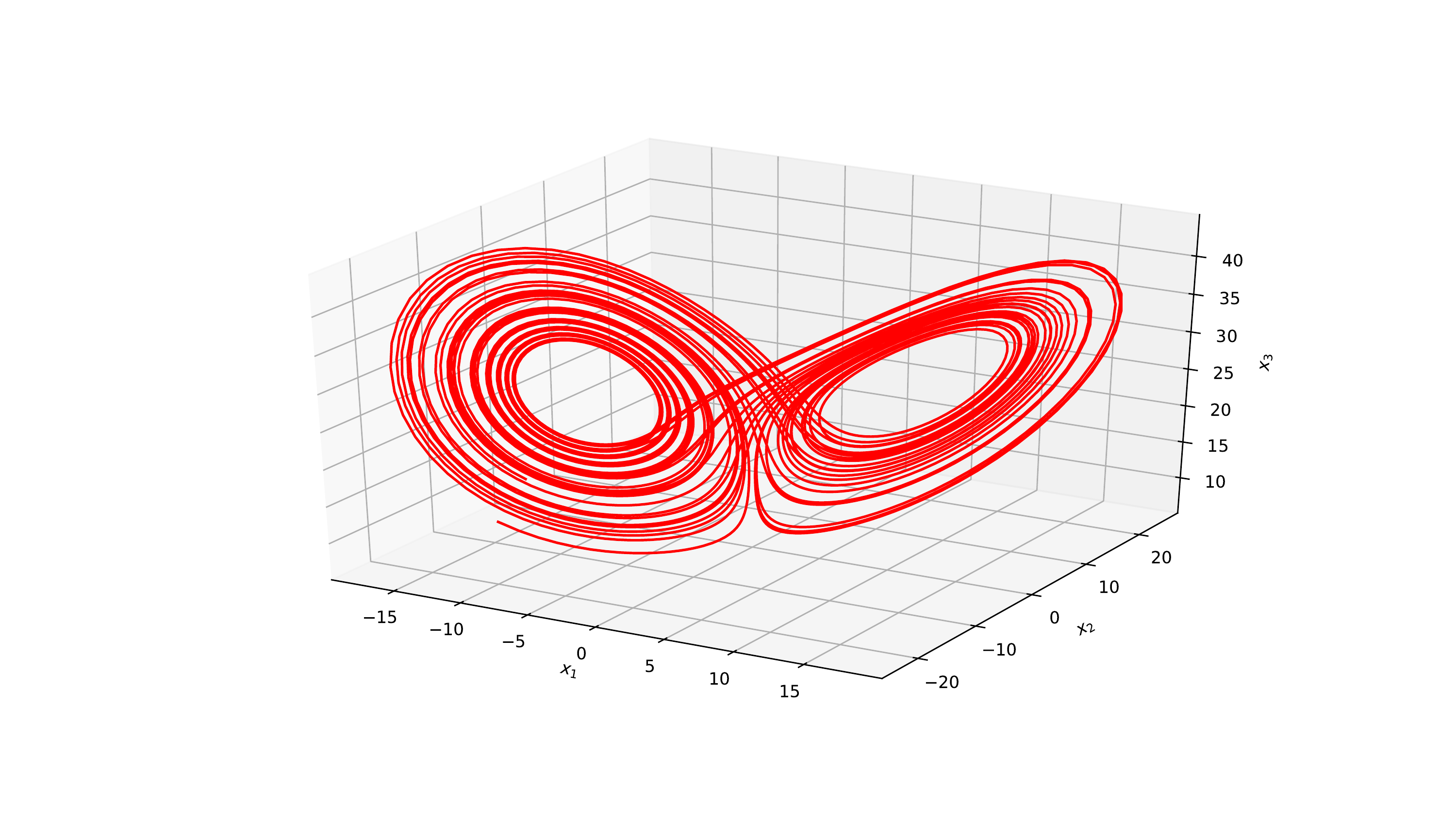} 
    \caption{The attractor of the Lorenz--63 system when $\sigma =10$, $\rho=28$ and  $\beta=8/3$.}
    \label{fig:true_lorenz}
\end{figure}

When $\sigma =11$, $\rho=28$ and  $\beta=8/3$, this system has a chaotic behavior, with the Lorenz attractor shown in Fig. \ref{fig:true_lorenz}.

Some characteristics of the L63 with the above set of parameters are as follows: 
\begin{itemize}
    \item The system is chaotic, a minor change in the initial condition will lead to a completely different trajectory in long term.
    \item The attractor of the L63 has a ``butterfly form", the particles frequently change side of the attractor. The density of the particles in two sides of the attractor is also similar.
\end{itemize}



\subsection{The Lorenz-96 system}
\label{subsec:L96}

The Lorenz-96 system (L96) \cite{lorenz_predictability:_1996} is a periodic 40-dimensional dynamical system governed by the following ODEs:

\textit{For $i = 1,..N_z$}: 
\begin{equation}
    \frac{\mathrm{d}\vect{z}_{t,i}}{\mathrm{d}t} = (\vect{z}_{t,i+1} - \vect{z}_{t,i-2})\vect{z}_{t,i-1} - \vect{z}_{t,i} + F       
\end{equation}

with $N_z = 40$,  $\vect{z}_{t,-1} = \vect{z}_{t,N_z-1}$,  $\vect{z}_{t,0} = \vect{z}_{t,N_z}$ and  $\vect{z}_{t,N_z+1} = \vect{z}_{t,1}$. 

We choose $F = 8$ to have chaotic system. 

\subsection{The stochastic Lorenz-63 system}
\label{subsec:L63s}

The stochastic Lorenz-63 system (L63s) is presented in \cite{chapron_large-scale_2018}. It is a modified version of the L63 to model situations where the large-scale characteristics of a physical event may be changed because of accumulated perturbations in fine scales. The governing equations of the L63s are as follow:

\begin{equation}
\begin{aligned}
    \label{eq:lorenz-63s}
    \mathrm{d}\vect{z}_{t,1} &=\left( \sigma \left( \vect{z}_{t,2}-\vect{z}_{t,1} \right) - \frac{4}{2\gamma}\vect{z}_{t,1} \right)\mathrm{d}t
    \\
    \mathrm{d}\vect{z}_{t,2} &=\left(\left(\rho - \vect{z}_{t,3} \right) \vect{z}_{t,1}-\vect{z}_{t,2} - \frac{4}{2\gamma}\vect{z}_{t,2} \right)\mathrm{d}t
    + \frac{\rho - \vect{z}_{t,3} }{\gamma^{0.5}}\mathrm{d}B_t
    \\
    \mathrm{d}\vect{z}_{t,3} &= \left(\vect{z}_{t,1}\vect{z}_{t,2}-\beta \vect{z}_{t,3} - \frac{8}{2\gamma}\vect{z}_{t,3} \right)\mathrm{d}t
    + \frac{\vect{z}_{t,2}}{\gamma^{0.5}}\mathrm{d}B_t
\end{aligned}
\end{equation}
with $B_t$ a Brownian motion.

In the L63s, the noise level is controlled by $\gamma$. The data used in this paper were generated with $\sigma =11$, $\rho=28$ and  $\beta=8/3$ and $\gamma=5$. With this set of parameters, the particles are easily trapped in one side of the attractor, as shown in Fig. 9 in the paper.
\section{Model setup}
\label{sec:setup}

\subsection{Models used for the L63 and the L63s}
\label{subsec:model_L63_L63s}

All the four models (BiNN\_EnKs, DAODEN\_determ, DAODEN\_MAP and DAODEN\_full) use the same dynammical sub-module: a BiNN. The architecture of this network is presented in Table. \ref{tab:BiNN_architecture}. The terms Linear and Bilinear are for the Linear and the Bilinear modules implemented in Pytorch.

\begin{table}[H]
    \caption{Architecture of the BiNN used for the L63 and the L63s.}
    \label{tab:BiNN_architecture}
    \centering
    \begin{tabular}{l|c}
        \toprule
        \textbf{Parameter} & \textbf{Value}  
        \\
        \midrule
        Number of Linear cells & $1$
        \\
        \midrule
        Linear cell size & $\left[3,3\right]$
        \\
        \midrule
        Linear cell activation & Linear
        \\
        \midrule
        Number of Bilinear cells & $3$
        \\
        \midrule
        Bilinear cell size & $\left[3,3,3\right]$
        \\
        \midrule
        Bilinear activation & Linear
        \\
        \bottomrule
    \end{tabular}
\end{table}

For BiNN\_EnKS, we used the EnKS implementation suggested in \cite{evensen_ensemble_2000}. The size of the ensemble was choosen as 50.  

As shown in Fig. 2 in the paper,
the inference scheme of DAODEN models is an LSTM-based network. The parameters of the inference sub-module of DAODEN\_full is presented in Table. \ref{tab:lstm_L63_architecture}. All the encoders and the decoders are MLPs. Similar architectures were used for DAODEN\_determ and DAODEN\_MAP, by removing the variance parts.

\begin{table}[H]
    \caption{Architecture of the inference scheme of DAODEN\_full used for the L63 and the L63s.}
    \label{tab:lstm_L63_architecture}
    \centering
    \begin{tabular}{l|c}
        \toprule
        \textbf{Parameter} & \textbf{Value}  
        \\
        \midrule
        LSTM layers & $2$
        \\
        \midrule
        LSTM hidden state dimension & $9$
        \\
        \midrule
        $MLP^{enc}$ size & $\left[3,7,3\right]$
        \\
        \midrule
        $MLP^{enc}$ activation & ReLU
        \\
        \midrule
        $MLP^{dec}$ size & $\left[21,7,6\right]$
        \\
        \midrule
        $MLP^{dec}$ activation & ReLU
        \\
        \bottomrule
    \end{tabular}
\end{table}

\subsection{Models used for the L96}
\label{subsec:model_L63_L63s}

For the L96, we used the convolutional version of BiNN, as presented in \cite{bocquet_bayesian_2020}. 


The architecture of the inference scheme is presented in Table. \ref{tab:lstm_L96_architecture}. 

\begin{table}[H]
    \caption{Architecture of the inference scheme of DAODEN\_determ used for the L96.}
    \label{tab:lstm_L96_architecture}
    \centering
    \begin{tabular}{l|c}
        \toprule
        \textbf{Parameter} & \textbf{Value}  
        \\
        \midrule
        LSTM layers & $2$
        \\
        \midrule
        LSTM hidden state dimension & $80$
        \\
        \midrule
        $MLP^{enc}$ size & $\left[40,80,40\right]$
        \\
        \midrule
        $MLP^{enc}$ activation & ReLU
        \\
        \midrule
        $MLP^{dec}$ size & $\left[200,80,40\right]$
        \\
        \midrule
        $MLP^{dec}$ activation & ReLU
        \\
        \bottomrule
    \end{tabular}
\end{table}

\subsection{Models used for the L63 with Legendre observations}
\label{subsec:model_L63LG}

The dynamical sub-module of the DAODEN\_determ model used in Section
V-H 
is the same as the one presented in Section \ref{subsec:model_L63_L63s}. 
The architecture of the inference scheme used in Section V-H 
is presented in Table. \ref{tab:lstm_L63LG_architecture}.

\begin{table}[H]
    \caption{Architecture of the inference scheme of DAODEN\_determ used for the L63 with Legendre observations}
    \label{tab:lstm_L63LG_architecture}
    \centering
    \begin{tabular}{l|c}
        \toprule
        \textbf{Parameter} & \textbf{Value}  
        \\
        \midrule
        LSTM layers & $2$
        \\
        \midrule
        LSTM hidden state dimension & $9$
        \\
        \midrule
        $MLP^{enc}$ size & $\left[128,64,32,3\right]$
        \\
        \midrule
        $MLP^{enc}$ activation & Sigmoid
        \\
        \midrule
        $MLP^{dec}$ size & $\left[21,32,64,128\right]$
        \\
        \midrule
        $MLP^{dec}$ activation & Sigmoid
        \\
        \bottomrule
    \end{tabular}
\end{table}
\section{Simulation of stochastic dynamics}
\label{sec:stochastic_simulation}

To simulate a stochastic sequence given the learnt stochastic model ($\mathcal{F}$ and $MLP^{var\_dyn}$ in the case of DAODEN\_full), we use the following algorithm:

\begin{algorithm}[t]
\caption{Generate stochastic sequence}
\label{alg:l63s}
 \KwResult{A sequence $\vect{S}$ of length $N$, generated by the model \{$\mathcal{F}$,$MLP^{var\_dyn}$\}, starting form the initial condition $\vect{x}_0$.}
 \textbf{Inputs}: $N$, $\mathcal{F}$, $MLP^{var\_dyn}$, $\vect{x}_0$\;
 $\vect{x} = \vect{x}_0$\;
 $\vect{S} = list()$\;
  t = 0\;
 \While{$t < N$}{
  $\mu = \mathcal{F}^1(\vect{x})$\;
  $\vect{d}^{dyn} = MLP^{var\_dyn}(\vect{x})$\;
  $\vect{x} \sim \mathcal{N}(\mu,\vect{d}^{dyn}\vect{I})$\;
  $\vect{S}$.append($\vect{x}$)\;
 }
\end{algorithm}


\end{document}